\documentclass{article}

\usepackage{microtype}
\usepackage{graphicx}
\usepackage{subfigure}
\usepackage{booktabs} 

\usepackage{threeparttable}
\usepackage{amsmath}
\usepackage{xcolor}
\usepackage{amssymb}
\usepackage{graphicx}
\usepackage{booktabs}
\usepackage[colorlinks=true, allcolors=blue]{hyperref}
\definecolor{red}{HTML}{D9423C}
\definecolor{purple}{HTML}{854C98}

\usepackage{hyperref}

\usepackage{amsmath}
\usepackage{amssymb}
\usepackage{mathtools}
\usepackage{amsthm}

\usepackage[capitalize,noabbrev]{cleveref}

\theoremstyle{plain}
\newtheorem{theorem}{Theorem}[section]
\newtheorem{proposition}[theorem]{Proposition}
\newtheorem{lemma}[theorem]{Lemma}

\theoremstyle{definition}
\newtheorem{definition}[theorem]{Definition}

\theoremstyle{remark}

\newcommand\numberthis{\addtocounter{equation}{1}\tag{\theequation}}

\usepackage{PRIMEarxiv}

\usepackage[utf8]{inputenc} 
\usepackage[T1]{fontenc}    
\usepackage{hyperref}       
\usepackage{url}            
\usepackage{booktabs}       
\usepackage{amsfonts}       
\usepackage{nicefrac}       
\usepackage{microtype}      
\usepackage{lipsum}
\usepackage{fancyhdr}       
\usepackage{graphicx}       
\graphicspath{{media/}}     

\title{Reveal the Mystery of DPO: The Connection between DPO and RL Algorithms 
}

\author{
  Xuerui Su\thanks{These authors contributed equally to this work.} \\
School of Mathematics and Statistics \\
Beijing Jiaotong University \\
\texttt{24110486@bjtu.edu.cn} \\
\And
Yue Wang\footnotemark[1] \\
Independent Researcher \\
\texttt{yuewang\_yw@foxmail.com} \\
   \And
   Jinhua Zhu \\
  University of Science and Technology of China \\
\texttt{teslazhu@mail.ustc.edu.cn} \\
   \And
   Mingyang Yi \\
  School of information \\
  Renmin University of China \\
  \texttt{yimingyang@ruc.edu.cn} \\
   \And
   Feng Xu \\
  School of Management \\
  Fudan University \\
  \texttt{fxu23@m.fudan.edu.cn} \\
   \And
  Zhiming Ma \\
  Academy of Mathematics and Systems Science \\
  \texttt{mazm@amt.ac.cn} \\
   \And
  Yuting Liu\thanks{Corresponding Author}  \\
School of Mathematics and Statistics \\
  Beijing Jiaotong University \\
  \texttt{ytliu@bjtu.edu.cn} \\
}

\begin{document}
\footnotetext[1]{These authors contributed equally to this work.}
\maketitle

\begin{abstract}
With the rapid development of Large Language Models (LLMs), numerous Reinforcement Learning from Human Feedback (RLHF) algorithms have been introduced to improve model safety, and alignment with human preferences. These algorithms can be divided into two main frameworks based on whether they require an explicit reward (or value) function for training: actor-critic-based Proximal Policy Optimization (PPO) and alignment-based Direct Preference Optimization (DPO). The mismatch between DPO and PPO, such as DPO’s use of a classification loss driven by human-preferred data, has raised confusion about whether DPO should be classified as a Reinforcement Learning (RL) algorithm. To address these ambiguities, we focus on three key aspects related to DPO, RL, and other RLHF algorithms: (1) the construction of the loss function; (2) the target distribution at which the algorithm converges; (3) the impact of key components within the loss function. Specifically, we first establish a unified framework named UDRRA connecting these algorithms based on the construction of their loss functions. Next, we uncover their target policy distributions within this framework. Finally, we investigate the critical components of DPO to understand their impact on the convergence rate. Our work provides a deeper understanding of the relationship between DPO, RL, and other RLHF algorithms, offering new insights for improving existing algorithms.
\end{abstract}


\section{Introduction}
\label{Intro}

Large Language Models (LLMs) are considered one of the most promising advancements toward Artificial General Intelligence (AGI) \cite{zhao2023survey, ouyang2022training, minaee2024large, achiam2023gpt}. In practice, post-training the LLM with Reinforcement Learning from Human Feedback (RLHF) has significantly improved its safety, compliance, and the alignment with human preferences \cite{arumugam2019deep, singh2022flava, bai2022training, dai2023safe}.

{While well-established RLHF approaches, such as the Proximal Policy Optimization (PPO-RLHF) \cite{PPO-basedRLHF}, have demonstrated significant success, Direct Preference Optimization (DPO) \cite{DPO} has also emerged and garnered widespread attention. DPO differs from PPO in that it does not require explicit modeling of the reward function, making it simpler and more efficient in practice. Instead, DPO directly optimizes the language model to align with human preferences by minimizing a simple classification loss function, which is based solely on the log probability of outputs generated by the LLMs. Since its inception, DPO has gained considerable traction, leading to the development of several derivatives, such as IPO \cite{IPO} and DRO \cite{DRO_Deepmind}, among others.}

Despite its practical success, DPO differs significantly from typical Reinforcement Learning (RL) algorithms, particularly in its underlying algorithmic structure. While most RL algorithms, such as PPO-RL \cite{PPO}, rely on an actor-critic framework and explicitly model value functions, DPO takes a distinct approach. Under a Bradly-Terry \cite{bradley1952rank} framework to modeling human preference, DPO transferred training policy model into minimizing a human-preference labeled loss function, without requiring an explicit reward model or training value function with policy gradient. This fundamental difference leads to limited theoretical understanding of how DPO connects with standard RL algorithms, particularly in terms of its theoretical advantages. As a result, more research is needed to clarify the specific scenarios where DPO might offer distinct benefits over typical RL algorithms.



However, addressing the relationship between DPO and RL algorithms is a non-trivial task. On one hand, DPO is rooted in the optimal solution analysis of the KL-constrained reward maximization problem, based on PPO-RLHF. On the other hand, the absence of a reward function in DPO complicates its classification within the typical RL framework. As a result, recent research has even suggested that DPO may not be a true RL algorithm \cite{rlhfwithoutrl}, further highlighting the need for a deeper theoretical investigation. In this paper, we aim to bridge this gap by investigating the connections between DPO, RL, and other RLHF algorithms. Specifically, we focus on three key aspects:
\begin{itemize}
\item 1. What are the distinctions and connections of the construction of the loss function between DPO, RL and other representative RLHF algorithms? 
\item 2. What is the target distribution of these loss function?  
\item 3. How do the key components within these algorithms affect the algorithm performance?  
\end{itemize}

To do so, we fist construct a unified framework (Figure \ref{Framework_figure}) that uniformly covers the standard RL algorithms (i.e., PPO and Soft Actor Critic (SAC) \cite{SAC}) and the DPO-based algorithms i.e., (IPO, DRO, DPO). Based on the construction of loss functions of these algorithms, we clearly reveal the connections between them in our framework. Besides, the target distribution under this framework is also clearly revealed. Notably, though some existing literature \cite{xu2024dpo,ivison2024unpacking,yan20243d} have also explored the connection between DPO and PPO. However, they mainly focus on the technical details of training LLM and the comparison of experimental results between DPO and PPO.
In contrast, our framework is more general, and is not restricted to any specific model scenario.

Then we analyze the target distributions of the series of methods mentioned in our framework. Although these methods have different requirements for the reward function, the same versions (e.g. the posterior version) all share the same target distribution. Furthermore, by analyzing the relationship between DPO and the PRA-P method in our framework, we proved that the target distribution of DPO is not $\bar{\pi}^\tau$ introduced in the original DPO's paper. Finally, considering the importance of DPO in training RLHF models, we further investigate the convergence rate of the DPO algorithm. Building on our theoretical results, we analyze how the hyper-parameter $\tau$ and the offline dataset of the algorithm influence its convergence performance.



To the best of our knowledge, we are the first to establish such a unified framework that provides a cohesive perspective on DPO Algorithm and a variety of RLHF algorithms. Our explorations provide us deeper understanding and new insights in improving the existing algorithms for RLHF. The contributions in this paper can be summarized as follows:
 
\begin{itemize}
\item  We established a unified framework to bridge the theoretical gap between DPO and RL algorithms.
\item We analyzed the target distribution of the loss function for algorithm in our Framework.
\item We explored the impact of the hyper parameter $\tau$ and the preference dataset on algorithm performance.
\end{itemize}

\section{Preliminary}\label{Preliminary}

\subsection{Problem Setup}

Consider a set of state or prompt  $\mathbb{X}$. Denote the response space as $\mathbb{Y}$, and the reward function as $r: \mathbb{X} \times \mathbb{Y} \rightarrow \mathbb{R}$. The optimization problem that most RLHF algorithm aim to solve is defined as where $\mathcal{D}$ is arbitrary distribution:
\begin{equation}\label{RLOpt}
\small{\min _{\pi_\theta} J \triangleq \min _{\pi_\theta} \mathbb{E}_{x \sim \mathcal{D}, y \sim \pi_\theta(y \mid x)}\left[-r(x, y)\right],}
\end{equation}
and define the target distribution $\pi^\delta=\arg\min _{\pi_\theta} J$. 

\begin{definition}\label{BDA}
Boltzmann Distribution Approximation Problem (BDAP). Motivated by the idea of Soft Actor Critic algorithm \cite{SAC} that they update the policy towards the exponential of the new Q-function, another important problem we consider is the Boltzmann Distribution Approximation Problem:
\begin{equation}\label{BDA_loss_eq}\small
\begin{gathered}
\min _{\pi^{\prime} \in \Pi} \mathbb{E}_{x \sim \mathcal{D}}\left[\mathrm{D}_{\mathrm{KL}}\left(\pi^{\prime}\left(\cdot \mid x\right) \| \pi^\tau\left(\cdot \mid x\right)\right)\right],\quad \pi^\tau\left(\cdot \mid x\right)\triangleq \frac{\exp \left(\tau r\left(x, \cdot\right)\right)}{Z\left(x\right)}.
\end{gathered}\normalsize
\end{equation}
\end{definition}
Problem (\ref{BDA_loss_eq}) describes the objective of optimizing $\pi^{\prime}$ towards the Boltzmann distribution of reward function by the KL divergence, though in principle any distribution distance is suitable. The partition function $Z\left(x\right)=\sum_{y\in\mathbb{Y}}\exp \left(\tau r\left(x, y\right)\right)$ normalizes the distribution.  $\tau$ is the temperature parameter. $\Pi$ is the set of policies. $\pi^\tau$ is essentially a soft approximation of the optimal solution $\pi^\delta$ of Problem (\ref{RLOpt}), while $\pi^\delta$ is sharp. This approximation is referred to as the Boltzmann approximation in physics. Proposition \ref{tau_tend_to_delta} shows that $\pi^\tau$ converges to $\pi^\delta$.

\begin{proposition}\label{tau_tend_to_delta}
Without loss of generality, assume that the reward function $r(x, y)$ has a unique maximum for any given $x$. Recall
$$\small \pi^\delta(y|x) = 
\begin{cases} 
1, & \text{if } y = \arg\max_{y^{\prime}} r(x, y^{\prime}), \\
0, & \text{otherwise}.
\end{cases} \normalsize$$
Then we have:
$  \lim_{\tau \to \infty} \pi^\tau(y|x) = \pi^\delta(y|x).$  
\end{proposition}

\subsection{Related Works}


\textbf{PPO-RL:} The Proximal Policy Optimization (PPO) algorithm \cite{PPO} in RL is based on the Trust Region Policy Optimization (TRPO) algorithm \cite{TRPO}. The KL penalty version of PPO in RL is below: 
\begin{equation}
\begin{aligned}
\mathcal{L}_{\text{PPO}}(\pi_\theta)=\hat{\mathbb{E}}_t[&-\frac{\pi_\theta\left(a_t \mid s_t\right)}{\pi_{\theta_{\text {old }}}\left(a_t \mid s_t\right)}\hat{A}_t +\frac{1}{\tau}\mathrm{D}_{\mathrm{KL}}\left(\pi_{\theta_{\text {old }}}(\cdot|s_t)\|\pi_\theta(\cdot|s_t)\right)].
\end{aligned}
\end{equation}
where $\theta_{\text{old}}$ is the vector of policy parameters before the update. $\hat{A}_t$ is an estimator of the advantage function at timestep $t$. For one step policy optimization problem, $\hat{A}_t=r(s_0,a_0)$ where $s_0, a_0$ are taken as the prompt and response separately. Based on the theory of TRPO, PPO-RL will converge to the optimal solution $\pi^\delta$ of Problem (\ref{RLOpt}).

\textbf{PPO-RLHF:} The PPO based RLHF \cite{PPO-basedRLHF} typically consists of two major stages: reward modeling and RL optimization. In the first stage, human annotators select the preferred answer $y_w$ over the less preferred one $y_l$, forming the preference pair $(y_w, y_l, x)$ based on a given input $x$. A reward model $r_\phi(x, y)$ is optimized by: 
\begin{equation}
\small{\mathcal{L}_R\left(r_\phi\right)=-\mathbb{E}_{\left(x, y_w, y_l\right) \sim \mathcal{D}_R}\left[\log \sigma\left(r_\phi\left(x, y_w\right)-r_\phi\left(x, y_l\right)\right)\right].}
\end{equation}
where $\sigma(\cdot)$ is the sigmoid function, $\mathcal{D}_{R}\triangleq \{(x,y_w,y_l)|x\sim\mathcal{D},y_w,y_l\sim\pi_0(\cdot|x),(y_w\succ y_l)\sim p^*(1|y_w,y_l,x)\}$, $\pi_0$ is the offline data sampling distribution. $\{z=1|y_1,y_2,x\}\triangleq \{r(x,y_1)\geq r(x,y_2)\}$ and $p^*$ is modeled by the BT model. In the RL optimization stage, the trained reward model evaluates the outputs of the language model, guiding policy optimization with loss function of the classic PPO algorithm:
\begin{equation}\label{PPO_loss}
\small{\mathbb{E}_{x \sim \mathcal{D}, y \sim \pi_\theta(\cdot\mid x)}\left[-r_\phi(x, y)+\frac{1}{\tau} \mathrm{D}_{\mathrm{KL}}\left(\pi_\theta(\cdot|x) \| \pi_{ref}(\cdot|x)\right)\right].}
\end{equation}
Typically the reference policy $\pi_{ref}(\cdot|x)$ is a pretrained or Supervised Fine-Tuned (SFT) model.

\textbf{DPO:} The Direct Preference Optimization (DPO) \cite{DPO} leverages the optimal policy form in Eq.\ref{PPO_loss} as theoretical support by representing the comparison probability (calculated under the assumption of BT model) of human preferences through the ratio between the policy $\pi_\theta$ and the reference policy $\pi_{ref}$. This approach eliminates the need for explicitly modeling the reward function. Thus DPO directly optimizes the policy by maximizing the log-likelihood function based on human preference feedback. 
\begin{equation}\label{DPO_eq}
\small{\begin{gathered}
\bar{h}_\theta\left(x, y_w, y_l\right)=\frac{1}{\tau} \log \frac{\pi_\theta\left(y_w \mid x\right)}{\pi_{\text {ref }}\left(y_w \mid x\right)}-\frac{1}{\tau} \log \frac{\pi_\theta\left(y_l \mid x\right)}{\pi_{\text {ref }}\left(y_l \mid x\right)},\ 
\mathcal{L}_{\mathrm{DPO}}\left(\pi_\theta ; \pi_{\text {ref }}\right)=-\mathbb{E}_{\left(x, y_w, y_l\right) \sim \mathcal{D}_R}\left[\log \sigma\left(\bar{h}_\theta\left(x, y_w, y_l\right)\right)\right].
\end{gathered}}
\end{equation}

\begin{figure*}[!ht]
\begin{center}
\centerline{\includegraphics[width=0.9\columnwidth,]{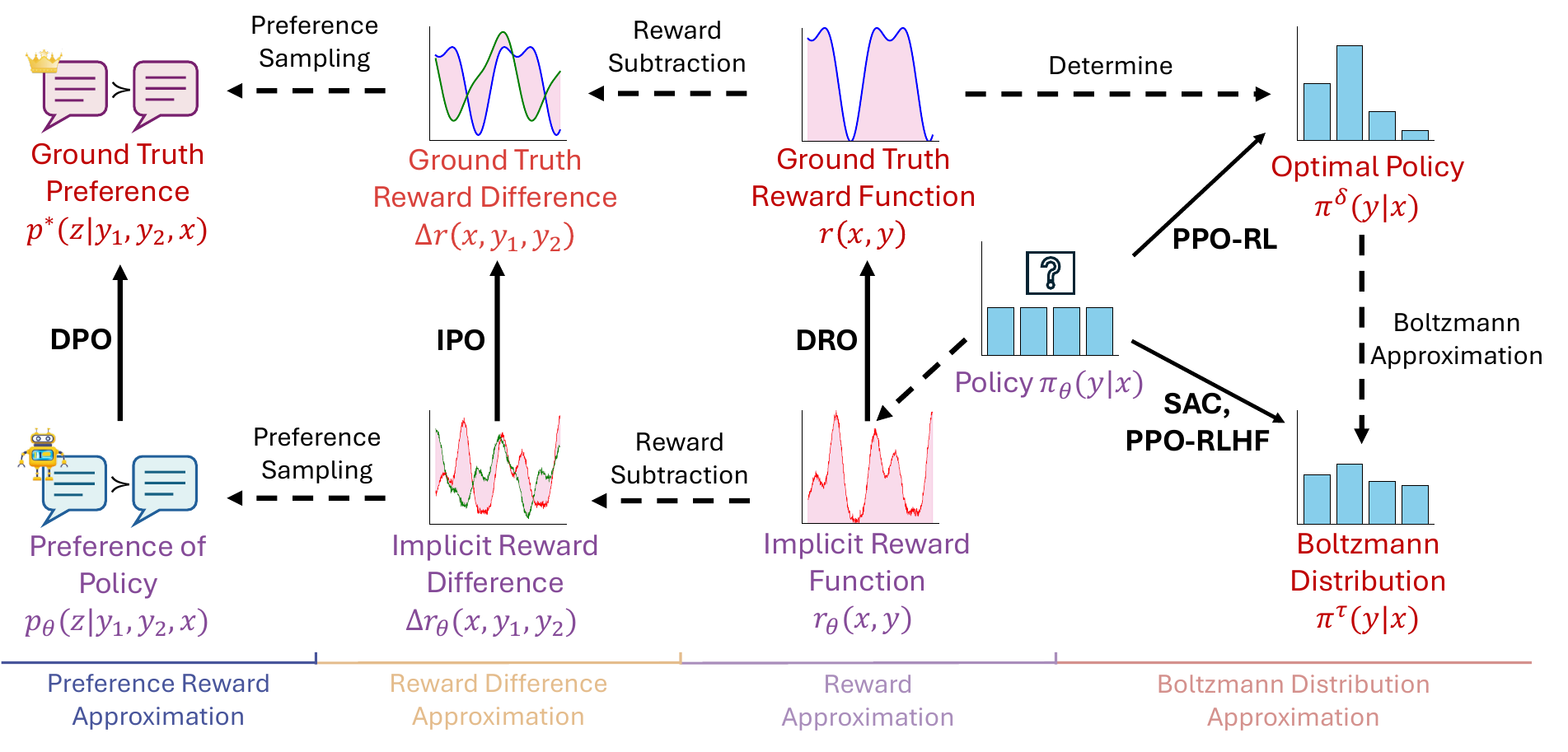}}
\vskip -0.1in
\caption{The unified UDRRA framework to connect DPO and other RLHF algorithms with typical RL methods. The algorithms in our framework can construct loss functions for four different scenarios (from right to left): (1) between \textcolor{purple}{$\pi_\theta(y|x)$} and \textcolor{red}{$\pi^\delta(y|x)$} (or \textcolor{red}{$\pi^\tau(y|x)$}); (2) between \textcolor{purple}{$r_\theta(x,y)$} and \textcolor{red}{$r(x,y)$}; (3) between \textcolor{purple}{$\Delta r_\theta(x,y_1,y_2)$} and \textcolor{red}{$\Delta r(x,y_1,y_2)$}; and (4) between \textcolor{purple}{$p_\theta(z|y_1,y_2,x)$} and \textcolor{red}{$p^*(z|y_1,y_2,x)$}. From scenario (1) to scenario (4), the requirements of the reward function $r(x,y)$ are progressively relaxed. The notations in the figure will be elaborated in Section \ref{Preliminary} and Section \ref{TRPRA_sec}. Notably, we utilize Proposition \ref{tau_tend_to_delta} to illustrate the relationship between $\pi^\delta$ and $\pi^\tau$. }
\label{Framework_figure}
\end{center}
\vskip -0.3in
\end{figure*}

\section{The Framework to \underline{U}nify \underline{D}PO, \underline{R}L and Other Representative \underline{R}LHF \underline{A}lgorithms}\label{TRPRA_sec}
In this section, we unify DPO, RL, and other representative RLHF algorithms within a unified framework (see Figure \ref{Framework_figure}) based on
the construction of loss functions. The distinctions of loss functions are mainly because how loss functions are defined across various reward function scenarios. Our framework clarifies these relationships, highlighting each method's strengths and use cases while bridging RL principles with RLHF techniques to improve methods like DPO.
\subsection{The UDRRA Framework}\label{Framework_sec}
Most RL algorithms aim to solve Problem (\ref{RLOpt}): given a reward function, find the optimal policy $\pi^\delta$. To balance exploration and exploitation, the target distribution $\pi^\delta$ is often approximated by a Boltzmann distribution $\pi^\tau$, turning the optimization into a distribution approximation problem. 
The critical challenge lies in designing a loss function that enables the parameterized policy $\pi_\theta$ to effectively approximate the target distribution $\pi^\tau$ through gradient-based methods. Our framework emphasizes the differences in loss function construction across various algorithms. Specifically, the construction of loss functions in current algorithms primarily addresses four types of scenarios involving policy distributions, reward functions, reward differences, and preferences, corresponding to scenarios (1)-(4) in Figure \ref{Framework_figure}. 


In scenario (1), the loss function aims to directly approximate $\pi^\delta$ or $\pi^\tau$ using $\pi_\theta$, as seen in algorithms like PPO-RL, SAC, and PPO-RLHF. In scenario (2), an implicit reward function $r_\theta(x, y)$ is constructed from $\pi_\theta$ (Eq.\ref{r_theta_def}) to minimize its difference from the true reward function $r(x, y)$, guiding $\pi_\theta$ towards $\pi^\tau$. However, this requires the partition function $Z(x)$, which is discussed later. Scenario (3) avoids calculating $Z(x)$ by aligning reward differences $\Delta r_\theta(x, y_1, y_2)$ and $\Delta r(x, y_1, y_2)$, focusing on differences rather than exact values.

In many real-world applications, noise often makes it difficult to determine the exact reward difference, as is common in RLHF scenarios. Instead, accessible information is typically a related random variable $ z $, reflecting the comparisons between reward values. In scenario (4), we model the conditional distribution of $ z $ based on the reward difference. Let $ p_\theta(z|y_1, y_2, x) $ approximate the true conditional preference distribution $ p^*(z|y_1, y_2, x) $. The loss function minimizes the distance between these distributions, relying on preference data without requiring the true reward function $ r(x, y) $.


In summary, from scenario (1) to scenario (4), the requirements on the reward function $r(x, y)$ are progressively relaxed. Our paper will illustrate the characteristics of loss function construction in different scenarios and prove the equivalence of different loss designs in the sense of target distribution. Furthermore, we will analyze the position of DPO within these scenarios and its relationship with our proposed method PRA-P for offering a new perspective for the theoretical investigation of DPO.

\subsubsection{Scenario (1): Boltzmann Distribution Approximation}\label{BDA_sec}
Proposition \ref{tau_tend_to_delta} supports the Boltzmann approximation, allowing $\pi^\tau$ to relax $\pi^\delta$. This relaxation is validated by RL algorithms like SAC. Therefore, in scenario (1) there is a natural relaxation method to approximate the solution to Problem (\ref{RLOpt}) by solving $\pi^\tau$. Here, we summarize the method based on the Boltzmann Distribution Approximation Problem in Definition \ref{BDA} using the KL divergence as the loss function. For the Forward-KL and Reverse-KL, we have Eq.\ref{forward-KL} and Eq.\ref{reverse-KL} (See derivation in Appendix \ref{forward-KL_proof}, \ref{reverse-KL_proof}):
{\begin{align*}
&\mathcal{L}_{\mathrm{Forward-BDA}}(\pi_\theta)=\mathbb{E}_{x\sim \mathcal{D}}\left[\mathrm{D}_{\mathrm{KL}}\left(\pi_\theta\left(\cdot \mid x\right) \| \pi^\tau\left(\cdot \mid x\right)\right)\right]=-\mathbb{E}_{x\sim \mathcal{D},y\sim\pi_\theta(\cdot|x)}\left[(\tau r(x,y) - \log(\pi_\theta(y|x)))\right], \numberthis\label{forward-KL} \\
&\mathcal{L}_{\mathrm{Reverse-BDA}}(\pi_\theta)=\mathbb{E}_{x\sim \mathcal{D}}\left[\mathrm{D}_{\mathrm{KL}}\left( \pi^\tau\left(\cdot \mid x\right) \| \pi_\theta\left(\cdot \mid x\right) \right)\right]=-\mathbb{E}_{x\sim \mathcal{D},y\sim\pi^\tau\left(\cdot \mid x\right)}[\log\pi_\theta(y|x)].\numberthis\label{reverse-KL}
\end{align*}}
We name the method using Eq.\ref{forward-KL} or Eq.\ref{reverse-KL} as the loss function as the \textbf{Boltzmann Distribution Approximation (BDA)} method. Eq.\ref{forward-KL} is exactly the loss function of SAC. 
\subsubsection{Scenario (2): Reward Approximation}
Denote the implicit reward function $r_\theta(x, y)$:
\begin{equation}\label{r_theta_def}
\small{r_\theta(x,y)\triangleq \frac{1}{\tau}\log(Z(x)\pi_\theta(y|x)).}
\end{equation}
Scenario (2) focuses on the intuition of using $r_\theta(x,y)$ to approximate the ground truth $r(x,y)$. Considering using $r_\theta(x,y)$ to approximate $r(x,y)$ is a typical regression problem, the Mean Square Error (MSE) is natural to be the loss function (See derivation in Appendix \ref{Derivation_RA_eq}):
\begin{equation}\label{RA_eq}
\small{\begin{aligned}
&\mathcal{L}_{\mathrm{RA}}(\pi_\theta)=\mathbb{E}_{x\sim \mathcal{D},y\sim\pi_\theta(\cdot|x)}\left[\left(r_\theta(x,y)- r(x,y)\right)^2\right] =\mathbb{E}_{x\sim \mathcal{D},y\sim\pi_\theta(\cdot|x)}\left[\frac{1}{\tau^2}\left(\log(\frac{\pi_\theta(y|x)}{\pi^\tau(y|x)})\right)^2\right].
\end{aligned}}
\end{equation}
We name the method using Eq.\ref{RA_eq} as the loss function as the \textbf{Reward Approximation (RA)} method. Due to the special design of $r_\theta(x,y)$, although the forms are different, Eq.\ref{forward-KL}, Eq.\ref{reverse-KL} and Eq.\ref{RA_eq} all effectively share the same target distribution. We will formally prove it later in Theorem \ref{policy_equivalence_1}.

In RLHF, the alignment task builds upon a reference model $\pi_{ref}(\cdot|x)$ rather than an initialized language model policy. Thus we incorporate $\pi_{ref}(y|x)$ via the posterior implicit reward function $\bar{r}_\theta(x,y)$. Treating $\pi_{ref}$ as the prior, the posterior distribution becomes $\bar{\pi}^\tau$:
\begin{equation}\label{pi_bar}
\small{\bar{\pi}^\tau(y|x)\triangleq \frac{\pi_{ref}(y|x)\exp(\tau r(x,y))}{\sum_{y'\in Y}\pi_{ref}(y'|x)\exp({\tau r(x,y')})}.}
\end{equation}
The corresponding posterior implicit reward function is 
\begin{equation}\label{r_theta_def_posterior}
\small{\bar{r}_\theta(x,y)=\frac{1}{\tau}\log(Z'(x)\frac{\pi_\theta(y|x)}{\pi_{ref}(y|x)}),}
\end{equation}
where $Z'(x)=\sum_{y'\in Y}\pi_{ref}(y'|x)\exp({\tau r(x,y')})$. Then Eq.\ref{RA_eq} will be changed into:
\begin{equation}\label{RA_eq_posterior}
\small{\begin{aligned}
&\mathcal{L}_{\mathrm{RA-P}}(\pi_\theta)=\mathbb{E}_{x\sim \mathcal{D},y\sim\pi_\theta(\cdot|x)}\left[\left(\bar{r}_\theta(x,y)- r(x,y)\right)^2\right] =\mathbb{E}_{x\sim \mathcal{D},y\sim\pi_\theta(\cdot|x)}\left[\left(\frac{1}{\tau}\log(Z'(x)\frac{\pi_\theta(y|x)}{\pi_{ref}(y|x)})- r(x,y)\right)^2\right].
\end{aligned}}
\end{equation}
We call the method using Eq.\ref{RA_eq_posterior} as loss function as the \textbf{{Reward Approximation-Posterior} (RA-P)} method. 

\subsubsection{Scenario (3): Reward Difference Approximation}
Define the general difference function $\Delta f(x, y_1, y_2)\triangleq f(x, y_1)-f(x, y_2)$. In scenario (3), the loss function is constructed to align the reward difference function, $\Delta r_\theta(x, y_1, y_2)=r_\theta(x,y_1)-r_\theta(x,y_2)$ and $\Delta r(x, y_1, y_2)=r(x,y_1)-r(x,y_2)$. Denote $\mathcal{D}_{\text{pw}}\triangleq\{(x, y_1, y_2) \mid x \in \mathcal{D}, y_1, y_2 \sim \pi_\theta(y|x)\}$. Similar to the RA method, using MSE, we have the loss function:
\begin{equation}\label{pair-wised_RA_eq}
\small{\begin{aligned}
&\mathcal{L}_{\mathrm{RDA}}(\pi_\theta)=\mathbb{E}_{\mathcal{D}_{\text{pw}}}\left[\left( \frac{1}{\tau}\log\frac{\pi_\theta(y_1|x)}{\pi_\theta(y_2|x)}-\Delta r(x, y_1, y_2)\right)^2\right].
\end{aligned}}
\end{equation}
The method using Eq.\ref{pair-wised_RA_eq} as the loss function is named \textbf{Reward Difference Approximation (RDA)}. Its posterior version, RDA-P, replaces $r_\theta$ with $\bar{r}_\theta$. All our methods have posterior versions, this paper only formulate RA-P and PRA-P as example. The RDA method simplifies computation by canceling the partition function $Z(x)$, which requires multiple reward function queries. Additionally, RDA only needs reward differences, not absolute values, while maintaining the same target distribution as BDA and RA. This is formally proven in Theorem \ref{policy_equivalence_2}.


\subsubsection{Scenario (4): Preference Reward Approximation}\label{Scenario4}
In scenario (4), we consider using $p^*(z \mid y_1, y_2, x) = \omega(r(x, y_{2-z}), r(x, y_{z+1}))$, where $z = 1$ if $r(x, y_1) > r(x, y_2)$, and $z = 0$ otherwise, to model the comparison probability between $r(x, y_1)$ and $r(x, y_2)$. Only $(x, y_1, y_2, z)$ sampled with $p^*(z \mid y_1, y_2, x)$ is available, while $r(x, y)$ is unobservable.  
The BDA, RA, and RDA methods are less effective in scenario (4). These methods rely on inverting $\omega$ to estimate $r(x, y)$. Table \ref{omega_table} and Appendix \ref{omega} discuss various $\omega$ formulations. In general, most $\omega^{-1}$ are difficult to compute because $p^*$ is not analytic. Thus developing methods that rely only on the forward evaluation of $\omega$ instead of $\omega^{-1}$ would greatly improve practicality in this scenario.

Therefore we propose the \textbf{Preference Reward Approximation (PRA)} method, assuming $\omega$ is known and data is sampled as $(x, y_w, y_l)$ where $\{r(x, y_w) > r(x, y_l)\}$ is true with probability $p^*(1\mid y_w, y_l, x)$. Define the parameterized comparison distribution as $p_\theta(z|y_1, y_2, x) = \omega(r_\theta(x, y_{2-z}), r_\theta(x, y_{z+1}))$. Assuming $\omega(x, y)$ is injective and satisfies the symmetric complementarity property $\omega(x, y) = 1 - \omega(y, x)$, the winning probability of $x$ against $y$ equals $y$'s losing probability against $x$, then we have:
\begin{equation}\label{PRA_eq}
\small{    \begin{aligned}
\mathcal{L}_{\mathrm{PRA}}(\pi_\theta)=&\mathbb{E}_{\mathcal{D}_{\theta}}\left[\mathrm{D}_{\mathrm{KL}}(p^*(z|y_1,y_2,x)||p_\theta(z|y_1,y_2,x))\right]\\
=&-\mathbb{E}_{(x,y_w,y_l)\sim \mathcal{D}_{\theta}}\left[\log \left( p_\theta(1|y_w,y_l,x)\right)\right]+\mathbb{E}_{x\sim \mathcal{D},y_1,y_2\sim \pi_\theta(y|x)}\left[M(x,y_1,y_2)\right].
\end{aligned}}
\end{equation}
Defer the proof in Appendix \ref{Derivation_PRA}. $\mathcal{D}_{\theta}\triangleq \{(x,y_w,y_l)|x\sim\mathcal{D},y_w,y_l\sim\pi_\theta(\cdot|x),(y_w\succ y_l)\sim p^*(1|y_w,y_l,x)\}$ and $M(x,y_1,y_2)=\sum_{z=0,1}p^*(z|y_1,y_2,x)\log p^*(z|y_1,y_2,x)$. 
PRA uses the KL divergence between $p^*$ and $p_\theta$ as the loss function, combining a cross-entropy term that avoids computing gradients of the reward function or $p^*$, and a regularization term involving integration over $p^*$ and $\pi_\theta$.

Similar to the RA-P method, we also propose a posterior version of PRA. Define $\bar{p}_\theta(z|y_1,y_2,x)=\omega(\bar{r}_\theta(x,y_{2-z}),\bar{r}_\theta(x,y_{z+1}))$. Based on Eq.\ref{PRA_eq}, we have:
\begin{equation}\label{PRA_posterior}
\small{    \begin{aligned}
\mathcal{L}_{\mathrm{PRA-P}}(\pi_\theta)=&\mathbb{E}_{\mathcal{D}_{\theta}}\left[\mathrm{D}_{\mathrm{KL}}(p^*(z|y_1,y_2,x)||\bar{p}_\theta(z|y_1,y_2,x))\right]\\
=&-\mathbb{E}_{(x,y_w,y_l)\sim \mathcal{D}_{\theta}}\left[\log \bar{p}_{\theta}\left(1|y_w, y_l,x\right)\right]+\mathbb{E}_{x\sim \mathcal{D},y_1,y_2\sim \pi_\theta(y|x)}\left[M(x,y_1,y_2)\right].
\end{aligned}}
\end{equation}
The method using Eq.\ref{PRA_posterior} as the loss function is name \textbf{{Preference Reward Approximation-Posterior} (PRA-P)} method. 
Firstly, PRA and PRA-P avoid calculating $\omega^{-1}$ while maintaining the same target distribution as previous methods. Theorem \ref{policy_equivalence_3} shows that PRA shares its target distribution with methods like BDA, while Theorem \ref{policy_equivalence_4} demonstrates that PRA-P and RA-P share the same target distribution. Secondly, the DPO algorithm aligns with scenario (4) and is closely related to PRA-P. In fact, the first term of Eq.\ref{PRA_posterior} matches the DPO loss \cite{DPO} if $\mathcal{D}_{\theta}$ is replaced with an offline dataset, though this introduces a distribution shift issue, which is discussed in Theorem \ref{DPO_PRA}.

\subsection{Summary}\label{Tree_sec}

Our framework is both logically reasonable and empirically valid. After developing the solution method for Problem \ref{BDA} based on different reward function scenarios, we found that previous studies have explored specific methodologies for the some scenarios and proposed efficient algorithms. For example, as shown in Section \ref{BDA_sec}, the loss function of SAC (Eq.10 in \cite{SAC}) aligns with the Forward-BDA method. To highlight these relationships, we summarize the correspondence between these algorithms and our methods in Table \ref{Corresponding_Algorithms}. Due to space constraints we only include a few algorithms to demonstrate the practical significance of our framework.

\begin{table}[ht] 
\centering
   \resizebox{0.4\hsize}{!}{
   \begin{tabular}{cc}
   \toprule
    Method & \#Corresponding Algorithms  \\
   \midrule
    Forward-BDA & SAC \cite{SAC}, DPG \cite{DPG} \\
    Reverse-BDA & RERPI \cite{RERPI} \\
    RA (RA-P)  & DRO \cite{DRO_Deepmind} \\
    RDA (RDA-P) & IPO \cite{IPO}, SVPO \cite{SVPO} \\
    PRA (PRA-P) & DPO \cite{DPO} \\
   \bottomrule
  \end{tabular}
  }
  \caption{Correspondence between the series of methods in our Framework and existing algorithms.}
  \label{Corresponding_Algorithms}
\end{table}
For scenario (1), the Reverse-BDA method, using reverse-KL (Eq.\ref{reverse-KL}) as a loss function for policy improvement, has been applied in finite state-action space control problems (see Eq.2 in RERPI\cite{RERPI}). However, in continuous state-action spaces, the term $ Z(x) $ is hard to compute, limiting the method's applicability. For scenario (2), the loss function of the Direct Reward Optimization (DRO) method matches the RA-P method’s loss function (Eq.\ref{RA_eq_posterior}), as shown the Eq.4 in \cite{DRO_Deepmind}. 
For scenario (3), the Step-level Value Preference Optimization (SVPO) algorithm \cite{SVPO} uses $\Delta r_\pi\left(\mathbf{s}_{t+1}^w, \mathbf{s}_{t+1}^l\right)$ to learn towards $\operatorname{sg}\left[\Delta r_\phi\left(\mathbf{s}_{t+1}^w, \mathbf{s}_{t+1}^l\right)\right]$ (Eq.10 in \cite{SVPO}), which corresponds to our RDA-P method. The RDA-P method is also a generalized version of Identity-PO (IPO) \cite{IPO}, relaxing the data requirement from querying reward differences to only ranking queries (i.e., $\mathbb{I}(r(x, y_1) - r(x, y_2))$, where $\mathbb{I}$ is the indicator function). Though $\mathbb{I}(r(x,y_1)-r(x,y_2))$ is more like a preference, we classify IPO as a relaxed version of RDA due to its similar form as RDA-P. Finally, for scenario (4), we will show in Theorem \ref{DPO_PRA} that DPO is simply an offline version of PRA-P.

\section{Target Distribution Analysis}\label{RL_Classification_sec}
In this section, we analyze the target distribution of the loss function mentioned in our framework. Although these methods have different requirements for the reward function, the same versions (e.g. the posterior version) all share the same target distribution. Additionally, as noted in Section \ref{Framework_sec}, we highlight how using an offline dataset in DPO introduces distribution shift issues, distinguishing it from online methods like SAC and PPO-RLHF.


\subsection{The Target Distribution Equivalence of Our Framework}
Here we state the approximation equivalence among the BDA, RA, RDA and PRA methods. 

\begin{theorem}\label{policy_equivalence_1}
 Define $\pi^*_{\mathrm{RA}}=\arg\min_{\pi_\theta}\mathcal{L}_{\mathrm{RA}}(\pi_\theta),\ \pi^*_{\mathrm{Forward-BDA}}=\arg\min_{\pi_\theta}\mathcal{L}_{\mathrm{Forward-BDA}}(\pi_\theta),\ \pi^*_{\mathrm{Reverse-BDA}}=\arg\min_{\pi_\theta}\mathcal{L}_{\mathrm{Reverse-BDA}}(\pi_\theta)$.
The following property holds:
\begin{equation}
    \pi^*_{\mathrm{Forward-BDA}} = \pi^*_{\mathrm{Reverse-BDA}} = \pi^*_{\mathrm{RA}} = \pi^\tau.
\end{equation}
\end{theorem}

See proof on Appendix \ref{Proof_Reward_policy}. Theorem \ref{policy_equivalence_1} demonstrate that the ``$\arg\min$'' of Eq.\ref{forward-KL}, \ref{reverse-KL}, \ref{RA_eq} are equivalent and equal to $\pi^\tau$. 

\begin{theorem}\label{policy_equivalence_2}
Recall $\mathcal{L}_{\mathrm{RDA}}(\pi_\theta)$ in Eq.\ref{pair-wised_RA_eq}. Define $\pi^*_{\mathrm{RDA}}=\arg\min_{\pi_\theta}\mathcal{L}_{\mathrm{RDA}}(\pi_\theta).$
Then $\pi^*_{\mathrm{RDA}} = \pi^\tau$.
\end{theorem}
Theorem \ref{policy_equivalence_2} proves that the target distribution corresponding to the loss function of the RDA method is $\pi^\tau$, which is consistent with the target distribution of the BDA and RA methods. See proof in Appendix \ref{Proof_RDA_lemma}.

\begin{theorem}\label{policy_equivalence_3}
Recall $\mathcal{L}_{\mathrm{PRA}}(\pi_\theta)$ in Eq.\ref{PRA_eq}. Define $\pi^*_{\mathrm{PRA}}=\arg\min_{\pi_\theta}\mathcal{L}_{\mathrm{PRA}}(\pi_\theta)$.
Then $\pi^*_{\mathrm{PRA}} = \pi^\tau$.
\end{theorem}
Theorem \ref{policy_equivalence_3} shows that although the loss function of the PRA method is designed by minimizing the distance between the two distributions $p^*, p_\theta$, its target distribution remains $\pi^\tau$. Defer proof in Appendix \ref{PRA_optimal_proof}. Similarly, we use Theorem \ref{policy_equivalence_4} to demonstrate the relationship between the target distributions of the posterior versions of the RA and PRA methods. See the proof in Appendix \ref{Reward_policy_posterior_proof}.

\begin{theorem}\label{policy_equivalence_4}
Recall $\mathcal{L}_{\mathrm{RA-P}}(\pi_\theta)$ in Eq.\ref{RA_eq_posterior} and $\mathcal{L}_{\mathrm{PRA-P}}(\pi_\theta)$ in Eq.\ref{PRA_posterior}. Define 
\begin{equation}
    \begin{aligned}
&\pi^*_{\mathrm{RA-P}}=\arg\min_{\pi_\theta}\mathcal{L}_{\mathrm{RA-P}}(\pi_\theta),\ \pi^*_{\mathrm{PRA-P}}=\arg\min_{\pi_\theta}\mathcal{L}_{\mathrm{PRA-P}}(\pi_\theta).
    \end{aligned}
\end{equation}
The following property holds:
\begin{equation}
    \pi^*_{\mathrm{RA-P}} = \pi^*_{\mathrm{PRA-P}} = \bar{\pi}^\tau.
\end{equation}
\end{theorem}
In summary, while the BDA, RA, RDA, and PRA methods are designed for different scenarios, they all share the same target distribution and aim to solve the same problem (BDAP) under varying conditions of reward function access. Specifically, BDA and RA require exact reward values, RDA only needs reward differences across variables $y$, and PRA relies on ordinal relationships of the reward function. Based on our analyzing about the methods in our UDRRA framework, we can more accurately select the appropriate method for different scenarios. 


\subsection{The Relationship between DPO and PRA-P (The Target Distribution of DPO)}\label{DPO_PRA_P_sec}
This section analyzes the target distribution of DPO. DPO mentions that the DPO target distribution is $\bar{\pi}^\tau$ which is also the the target distribution of the PRA-P method (Eq.4 in \cite{DPO}). However, we will show that there is a distribution shift between the the target distribution $\bar{\pi}^\tau$ of the PRA-P method and the the target distribution of DPO. We use Theorem \ref{DPO_PRA} to strictly analyze the differences and connections between DPO and PRA-P. 
\begin{theorem}
    \label{DPO_PRA}
When $p^*$ is modeled by the BT model:
\begin{equation}\label{BTmodel}
\begin{aligned}
p^*(1|y_1,y_2,x) &= \omega(r(x,y_{2-z}),r(x,y_{z+1}))=\sigma(r(x,y_1)- r(x,y_2) ),
\end{aligned}
\end{equation}
then we have the following equality:
\begin{equation}
\small{\mathcal{L}_{\mathrm{PRA-P}}\left(\pi_\theta\right) = \mathcal{L}_{\mathrm{DPO}}\left(\pi_\theta\right) + \eta_1(\pi_\theta, \pi_0) + \eta_2(\pi_\theta),}
\end{equation}
where $\eta_1(\pi_\theta, \pi_0)$ equals to 0 if and only if $\pi_\theta=\pi_0$ and $\eta_2(\pi_\theta)\triangleq\mathbb{E}_{x\sim \mathcal{D},y_1,y_2\sim \pi_\theta(y|x)}\left[M(x,y_1,y_2)\right]$. 
\end{theorem}
Theorem \ref{DPO_PRA} tells us that since $\mathcal{D}_{R}$ in DPO is an offline dataset, a shift term $\eta_1(\pi_\theta, \pi_0)$ is introduced between DPO and PRA-P. Because of the difference between $\pi_\theta$ and $\pi_0$, the existence of the non-zero term $\eta_1(\pi_\theta, \pi_0)$ makes the target distribution of DPO deviate from the target distribution $\bar{\pi}^\tau$ of the PRA-P method. We call this phenomenon as distribution shift. Defer the proofs to Appendix \ref{DPO_PRA_proof}. 
Moreover, $\eta_2(\pi_\theta)$ is a regularization term which serves to increase the information entropy ($-M(x,y_1,y_2)$) of the comparison probability $p^*$ under the current policy distribution $\pi_\theta$. The ignorance of this regularization term will make the DPO deviate further from the target distribution $\bar{\pi}^\tau$. Lastly, we emphasize that the $\omega$ function in the PRA method only needs to satisfy the symmetric complementarity property and is not limited to the BT model, which is a special case. We list more $\omega$ functions meeting this requirement in Table \ref{omega_table}.

\section{Components Influence in Algorithms within UDRRA}\label{Convergence_sec}

After analyzing the connections and distinctions between algorithms like DPO, the remaining question is how the common components in algorithms within UDRRA affect their performance. 
Here we focus on the impact of hyperparameters $\tau$ and preference datasets on algorithm performance.


\subsection{Q1: The Influence of $\tau$ on Algorithm Performance}\label{DPO_rate}
For Q1, Increasing hyper-parameter $\tau$ can speed up the convergence of DPO, which will formulate in Theorem \ref{DPO_theorem}. As $ \tau $ increases, the optimal solution $\bar{\pi}^\tau$ of DPO tends to $ \pi^\delta $ (See Proposition \ref{tau_tend_to_delta}), while deviating from $ \pi_{\text{ref}} $. Thus there exists a trade-off between the convergence rate and maintaining proximity to $ \pi_{\text{ref}}$. 
\begin{theorem}\label{DPO_theorem}
Assume $\pi_\theta$ is constructed by Definition \ref{Softmax}. Consider the DPO loss function  $\mathcal{L}_{\mathrm{DPO}}(\pi_{\theta} ; \pi_{\text {ref }})$. Given the learning rate $\alpha_t$ satisfying finite squared summability, suppose the parameters $\theta$ are updated by:
\begin{equation}\label{SGD_update}
    \theta_{t+1}=\theta_{t} - \alpha_t g(x,y_w,y_l,\theta_t),
\end{equation}
where $g(x,y_w,y_l,\theta_t)$ is a stochastic gradient of $\mathcal{L}_{\mathrm{DPO}}(\pi_{\theta_t} ; \pi_{\text {ref }})$. Assume $\|g(\cdot,\cdot,\cdot,\cdot)\|^2\leq G^2$. Denote $\mathcal{L}_{\mathrm{DPO}}^*=\min_{\pi_\theta} \mathcal{L}_{\mathrm{DPO}}(\pi_{\theta} ; \pi_{\text {ref }})$, then:
\begin{equation}
{    \begin{aligned}
\min_{1\leq i\leq T}&||\nabla_\theta \mathcal{L}_{\mathrm{DPO}}(\pi_{\theta_i} ; \pi_{\text {ref }})||^2_2\leq \frac{2G^2\sum_{t=1}^{T-1}\alpha_t^2}{\tau^2\sum_{t=1}^{T-1}\alpha_t} + \frac{\mathcal{L}_{\mathrm{DPO}}(\pi_{\theta_1})-\mathcal{L}_{\mathrm{DPO}}^*}{\sum_{t=1}^{T-1}\alpha_t}.
    \end{aligned}}
\end{equation}
\end{theorem}
See the proof in Appendix \ref{DPO_theorem_proof}. We assume the policy is a softmax policy (Definition \ref{Softmax}) and show that the gradient norm of the DPO loss function has an upper bound under non-convexity, which is negatively correlated with the hyper-parameter $\tau$. Thus we have that $\tau$ is inversely related to the convergence rate of DPO.

This paper focuses on the impact of $\tau$, rather than the design of the LLMs. Thus, we simplify Theorem \ref{DPO_theorem} to apply the convergence theorem for SGD in non-convex settings. Our argument is that, as long as the policy distribution is a softmax policy (Definition \ref{Softmax}), we show that the loss functions for all four methods in our framework, including DPO, are $L$-smooth with respect to the policy parameters $\theta$ leading to a sub-linear convergence rate of $O(\frac{1}{T})$.

\begin{table*}[ht] 
\centering
   \resizebox{0.85\hsize}{!}{
   \begin{tabular}{cccc}
   \toprule
    Method & \#Loss Functions & \#Smooth Coefficient ($L$) & \#$p(x,y;\pi_\theta)$ \\
   \midrule
    Forward-BDA & Eq.\ref{forward-KL}   & $6\epsilon_1+10$ & $\pi_\theta(y|x), x\sim\mathcal{D}$ \\
    Reverse-BDA  & Eq.\ref{reverse-KL}   & $2$ & $\pi^\tau(y|x), x\sim\mathcal{D}$\\
    RA & Eq.\ref{RA_eq} and Eq.\ref{RA_eq_posterior}   & $3\epsilon_1^2 + \frac{18\epsilon_1}{\tau} + \frac{8}{\tau^2} + \max\left\{ \epsilon_1^2 + \frac{2}{\tau} \epsilon_1, \frac{1}{\tau} \right\}$  & $\pi_\theta(y|x), x\sim\mathcal{D}$ \\
    RDA  & Eq.\ref{pair-wised_RA_eq}   & $20\epsilon_2^2+\frac{32\epsilon_2}{\tau}+\frac{8}{\tau^2}$  & $\pi_\theta(y|x), x\sim\mathcal{D}$ \\
    PRA  & Eq.\ref{PRA_eq} and Eq.\ref{PRA_posterior}   & $20\log(1+ e^{\frac{d}{\tau}})+\frac{16\epsilon_3}{\tau}+\frac{4}{\tau^2}+16\log2$  & $\pi_\theta(y|x), x\sim\mathcal{D}$ \\
    DPO & Eq.\ref{DPO_eq}  & $\frac{4}{\tau^2}$  & $\pi_0(y|x), x\sim\mathcal{D}$ \\
   \bottomrule
  \end{tabular}
  }
  \caption{Smooth Coefficients for different methods. $\pi_\theta$ is the current policy to be optimized, and $\pi_0$ is the sampling distribution for DPO's offline dataset. Let $|\log(\pi_\theta(y|x)) - \log(\pi^\tau(y|x))| \leq \epsilon_1$, where $\epsilon_1$ represents the maximum likelihood difference between $\pi_\theta$ and $\pi^\tau$. Let $|(r_\theta(x,y_1) - r_\theta(x,y_2)) - (r(x,y_1) - r(x,y_2))| \leq \epsilon_2$, where $\epsilon_2$ is the maximum deviation between $r_\theta$ and $r$. Define $d$ as the diameter of the smallest manifold sphere containing the domain of $\theta$ (as per Definition \ref{Softmax}), i.e. $|\theta(x_1,y_1) - \theta(x_2,y_2)| \leq d$. Similarly, let $|p^*(z|y_1,y_2,x) - p_\theta(z|y_1,y_2,x)| \leq \epsilon_3$, where $\epsilon_3$ bounds the difference between $\bar{p}_\theta(z|y_1,y_2,x)$ and $p^*(z|x,y_1,y_2)$. 
  }
  \label{smoothCoefficients}
  \vspace{-0.1in}
\end{table*}

Thus we summarize the convergence properties of DPO and related methods (BDA, RA, RDA, PRA) in Table \ref{smoothCoefficients}, corresponding to Lemma \ref{BDA_L_RKL_coef}-\ref{PRA_L_coef}. Our goal is to compare these methods from a unified perspective, focusing on the smoothness coefficient $L$, which influences convergence. The variations in $L$ arise from differences in how their loss functions are constructed. For Table \ref{smoothCoefficients}, we summarized several interesting conclusions below. 

First, the Forward-BDA method has a larger smooth coefficient ($6\epsilon_1 + 10$) than the Reverse-BDA method, indicating slower convergence but a more sampling-friendly policy $\pi_\theta$. Second, the RA method outperforms both BDA methods by setting $\epsilon_1$ and $\tau$ to reduce the smooth coefficient. For instance, when $\epsilon_1$ and $\tau$ satisfy $0 \leq \epsilon_1\leq\frac{\sqrt{1+\tau}-1}{\tau} \leq \frac{\sqrt{6}-1}{5}$, the smooth coefficient of the RA method is bounded by 2, ensuring faster convergence. Third, the smooth coefficients of the RDA and PRA methods can be reduced by adjusting $\epsilon_2, \epsilon_3, \tau$, and the parameter domain diameter $d$, to speed up their convergence. Lastly, the DPO method has a higher smooth coefficient than PRA, with additional terms like $20\log(1 + e^{\frac{d}{\tau}}) + \frac{16\epsilon_3}{\tau} + 16\log 2$, indicating that DPO converges faster than PRA due to the distribution shift bias.


\subsection{Q2: The Influence of Preference Dataset on Algorithm Performance}

Since DPO is a representative algorithm in RLHF, this paper uses it to answer Q2. Essentially, the design of preference dataset $\mathcal{D}_R$ depends on the design of the sampling strategy $\pi_0$, in other words, the relationship between the sampling strategy $\pi_0$ and the current strategy $\pi_\theta$. 
Thus for Q2, we use Theorem \ref{data_select_coro} to show that data $(x, y_1, y_2)$ should be selected where both the reward values and policy probabilities have a large and consistent margin between $y_1$ and $y_2$. Larger margins accelerate convergence, while consistent margins reduce distribution shift.

To prove Theorem \ref{data_select_coro}, we first consider a simplified question: should the sampling strategy $\pi_0$ prioritize samples with large or small reward differences or margins when identifying winners and losers? The challenge is that $\pi_0$ and $\pi_\theta$ are interdependent, as we optimize $\pi_\theta$. To address this, Lemma \ref{data_select} quantifies their relationship when $\pi_0$ is a uniform distribution. Define event sets $\Omega_1 = \{(y_1, y_2, x) \mid |\log\frac{p^*(1 \mid y_1, y_2, x)}{p^*(0 \mid y_1, y_2, x)}| \geq \epsilon_0\}$ and $\Omega_2 = \{(y_1, y_2, x) \mid |\log\frac{\pi_\theta(y_1 \mid x)\pi_{\text{ref}}(y_2 \mid x)}{\pi_\theta(y_2 \mid x)\pi_{\text{ref}}(y_1 \mid x)}| \geq \epsilon_0\}$. Let $\gamma(x) = \frac{|\Omega_1 \cap \Omega_2|}{K^2}\leq\gamma$, where $K=|\mathbb{Y}|$. $\epsilon_0$ represents the log-probability significance threshold of $p^*$. Intuitively, the parameter $\gamma$ reflects the alignment between $\pi_0$ and $\pi_\theta$ for large-margin data pairs $(x, y_1, y_2)$. 

\begin{lemma}\label{data_select}
Assume $\|g_t\|^2\leq G^2$. See $\mathcal{L}_{\mathrm{DPO}}(\pi_{\theta} ; \pi_{\text {ref }})$ on Eq.\ref{DPO_eq}. Given the learning rate $\alpha_t$ satisfying finite squared summability, for DPO, let $\gamma(x)\leq\gamma$, when $\pi_0$ is a uniform distribution, then:
\begin{equation}\label{data_select_eq}
\small{    \begin{aligned}
\min_{1\leq i\leq T}&||\nabla_\theta \mathcal{L}_{\mathrm{DPO}}(\pi_{\theta_i} ; \pi_{\text {ref }})||^2_2\leq \frac{\mathcal{L}_{\mathrm{DPO}}(\pi_{\theta_1})-\mathcal{L}_{\mathrm{DPO}}^*}{\sum_{t=1}^{T-1}\alpha_t} + \frac{2(\gamma c_0+1)G^2\sum_{t=1}^{T-1}\alpha_t^2}{\tau^2\sum_{t=1}^{T-1}\alpha_t}.
    \end{aligned}}
\end{equation} 
where $\mathcal{L}_{\mathrm{DPO}}^*=\min_{\pi_\theta} \mathcal{L}_{\mathrm{DPO}}(\pi_{\theta} ; \pi_{\text {ref }})$ and $c_0=\sigma(\frac{\epsilon_0}{\tau})\sigma(-\frac{\epsilon_0}{\tau})-1\in (-1,0)$. 
\end{lemma}

Assuming $\pi_0$ is uniform, Lemma \ref{data_select} prepares for analyzing how sampling certain types of data affects convergence. Since $c_0$ is negative, a higher $\gamma$ value indicates better alignment and faster convergence, as shown in the second term of Eq.\ref{data_select_eq}.


Based on Lemma \ref{data_select}, we further demonstrate with Theorem \ref{data_select_coro} that when the data pairs $(x, y_1, y_2)$ satisfy that $(x, y_1, y_2)\in\Omega_1 \cap \Omega_2$ and are sampled more, the convergence rate can be accelerated. In Theorem \ref{data_select_coro}, $\mu$ represents a preference for selecting data pairs with large margins that are also consistent with $\pi_\theta$. Compared to the upper bound of inequality \ref{data_select_eq}, the upper bound of inequality \ref{data_select_coro_eq} is smaller, indicating that the data selection distribution $\pi_1$ plays a beneficial role in accelerating the DPO optimization process. Thus in practice, we can sample data according to $\pi_1$ to achieve faster convergence for DPO. See proof for Theorem \ref{data_select_coro} in Appendix \ref{data_select_coro_proof}.
\begin{theorem}\label{data_select_coro}
Define joint conditional probability distribution $\pi_1(y_1,y_2|x)= \frac{\mu}{K^2}\ \text{if}\ (x,y_1,y_2)\in(\Omega_1 \cap \Omega_2); \frac{1-\mu\gamma}{(1-\gamma)K^2} \text{else}$ where $\mu\in(0,1)$. Assume $\|g_t\|^2\leq G^2$. Given the learning rate $\alpha_t$  satisfying finite squared summability, for DPO, we have:
\begin{equation}\label{data_select_coro_eq}
\small{    \begin{aligned}
\min_{1\leq i\leq T}&||\nabla_\theta \mathcal{L}_{\mathrm{DPO}}(\pi_{\theta_i} ; \pi_{\text {ref }})||^2_2\leq \frac{\mathcal{L}_{\mathrm{DPO}}(\pi_{\theta_1})-\mathcal{L}_{\mathrm{DPO}}^*}{\sum_{t=1}^{T-1}\alpha_t} + \frac{2(\mu\gamma c_0+1)G^2\sum_{t=1}^{T-1}\alpha_t^2}{\tau^2\sum_{t=1}^{T-1}\alpha_t}.
    \end{aligned}}
\end{equation}
\end{theorem}


In summary, we conduct the answers for Q1 and Q2:
\begin{itemize}
\item About Q1: 
A larger value of $ \tau $ leads to faster convergence of DPO. As $ \tau $ increases, the optimal solution $ \bar{\pi}^\tau $ of DPO converges to $ \pi^\delta$, while deviating from $ \pi_{\text{ref}} $. Hence, there is a trade-off between the convergence speed and the preservation of proximity to $\pi_{\text{ref}}$.
\item About Q2: Selecting data where both the current policy probability values ($\pi(y_1|x)$ and $\pi(y_2|x)$) and the reward values ($r(x,y_1)$ and $r(x,y_1)$) consistently exhibit a large distance (i.e. large margin) can accelerate the convergence of the DPO algorithm. And with the consistence, the distribution shift problem will reduce.
\end{itemize}

\section{Conclusion}\label{conclusion_sec}
In this paper, we investigated the connections of DPO, RL and other RLHF algorithms. We proposed UDRRA to link these algorithms through their different reward function scenarios. Then we analyzed the target distributions corresponding to the methods in UDRRA and pointed out that the distribution shift problem of DPO comparing PRA-P. Furthermore, we examined the impact of key components of the DPO loss on algorithm performance. Our findings offer a deeper understanding of DPO’s theoretical positioning and practical implications, providing a foundation for future developments in RLHF algorithms.


\bibliographystyle{unsrt}  
\bibliography{references}

\newpage
\appendix
\section*{Appendix Contents}
\begin{itemize}
\item Appendix \ref{appendix_related_work}: Related Work.
\item Appendix \ref{omega}: Discussion about $p^*(z \mid y_1, y_2, x)=\omega(r(x,y_{2-z}),r(x,y_{z+1}))$.
\item Appendix \ref{Proof_Preliminary}: Proof in Section \ref{Preliminary}.
\item \quad Appendix \ref{tau_tend_to_delta_proof}: Proof of Proposition \ref{tau_tend_to_delta}.
\item \quad Appendix \ref{Softmax_Transform_sec}: {The Softmax Transform of Policy $\pi_\theta$}.
\item Appendix \ref{TRPRA_sec_proof}: Proof in Section \ref{TRPRA_sec}.
\item \quad Appendix \ref{forward-KL_proof}: Derivation of Equation \ref{forward-KL}.
\item \quad Appendix \ref{reverse-KL_proof}: Derivation of Equation \ref{reverse-KL}.
\item \quad Appendix \ref{Derivation_PRA}: Derivation of Equation \ref{PRA_eq}.
\item \quad Appendix \ref{Derivation_RA_eq}: Derivation of Equation \ref{RA_eq}.
\item \quad Appendix \ref{DPO_eq_from_PRA_proof}: Derivation of Equation \ref{DPO_eq}.
\item Appendix \ref{RL_Classification_sec_proof}: Proof in Section \ref{RL_Classification_sec}.
\item \quad Appendix \ref{Proof_Reward_policy}: The Target Distribution Equivalence of Eq.\ref{forward-KL}, Eq.\ref{reverse-KL} and Eq.\ref{RA_eq}.
\item \quad Appendix \ref{Proof_RDA_lemma}: The Target Distribution Equivalence of Eq.\ref{RA_eq} and Eq.\ref{pair-wised_RA_eq}.
\item \quad Appendix \ref{PRA_optimal_proof}: The Target Distribution Equivalence of Eq.\ref{pair-wised_RA_eq} and Eq.\ref{PRA_eq}.
\item \quad Appendix \ref{Reward_policy_posterior_proof}: Proof of The Target Distribution Equivalence of Eq.\ref{RA_eq_posterior} and Eq.\ref{PRA_posterior}.
\item \quad Appendix \ref{DPO_PRA_proof}: Proof of Theorem \ref{DPO_PRA}.
\item Appendix \ref{Convergence_sec_proof}: Proof in Section \ref{Convergence_sec}.
\item \quad Appendix \ref{SGD_theorem_proof}: Proof of Proposition \ref{SGD_theorem}.
\item \quad Appendix \ref{DPO_theorem_proof}: Proof of Theorem \ref{DPO_theorem}.
\item \quad Appendix \ref{data_select_proof}: Proof of Lemma \ref{data_select}.
\item \quad Appendix \ref{data_select_coro_proof}: {Proof of Theorem \ref{data_select_coro}}.
\item \quad Appendix \ref{pq_equal_sec}: {Proposition \ref{pq_equal}}.
\item \quad Appendix \ref{SubLemma_sec}: {Sub Lemma for Lemma \ref{BDA_L_RKL_coef}-\ref{PRA_L_coef}}.
\item \quad Appendix \ref{BDA_L_RKL_coef_proof}: Proof of Lemma \ref{BDA_L_RKL_coef}.
\item \quad Appendix \ref{BDA_L_FKL_coef_proof}: {Proof of Lemma \ref{BDA_L_FKL_coef}}.
\item \quad Appendix \ref{RA_L_coef_proof}: {Proof of Lemma \ref{RA_L_coef}}.
\item \quad Appendix \ref{RDA_L_coef_proof}: {Proof of Lemma \ref{RDA_L_coef}}.
\item \quad Appendix \ref{PRA_L_coef_proof}: {Proof of Lemma \ref{PRA_L_coef}}.
\end{itemize}
\newpage

\section{Related Work}\label{appendix_related_work}
\textbf{Reinforcement Learning from Human Feedback (RLHF).} The Large Language Models (LLMs) \cite{zhao2023survey,chang2024survey,hadi2024large,minaee2024large,hadi2023survey,achiam2023gpt,bubeck2023sparks} is one of the most promising evolutions towards Artificial General Intelligence (AGI). The success of this transformation lies a critical component: Reinforcement Learning from Human Feedback (RLHF) or Human Alignment (HA), which is the final and crucial step in LLMs' training \cite{arumugam2019deep,ouyang2022training,singh2022flava,bai2022training,dai2023safe}. Motivated by the instability, complexity, and incurring significant computational costs of the RLHF process, \cite{DPO} proposed an algorithmic framework for directly optimizing the language model to follow human preferences, namely the DPO algorithm, based on the analysis of the optimal solutions of classic RL algorithm PPO \cite{schulman2017proximal,engstrom2019implementation}. DPO has received widespread attention since its inception, and hundreds of algorithms for improving DPO have been derived within a year, e.g. IPO\cite{IPO}, KTO\cite{KTO}, SimPO\cite{SimPO}, ORPO\cite{ORPO}, etc. 

\textbf{DPO $\&$ PPO discussing.}
The success of DPO \cite{DPO} benefits from the optimal solution analysis of the KL-constrained reward maximization objective (Eq.3 in \cite{DPO}) come from the classical Reinforcement Learning (RL) algorithm, PPO \cite{PPO}. Therefore, the performance difference between the DPO algorithm and the PPO based RLHF method \cite{PPO-basedRLHF} has gradually attracted the attention of researchers in the RL community \cite{ivison2024unpacking}. In general, while the DPO algorithm can fit the static training dataset comparably, it generalizes less effectively than PPO based RLHF \cite{lin2024limited}. While PPO based RLHF usually performs better in the state-of-the-art production-level LLMs \cite{yan20243d}, its correct fine-tuning usually requires more sophisticated techniques \cite{xu2024dpo}.

There are many works that empirically discuss the relationship between the DPO algorithm and the PPO algorithm in RLHF. From the perspective of algorithm design, \cite{xu2024dpo,ivison2024unpacking,yan20243d} pointed out that the PPO algorithm is generally better than the DPO algorithm in human preference alignment tasks, but the PPO algorithm requires various additional tricks, such as advantage normalization, large batch size, and exponential moving average update for the reference model. \cite{zhong2024dpo} combined the advantages of DPO and PPO to create a more effective algorithm named RTO at the token-level. From the perspective of data source, \cite{tang2024understanding} classified algorithms such as DPO as offline algorithms, while the PPO based RLHF is an online algorithm. This demonstration is consistent with our paper. \cite{li2023policy,lin2024limited} showed that compared with the DPO algorithm that only relies on static data sets, the PPO algorithm can use sufficient non-preferred data for policy optimization to significantly improve performance by relying on the generalization ability of its Reward model.

\section{Discussion about $p^*(z \mid y_1, y_2, x)=\omega(r(x,y_{2-z}),r(x,y_{z+1}))$}\label{omega}
As for modeling the comparison probability $p^*(z \mid y_1, y_2, x)$ between $r(x,y_1)$ and $r(x,y_2)$, i.e. the pair $(x, y_1, y_2, z)$ is sampled with probability $p^*(z \mid y_1, y_2, x)=\omega(r(x,y_{2-z}),r(x,y_{z+1}))$ where $z\in\{0,1\}$, in general, the difficulty of estimating the reward depends on the complexity of $\omega^{-1}$. Table \ref{omega_table} presents several special forms of the $\omega$ function, including cases based on the BT model assumptions \cite{bradley1952rank}, along with an analysis of their relevant properties.
\begin{table}[ht] 
\centering
\begin{threeparttable} 
   \resizebox{1\hsize}{!}{
   \begin{tabular}{cccc}
   \toprule
    $\omega(r(x,y_{2-z}),r(x,y_{z+1}))$ & \#Requirement about $r(x,y)$ & \#Inverse Formula about $r(x,y_{1})-r(x,y_{2})$ & \#Corresponding Algorithm \\
   \midrule
    \tnote{1}$\quad \frac{\exp(\eta r(x,y_{2-z}))}{\exp(\eta r(x,y_{2-z}))+\exp(\eta r(x,y_{z+1}))}$  & $[-\infty,\infty]$ & $\frac{1}{\eta}\log(\frac{p^*(1 \mid y_1, y_2, x)}{1-p^*(1 \mid y_1, y_2, x)})$ & DPO\cite{DPO} \\
    $\frac{(\eta r(x,y_{2-z}))}{(\eta r(x,y_{2-z}))+(\eta r(x,y_{z+1}))}$  & $[0,\infty]$ & N/A  & TYPO\cite{TYPO}  \\
    \tnote{2}$\quad \frac{1}{2}+\frac{1}{2}\tanh(r(x,y_{2-z})-r(x,y_{z+1}))$  & $[-\infty,\infty]$ & $\frac{1}{2}\log(\frac{p^*(1 \mid y_1, y_2, x)}{1-p^*(1 \mid y_1, y_2, x)})$ & N/A \\
    $\frac{1}{2}+\frac{1}{2}\sin(r(x,y_{2-z})-r(x,y_{z+1}))$  & $[-\frac{\pi}{2},\frac{\pi}{2}]$ & $\arcsin(2p^*(1 \mid y_1, y_2, x)-1)$ & N/A \\
    $\mathbb{I}(r(x,y_{2-z})-r(x,y_{z+1}))$ & $[-\infty,\infty]$ & $ 1 $ & IPO\cite{IPO} \\
    \tnote{4}$\quad \max\{0,1-\eta(r(x,y_{2-z})-r(x,y_{z+1}))\}$ & $[-\infty,\infty]$ & N/A & SLiC\cite{SLiC} \\
    \tnote{3}$\ $\tnote{4}$\quad \frac{\exp(\eta r(x,y_{2-z}))}{\exp(\eta r(x,y_{2-z}))+\exp(\eta r(x,y_{\text{average}}))}$ & $[-\infty,\infty]$ & $\frac{1}{\eta}\log(\frac{p^*(1 \mid y_1, y_2, x)(1-p^*(0 \mid y_1, y_2, x))}{p^*(0 \mid y_1, y_2, x)(1-p^*(1 \mid y_1, y_2, x))})$ & KTO\cite{KTO} \\
    \tnote{4}$\quad \frac{1}{(1+\exp(\eta(r(x,y_{2-z})-r(x,y_{z+1}))))^2}$  & $[-\infty,\infty]$ & $\frac{1}{\eta}\log(\frac{\sqrt{p^*(1 \mid y_1, y_2, x)}}{1-\sqrt{p^*(1 \mid y_1, y_2, x)}})$  & N/A  \\
    \tnote{4}$\quad \exp(\eta(r(x,y_{2-z})-r(x,y_{z+1})))$  & $[-\infty,\infty]$ & $\frac{1}{\eta}\log(p^*(1 \mid y_1, y_2, x))$  & N/A  \\
   \bottomrule
  \end{tabular}
  }
  \begin{tablenotes}
  \footnotesize  
     \item[1] Here is the generalized BT model expression, like other work did, we simplified $\eta=1$ in Eq.\ref{BTmodel}.
     \item[2] $\frac{1}{2}+\frac{1}{2}\tanh(r(x,y_{1})-r(x,y_{2}))$ actually is a BT model expression when $\eta=2$.
     \item[3] $r(x,y_{\text{average}})$ is the reference point $z_{\text{ref}}$ in KTO\cite{KTO}. 
     \item[4] These forms of $\omega$ may not strictly satisfy $p^*(1 \mid y_1, y_2, x)=(1-p^*(0 \mid y_1, y_2, x))$.
   \end{tablenotes}
   \end{threeparttable}       
   \caption{Various expressions of function $\omega(r(x,y_{2-z}),r(x,y_{z+1}))$.}
  \label{omega_table}
\end{table}

\section{Proof in Section \ref{Preliminary}}\label{Proof_Preliminary}

\subsection{Proof of Proposition \ref{tau_tend_to_delta}}\label{tau_tend_to_delta_proof}
\textbf{Proposition \ref{tau_tend_to_delta}:} Without loss of generality, assume that the reward function $r(x, y)$ has a unique maximum for any given $x$. Define
$$\small \pi^\delta(y|x) = 
\begin{cases} 
1, & \text{if } y = \arg\max_{y^{\prime}} r(x, y^{\prime}), \\
0, & \text{otherwise}.
\end{cases} \normalsize$$
Then the following limit holds:
$  \lim_{\tau \to \infty} \pi^\tau(y|x) = \pi^\delta(y|x).$  

\textbf{Proof:} Given $x$ and defining $y^* = \arg\max\ r(x,y)$, we consider the limit of $\pi^\tau(y|x)$ at the point $(x, y^*)$ as $\tau$ approaches infinity:
\begin{equation}
    \small{\begin{aligned}
&\lim_{\tau \to \infty} \pi^\tau(y^*|x) = \lim_{\tau \to \infty} \frac{\exp(\tau r(x, y^*))}{\sum_{y \in Y} \exp(\tau r(x, y))} = \frac{1}{\lim_{\tau \to \infty} \sum_{y \in Y} \exp(\tau (r(x, y) - r(x, y^*)))} \\
&= \frac{1}{\sum_{y \in Y} \lim_{\tau \to \infty} \exp(\tau (r(x, y) - r(x, y^*)))} = \frac{1}{1 + 0 + \cdots + 0} = 1.
\end{aligned}}
\end{equation}

Next, for $y \neq y^*$, we consider the limit of $\pi^\tau(y|x)$ at the point $(x, y)$:
\begin{equation}
    \small{\begin{aligned}
&\lim_{\tau \to \infty} \pi^\tau(y|x) = \lim_{\tau \to \infty} \frac{\exp(\tau r(x, y))}{\sum_{y' \in Y} \exp(\tau r(x, y'))} = \frac{1}{\lim_{\tau \to \infty} \sum_{y' \in Y} \exp(\tau (r(x, y') - r(x, y)))} \\
&= \frac{1}{\sum_{y' \in Y} \lim_{\tau \to \infty} \exp(\tau (r(x, y') - r(x, y)))} \\
&= \frac{1}{\sum_{y' \in Y, \ r(x, y') > r(x, y)} \lim_{\tau \to \infty} \exp(\tau (r(x, y') - r(x, y))) + \sum_{y' \in Y, \ r(x, y') \leq r(x, y)} \lim_{\tau \to \infty} \exp(\tau (r(x, y') - r(x, y)))} \\
&= \frac{1}{\sum_{y' \in Y, \ r(x, y') > r(x, y)} \lim_{\tau \to \infty} \exp(\tau (r(x, y') - r(x, y)))} = 0.
\end{aligned}}
\end{equation}

Thus, we conclude that $\lim_{\tau \to \infty} \pi^\tau(y|x) = \pi^\delta(y|x)$. Proof finished.

\subsection{The Softmax Transform of Policy $\pi_\theta$}\label{Softmax_Transform_sec}
Given a policy $\pi: \mathbb{X} \rightarrow \Delta(\mathbb{Y})$, the softmax transform of a vector exponentiates the components of the vector and normalizes it so that the result lies in the simplex. This can be used to transform vectors assigned to state-action pairs into policies:
\begin{definition}
\label{Softmax}
    (Softmax transform.) Given the function $\theta: \mathbb{X} \times \mathbb{Y} \rightarrow$ $\mathbb{R}$, the softmax transform of $\theta$ is defined as $\pi_\theta(\cdot \mid x)\triangleq \operatorname{softmax}(\theta(x, \cdot))$, where for all $y \in \mathbb{Y}$,
\begin{equation}
    \pi_\theta(y \mid x)\triangleq \frac{\exp \{\theta(x, y)\}}{\sum_{y^{\prime}} \exp \left\{\theta\left(x, y^{\prime}\right)\right\}}.
\end{equation}
\end{definition}

We call the values $\theta(x, y)$ the logit values and the function $\theta$ itself a logit function due to its origin in logistic regression. This paper assume the set is parameterized by a softmax function, i.e., $\Pi = \{\pi_\theta \mid \pi_\theta(\cdot \mid x) \triangleq  \operatorname{softmax}(\theta(x, \cdot)), \, \theta : \mathbb{X} \times \mathbb{Y} \to \mathbb{R}\}$, which is commonly used in the proof of the policy gradient theorem \cite{mei2020global}.

\section{Proof in Section \ref{TRPRA_sec}}\label{TRPRA_sec_proof}
\subsection{Derivation of Equation \ref{forward-KL}}\label{forward-KL_proof}
\begin{equation}
    \begin{aligned}  
&\mathbb{E}_{x\sim \mathcal{D}}\left[\mathrm{D}_{\mathrm{KL}}\left(\pi_\theta\left(\cdot \mid x\right) \| \pi^\tau\left(\cdot \mid x\right)\right)\right]=\mathbb{E}_{x\sim \mathcal{D}}[\mathrm{D}_{\mathrm{KL}}(\pi_\theta(y|x)||\frac{\exp(\tau r(x,y))}{Z(x)})] \\ 
=& \mathbb{E}_{x\sim \mathcal{D}}\left[\sum_{y\in Y} \pi_\theta(y|x) \log\left(\frac{\pi_\theta(y|x)}{\frac{\exp(\tau r(x,y))}{Z(x)}}\right)\right] = \mathbb{E}_{x\sim \mathcal{D}}\left[\sum_{y\in Y} \pi_\theta(y|x) \log\left(\frac{Z(x)\pi_\theta(y|x)}{\exp(\tau r(x,y))}\right)\right] \\   
=& \mathbb{E}_{x\sim \mathcal{D}}\left[\sum_{y\in Y} -\pi_\theta(y|x)(\tau r(x,y) - \log(\pi_\theta(y|x))) + \log(Z(x))\right] \\  
=& \mathbb{E}_{x\sim \mathcal{D}}\left[\sum_{y\in Y} -\pi_\theta(y|x)(\tau r(x,y) - \log(\pi_\theta(y|x)))\right] + \mathbb{E}_{x\sim \mathcal{D}}[\log(Z(x))] \\
\equiv&\mathbb{E}_{x\sim \mathcal{D}}\left[\sum_{y\in Y} -\pi_\theta(y|x)(\tau r(x,y) - \log(\pi_\theta(y|x)))\right]\\
=&\mathbb{E}_{x\sim \mathcal{D},y\sim\pi_\theta(y|x)}\left[-(\tau r(x,y) - \log(\pi_\theta(y|x)))\right].
\end{aligned}
\end{equation}
The ``$\equiv$" means that the left and right ends differ by a term that is unrelated to $\theta$.

\subsection{Derivation of Equation \ref{reverse-KL}}\label{reverse-KL_proof}
\begin{equation}
    \begin{aligned}
&\mathbb{E}_{x\sim \mathcal{D}}\left[\mathrm{D}_{\mathrm{KL}}\left( \pi^\tau\left(\cdot \mid x\right) \| \pi_\theta\left(\cdot \mid x\right) \right)\right] \equiv\mathbb{E}_{x\sim \mathcal{D}}[\sum_{y\in Y}\frac{\exp(\tau r(x,y))}{Z(x)}(-\log\pi_\theta(y|x))]\\
=&-\mathbb{E}_{x\sim \mathcal{D},y\sim\pi^\tau\left(\cdot \mid x\right)}[\log\pi_\theta(y|x)]=\mathbb{E}_{x\sim \mathcal{D}}[\sum_{y\in Y}\pi_\theta(y|x)\frac{1}{\pi_\theta(y|x)}\frac{\exp(\tau r(x,y))}{Z(x)}(-\log\pi_\theta(y|x))]\\
=&-\mathbb{E}_{x\sim \mathcal{D},y\sim\pi_\theta(\cdot|x)}[\frac{\exp(\tau r(x,y))}{Z(x)\pi_\theta(y|x)}\log\pi_\theta(y|x)].
\end{aligned}
\end{equation}

\subsection{Derivation of Equation \ref{PRA_eq}}\label{Derivation_PRA}
\begin{equation}
    \begin{aligned}
&\mathcal{L}_{\mathrm{PRA}}(\pi_\theta)=\mathbb{E}_{\mathcal{D}_{\theta}}\left[\mathrm{D}_{\mathrm{KL}}(p^*(z|y_1,y_2,x)||p_\theta(z|y_1,y_2,x))\right]\\
=&-\mathbb{E}_{x\sim \mathcal{D},y_1,y_2\sim \pi_\theta(y|x)}\left[p^*(1|y_1,y_2,x)\log p_\theta\left( 1|y_1,y_2,x\right)+p^*(0|y_1,y_2,x)\log p_\theta\left( 0|y_1,y_2,x\right)\right]\\
&+\mathbb{E}_{x\sim \mathcal{D},y_1,y_2\sim \pi_\theta(y|x)}\left[p^*(1|y_1,y_2,x)\log p^*(1|y_1,y_2,x)+p^*(0|y_1,y_2,x)\log p^*(0|y_1,y_2,x)\right]\\
=&-\mathbb{E}_{x\sim \mathcal{D},y_1,y_2\sim \pi_\theta(y|x)}\left[p^*(1|y_1,y_2,x)\log p_\theta\left( 1|y_1,y_2,x\right)+p^*(0|y_1,y_2,x)\log p_\theta\left( 0|y_1,y_2,x\right)\right]\\
&+\mathbb{E}_{x\sim \mathcal{D},y_1,y_2\sim \pi_\theta(y|x)}\left[M(x,y_1,y_2)\right]\\
=&-\mathbb{E}_{x\sim \mathcal{D},y_1,y_2\sim \pi_\theta(y|x)}\left[p^*(1|y_1,y_2,x)\log p_\theta\left( 1|y_1,y_2,x\right)+p^*(1|y_2,y_1,x)\log p_\theta\left( 1|y_2,y_1,x\right)\right]\\
&+\mathbb{E}_{x\sim \mathcal{D},y_1,y_2\sim \pi_\theta(y|x)}\left[M(x,y_1,y_2)\right]\\
=&-\mathbb{E}_{x\sim \mathcal{D},y_1,y_2\sim \pi_\theta(y|x)}\left[p^*(1|y_w,y_l,x)\log p_\theta\left( 1|y_w,y_l,x\right)+p^*(1|y_l,y_w,x)\log p_\theta\left( 1|y_l,y_w,x\right)\right]\\
&+\mathbb{E}_{x\sim \mathcal{D},y_1,y_2\sim \pi_\theta(y|x)}\left[M(x,y_1,y_2)\right]\\
=&-\mathbb{E}_{(x,y_w,y_l)\sim \mathcal{D}_{\theta}}\left[\log p_\theta\left( 1|y_w,y_l,x\right)\right]+\mathbb{E}_{x\sim \mathcal{D},y_1,y_2\sim \pi_\theta(y|x)}\left[M(x,y_1,y_2)\right].
\end{aligned}
\end{equation}
where $\sigma(x)=\frac{1}{1+e^{-x}}$ is the sigmoid function, $M(x,y_1,y_2)=\sum_{z=0,1}p^*(z|y_1,y_2,x)\log p^*(z|y_1,y_2,x)$ and $(x,y_1,y_2)\sim \mathcal{D}_{\theta}\triangleq y_1\succ y_2\sim p^*(z=1|y_1,y_2,x),y_1,y_2\sim\pi_\theta(y|x),x\sim\mathcal{D}$. The third equality is because the symmetric complementarity property of $\omega(\cdot,\cdot)$:
\begin{equation}
p^*(1|y_2,y_1,x)=\omega(r(x,y_2),r(x,y_1))=1-\omega(r(x,y_1),r(x,y_2))=1-p^*(1|y_1,y_2,x)=p^*(0|y_1,y_2,x).
\end{equation}

\subsection{Derivation of Equation \ref{RA_eq}}\label{Derivation_RA_eq}
\begin{equation}
\begin{aligned}
&\mathbb{E}_{x\sim \mathcal{D},y\sim\pi_\theta(\cdot|x)}\left[\left(r_\theta(x,y)- r(x,y)\right)^2\right]=\mathbb{E}_{x\sim \mathcal{D},y\sim\pi_\theta(y|x)}\left(\frac{1}{\tau}\log(\pi_\theta(y|x)) + \frac{1}{\tau}\log(Z(x))-r(x,y) \right)^2\\
=&\frac{1}{\tau^2}\mathbb{E}_{x\sim \mathcal{D},y\sim\pi_\theta(\cdot|x)}\left[\left(\log(\pi_\theta(y|x)) - \log(\frac{\exp(\tau r(x,y))}{Z(x)})\right)^2\right]=\frac{1}{\tau^2}\mathbb{E}_{x\sim \mathcal{D},y\sim\pi_\theta(\cdot|x)}\left[\left(\log(\frac{\pi_\theta(y|x)}{\pi^\tau(y|x)})\right)^2\right].
\end{aligned}
\end{equation}

\subsection{Derivation of Equation \ref{DPO_eq}}\label{DPO_eq_from_PRA_proof}
\begin{equation}
    \begin{aligned}
&\mathbb{E}_{x\sim \mathcal{D},y_1,y_2\sim \pi_0(y|x)}\left[\mathbb{D}_{\mathrm{KL}}(p^*(z|y_1,y_2,x)||\bar{p}_\theta(z|y_1,y_2,x))\right]\\
\equiv&-\mathbb{E}_{x\sim \mathcal{D},y_1,y_2\sim \pi_0(y|x)}\left[p^*(1|y_1,y_2,x)\log\sigma\left( \bar{h}_\theta(x,y_1,y_2)\right)+p^*(0|y_1,y_2,x)\log\sigma\left( \bar{h}_\theta(x,y_2,y_1)\right)\right]\\
=&-\mathbb{E}_{x\sim \mathcal{D},y_1,y_2\sim \pi_0(y|x)}\left[p^*(1|y_1,y_2,x)\log\sigma\left( \bar{h}_\theta(x,y_1,y_2)\right)+p^*(1|y_2,y_1,x)\log\sigma\left( \bar{h}_\theta(x,y_2,y_1)\right)\right]\\
=&-\mathbb{E}_{x\sim \mathcal{D},y_1,y_2\sim \pi_0(y|x)}\left[p^*(1|y_w,y_l,x)\log\sigma\left( \bar{h}_\theta(x,y_w,y_l)\right)+p^*(1|y_l,y_w,x)\log\sigma\left( \bar{h}_\theta(x,y_l,y_w)\right)\right]\\
=&-\mathbb{E}_{(x,y_w,y_l)\sim \mathcal{D}_R}\left[\log\sigma\left( \bar{h}_\theta(x,y_w,y_l)\right)\right]\\
=&-\mathbb{E}_{(x,y_w,y_l)\sim \mathcal{D}_R}\left[\log\sigma\left( \frac{1}{\tau}\log\frac{\pi_\theta(y_w|x)}{\pi_{ref}(y_w|x)}-\frac{1}{\tau}\log\frac{\pi_\theta(y_l|x)}{\pi_{ref}(y_l|x)}\right)\right].
\end{aligned}
\end{equation}
where $z\in\{0,1\}$, $\bar{p}_\theta(0|y_1,y_2,x)=\sigma\left( \bar{h}_\theta(x,y_2,y_1)\right)$ and $\bar{p}_\theta(1|y_1,y_2,x)=\sigma\left( \bar{h}_\theta(x,y_1,y_2)\right)$.

\section{Proof in Section \ref{RL_Classification_sec}}\label{RL_Classification_sec_proof}


\subsection{The Target Distribution Equivalence of Eq.\ref{forward-KL}, Eq.\ref{reverse-KL} and Eq.\ref{RA_eq}}\label{Proof_Reward_policy}
Denote:
\begin{equation}
\begin{aligned}
\pi^*_{\mathrm{Forward-KL}}&=\arg\min_{\pi_\theta} -\mathbb{E}_{x\sim \mathcal{D},y\sim\pi_\theta(\cdot|x)}\left[(\tau r(x,y) - \log(\pi_\theta(y|x)))\right], \\
\pi^*_{\mathrm{Reverse-KL}}&=\arg\min_{\pi_\theta} -\mathbb{E}_{x\sim \mathcal{D},y\sim\pi^\tau\left(\cdot \mid x\right)}[\log\pi_\theta(y|x)], \\
\pi^*_{\mathrm{RA}}&=\arg\min_{\pi_\theta}\mathbb{E}_{x\sim \mathcal{D},y\sim\pi_\theta(\cdot|x)}\left[\frac{1}{\tau^2}\left(\log(\frac{\pi_\theta(y|x)}{\pi^\tau(y|x)})\right)^2\right],
\end{aligned}
\end{equation}
then $\pi^*_{\mathrm{Forward-KL}}=\pi^*_{\mathrm{Reverse-KL}}=\pi^*_{\mathrm{RA}}=\pi^\tau$.

\textbf{Proof:} First, we prove $\pi^*_{\mathrm{Forward-KL}}=\pi^\tau$:
\begin{equation}
    \begin{aligned}
&-\mathbb{E}_{x\sim \mathcal{D},y\sim\pi_\theta(\cdot|x)}\left[\tau r(x,y) - \log(\pi_\theta(y|x))\right]-(-\mathbb{E}_{x\sim \mathcal{D},y\sim\pi^\tau(\cdot|x)}\left[\tau r(x,y) - \log(\pi^\tau(y|x))\right] )\\
=& -\mathbb{E}_{x\sim \mathcal{D},y\sim\pi_\theta(\cdot|x)}\left[\tau r(x,y) - \log(\pi_\theta(y|x))\right]+\mathbb{E}_{x\sim \mathcal{D},y\sim\pi^\tau(\cdot|x)}\left[\tau r(x,y) - \log(\frac{\exp \left(\tau r\left(x, y\right)\right)}{Z(x)} )\right] \\
=& -\mathbb{E}_{x\sim \mathcal{D},y\sim\pi_\theta(\cdot|x)}\left[\tau r(x,y) - \log(\pi_\theta(y|x))\right]+\mathbb{E}_{x\sim \mathcal{D},y\sim\pi^\tau(\cdot|x)}\left[\log({Z(x)} )\right] \\
=& \mathbb{E}_{x\sim \mathcal{D},y\sim\pi_\theta(\cdot|x)}\left[-\tau r(x,y) + \log(\pi_\theta(y|x)) + \frac{\pi^\tau(y|x)}{\pi_\theta(y|x)}\log({Z(x)} ) \right] \\
=& \mathbb{E}_{x\sim \mathcal{D},y\sim\pi_\theta(\cdot|x)}\left[-(\log\exp(\tau r(x,y)) -\log({Z(x)} )+\log({Z(x)})) + \log(\pi_\theta(y|x)) + \frac{\pi^\tau(y|x)}{\pi_\theta(y|x)}\log({Z(x)} ) \right] \\
=& \mathbb{E}_{x\sim \mathcal{D},y\sim\pi_\theta(\cdot|x)}\left[-\log(\pi^\tau(y|x))-\log({Z(x)}) + \log(\pi_\theta(y|x)) + \frac{\pi^\tau(y|x)}{\pi_\theta(y|x)}\log({Z(x)} ) \right] \\
=& \mathbb{E}_{x\sim \mathcal{D},y\sim\pi_\theta(\cdot|x)}\left[\log(\frac{\pi_\theta(y|x)}{\pi^\tau(y|x)}) + \frac{\pi^\tau(y|x)-\pi_\theta(y|x)}{\pi_\theta(y|x)}\log({Z(x)} ) \right] \\
=& \mathbb{E}_{x\sim \mathcal{D}}\left[\mathrm{D}_{\mathrm{KL}}\left(\pi_\theta\left(\cdot \mid x\right) \| \pi^\tau\left(\cdot \mid x\right)\right)\right]\geq0.\\
    \end{aligned}
\end{equation}
From the above equation, we can see that when $\pi_\theta$ is equal to $\pi^\tau$, $-\mathbb{E}_{x\sim \mathcal{D},y\sim\pi_\theta(\cdot|x)}\left[\tau r(x,y) - \log(\pi_\theta(y|x))\right]$ reaches the minimum value $-\mathbb{E}_{x\sim \mathcal{D},y\sim\pi^\tau(\cdot|x)}\left[\tau r(x,y) - \log(\pi^\tau(y|x))\right]$. Therefore $\pi^*_{\mathrm{Forward-KL}}=\pi^\tau$.

Second, similarly we can prove $\pi^*_{\mathrm{Reverse-KL}}=\pi^\tau$:
\begin{equation}
    \begin{aligned}
&-\mathbb{E}_{x\sim \mathcal{D},y\sim\pi^\tau\left(\cdot \mid x\right)}[\log\pi_\theta(y|x)]-(-\mathbb{E}_{x\sim \mathcal{D},y\sim\pi^\tau\left(\cdot \mid x\right)}[\log\pi^\tau(y|x)] )=\mathbb{E}_{x\sim \mathcal{D}}\left[\mathrm{D}_{\mathrm{KL}}\left(\pi^\tau\left(\cdot \mid x\right) \| \pi_\theta\left(\cdot \mid x\right)\right)\right] \geq0.\\
    \end{aligned}
\end{equation}

At last, we prove $\pi^*_{\mathrm{RA}}=\pi^\tau$ like above:
\begin{equation}
\small{    \begin{aligned}
\mathbb{E}_{x\sim \mathcal{D},y\sim\pi_\theta(\cdot|x)}\left[\frac{1}{\tau^2}\left(\log(\frac{\pi_\theta(y|x)}{\pi^\tau(y|x)})\right)^2\right]-\mathbb{E}_{x\sim \mathcal{D},y\sim\pi_\theta(\cdot|x)}\left[\frac{1}{\tau^2}\left(\log(\frac{\pi^\tau(y|x)}{\pi^\tau(y|x)})\right)^2\right]=\mathbb{E}_{x\sim \mathcal{D},y\sim\pi_\theta(\cdot|x)}\left[\frac{1}{\tau^2}\left(\log(\frac{\pi_\theta(y|x)}{\pi^\tau(y|x)})\right)^2\right] \geq0.
    \end{aligned}}
\end{equation}
Proof finished.

\subsection{The Target Distribution Equivalence of Eq.\ref{RA_eq} and Eq.\ref{pair-wised_RA_eq} }\label{Proof_RDA_lemma}
Here we prove the target distribution equivalence of Eq.\ref{RA_eq} and Eq.\ref{pair-wised_RA_eq}. Denote
\begin{equation}
\begin{aligned}
\pi^*_{\mathrm{RDA}}&=\arg\min_{\pi_\theta}\mathbb{E}_{\mathcal{D}_{\text{pw}}}\left[\left( \frac{1}{\tau}\log\frac{\pi_\theta(y_1|x)}{\pi_\theta(y_2|x)}-(r(x,y_1)-r(x,y_2))\right)^2\right],
\end{aligned}
\end{equation}
then $\pi^*_{\mathrm{RDA}}=\pi^*_{\mathrm{RA}}=\pi^\tau$.

\textbf{Proof:} 
\begin{equation}
\begin{aligned}
&\mathbb{E}_{\mathcal{D}_{\text{pw}}}\left[\left( \frac{1}{\tau}\log\frac{\pi_\theta(y_1|x)}{\pi_\theta(y_2|x)}-(r(x,y_1)-r(x,y_2))\right)^2\right]-\mathbb{E}_{\mathcal{D}_{\text{pw}}}\left[\left( \frac{1}{\tau}\log\frac{\pi^\tau(y_1|x)}{\pi^\tau(y_2|x)}-(r(x,y_1)-r(x,y_2))\right)^2\right]\\
=&\mathbb{E}_{\mathcal{D}_{\text{pw}}}\left[\left( \frac{1}{\tau}\log\frac{\pi_\theta(y_1|x)}{\pi_\theta(y_2|x)}-(r(x,y_1)-r(x,y_2))\right)^2\right] \geq0.
    \end{aligned}
\end{equation} 
When $\pi_\theta$ is equal to $\pi^\tau$, $\mathbb{E}_{\mathcal{D}_{\text{pw}}}\left[\left( \frac{1}{\tau}\log\frac{\pi_\theta(y_1|x)}{\pi_\theta(y_2|x)}-(r(x,y_1)-r(x,y_2))\right)^2\right]$ reaches the minimum value 0. Therefore $\pi^*_{\mathrm{RDA}}=\pi^*_{\mathrm{RA}}=\pi^\tau$. Proof finished.
\subsection{The Target Distribution Equivalence of Eq.\ref{pair-wised_RA_eq} and Eq.\ref{PRA_eq}}\label{PRA_optimal_proof}
Denote
\begin{equation} 
    \begin{aligned}
&\pi^*_{\mathrm{PRA}}&=\arg\min_{\pi_\theta}\mathbb{E}_{\mathcal{D}_{\theta}}\left[\mathrm{D}_{\mathrm{KL}}(p^*(z|y_1,y_2,x)||p_\theta(z|y_1,y_2,x))\right],
\end{aligned}
\end{equation}
then $\pi^*_{\mathrm{PRA}}=\pi^*_{\mathrm{RDA}}=\pi^\tau$.

\textbf{Proof:} Because $r_\theta(x, y) = \frac{1}{\tau} \log(Z(x)\pi_\theta(y|x))$, $p^*(z \mid y_1, y_2, x)=\omega(r(x,y_{2-z}),r(x,y_{z+1}))$ and $p_\theta(z|y_1,y_2,x)=\omega(r_\theta(x,y_{2-z}),r_\theta(x,y_{z+1}))$, when $\pi_\theta$ is equal to $\pi^\tau$, $r_\theta(x, y) = \frac{1}{\tau} \log(Z(x)\pi_\theta(y|x))=r(x, y)$. Then $p_\theta(z|y_1,y_2,x)=p^*(z|y_1,y_2,x)$ and $\mathbb{E}_{\mathcal{D}_{\theta}}\left[\mathrm{D}_{\mathrm{KL}}(p^*(z|y_1,y_2,x)||p_\theta(z|y_1,y_2,x))\right]=0$, then $\pi^*_{\mathrm{PRA}}=\pi^\tau=\pi^*_{\mathrm{RDA}}$. Proof finished.

\subsection{Proof of The Target Distribution Equivalence of Eq.\ref{RA_eq_posterior} and Eq.\ref{PRA_posterior}}\label{Reward_policy_posterior_proof}
\textbf{Proof:}
See the definition of $\bar{\pi}^\tau$ in Eq.\ref{pi_bar}. Denote
\begin{equation}
{\begin{aligned}
\pi^*_{\mathrm{RA-P}}&=\arg\min_{\pi_\theta}\mathbb{E}_{x\sim \mathcal{D},y\sim\pi_\theta(\cdot|x)}\left[\left(\frac{1}{\tau}\log(Z'(x)\frac{\pi_\theta(y|x)}{\pi_{ref}(y|x)})- r(x,y)\right)^2\right] \\
\pi^*_{\mathrm{PRA-P}}&=\arg\min_{\pi_\theta}\mathbb{E}_{\mathcal{D}_{\theta}}\left[\mathrm{D}_{\mathrm{KL}}(p^*(z|y_1,y_2,x)||\bar{p}_\theta(z|y_1,y_2,x))\right].
\end{aligned}}
\end{equation}
First, when $\pi_\theta$ is equal to $\bar{\pi}^\tau$, then
\begin{equation}
{\begin{aligned}
&\mathbb{E}_{x\sim \mathcal{D},y\sim\pi_\theta(\cdot|x)}\left[\left(\frac{1}{\tau}\log(Z'(x)\frac{\pi_\theta(y|x)}{\pi_{ref}(y|x)})- r(x,y)\right)^2\right] \\
=&\mathbb{E}_{x\sim \mathcal{D},y\sim\bar{\pi}^\tau(\cdot|x)}\left[\left(\frac{1}{\tau}\log(Z'(x)\frac{\bar{\pi}^\tau(y|x)}{\pi_{ref}(y|x)})- r(x,y)\right)^2\right] \\
=& \mathbb{E}_{x\sim \mathcal{D},y\sim\bar{\pi}^\tau(\cdot|x)}\left[\left(\frac{1}{\tau}\log(\exp(\tau r(x,y))- r(x,y)\right)^2\right]=0.
\end{aligned}}
\end{equation}
Because $\mathbb{E}_{x\sim \mathcal{D},y\sim\pi_\theta(\cdot|x)}\left[\left(\frac{1}{\tau}\log(Z'(x)\frac{\pi_\theta(y|x)}{\pi_{ref}(y|x)})- r(x,y)\right)^2\right]\geq0$, then $\pi^*_{\mathrm{RA-P}}=\bar{\pi}^\tau$. 

Second, when $\pi_\theta$ is equal to $\bar{\pi}^\tau$, $r_\theta(x, y) = \frac{1}{\tau} \log(Z'(x)\frac{\pi_\theta(y|x)}{\pi_{ref}(y|x)}))=\frac{1}{\tau} \log(Z'(x)\frac{\bar{\pi}^\tau(y|x)}{\pi_{ref}(y|x)}))=r(x, y)$. Then $\bar{p}_\theta(z|y_1,y_2,x)=p^*(z|y_1,y_2,x)$ and $\mathbb{E}_{\mathcal{D}_{\theta}}\left[\mathrm{D}_{\mathrm{KL}}(p^*(z|y_1,y_2,x)||\bar{p}_\theta(z|y_1,y_2,x))\right]=0$, then $\pi^*_{\mathrm{PRA-P}}=\bar{\pi}^\tau$. Then $\pi^*_{\mathrm{RA-P}}=\pi^*_{\mathrm{PRA-P}}=\bar{\pi}^\tau$. Proof finished.

\subsection{Proof of Theorem \ref{DPO_PRA}}\label{DPO_PRA_proof}
Theorem \ref{DPO_PRA}: Denote $$
\small{\begin{aligned}
&\mathcal{L}_{\mathrm{DPO}}\left(\pi_\theta\right) =\mathbb{E}_{\mathcal{D}_R}\left[-\log \sigma\left(\bar{h}_\theta\left(x, y_w, y_l\right)\right)\right], \\
&\mathcal{L}_{\mathrm{PRA-P}}\left(\pi_\theta\right) =\mathbb{E}_{\mathcal{D}_{\theta}}\left[\mathrm{D}_{\mathrm{KL}}(p^*(z|y_1,y_2,x)||\bar{p}_\theta(z|y_1,y_2,x))\right].
    \end{aligned}}$$
where $\mathcal{D}_{R}\triangleq \{(x,y_w,y_l)|x\sim\mathcal{D},y_w,y_l\sim\pi_0(\cdot|x),(y_w\succ y_l)\sim p^*(1|y_w,y_l,x)\}$ and $\mathcal{D}_{\theta}\triangleq \{(x,y_1,y_2)|x\sim\mathcal{D},y_1,y_2\sim\pi_\theta(\cdot|x),(y_1\succ y_2)\sim p^*(1|y_1,y_2,x)\}$. When $p^*$ is modeled by the BT model with Eq.\ref{BTmodel}:
\begin{equation}
\begin{aligned}
    p^*(1|y_1,y_2,x) &= \omega(r(x,y_{2-z}),r(x,y_{z+1}))=\sigma(r(x,y_1)- r(x,y_2) ),
\end{aligned}
\end{equation}
then we have the following equality:
\begin{equation}
\small{\mathcal{L}_{\mathrm{PRA-P}}\left(\pi_\theta\right) = \mathcal{L}_{\mathrm{DPO}}\left(\pi_\theta\right) + \eta_1(\pi_\theta, \pi_0) + \eta_2(\pi_\theta).}
\end{equation}
where $\eta_1(\pi_\theta, \pi_0)$ equals to 0 if and only if $\pi_\theta=\pi_0$ and $\eta_2(\pi_\theta)\triangleq\mathbb{E}_{x\sim \mathcal{D},y_1,y_2\sim \pi_\theta(y|x)}\left[M(x,y_1,y_2)\right]$. 

\textbf{Proof:} 

\begin{equation}
    \begin{aligned}
&\mathcal{L}_{\mathrm{PRA-P}}\left(\pi_\theta\right) =\mathbb{E}_{\mathcal{D}_{\theta}}\left[\mathrm{D}_{\mathrm{KL}}(p^*(z|y_1,y_2,x)||\bar{p}_\theta\left( x,y_1,y_2\right))\right]\\
=&-\mathbb{E}_{x\sim \mathcal{D},y_1,y_2\sim \pi_\theta(y|x)}\left[p^*(z=1|y_1,y_2,x)\log \bar{p}_\theta\left( x,y_1,y_2\right)+p^*(z=0|y_1,y_2,x)\log \bar{p}_\theta\left( x,y_2,y_1\right)\right]\\
&+\mathbb{E}_{x\sim \mathcal{D},y_1,y_2\sim \pi_\theta(y|x)}\left[p^*(z=1|y_1,y_2,x)\log p^*(z=1|y_1,y_2,x)+p^*(z=0|y_1,y_2,x)\log p^*(z=0|y_1,y_2,x)\right]\\
=&-\mathbb{E}_{x\sim \mathcal{D},y_1,y_2\sim \pi_\theta(y|x)}\left[p^*(z=1|y_1,y_2,x)\log \bar{p}_\theta\left( x,y_1,y_2\right)+p^*(z=0|y_1,y_2,x)\log \bar{p}_\theta\left( x,y_2,y_1\right)\right]\\
&+\mathbb{E}_{x\sim \mathcal{D},y_1,y_2\sim \pi_\theta(y|x)}\left[M(x,y_1,y_2)\right].
\end{aligned}
\end{equation}
where $\sigma(x)=\frac{1}{1+e^{-x}}$ is the sigmoid function, $M(x,y_1,y_2)=\sum_{z=0,1}p^*(z|y_1,y_2,x)\log p^*(z|y_1,y_2,x)$ and $(x,y_1,y_2)\sim \mathcal{D}_{\theta}\triangleq y_1\succ y_2\sim p^*(z=1|y_1,y_2,x),y_1,y_2\sim\pi_\theta(y|x),x\sim\mathcal{D}$.

Denote $\eta_2(\pi_\theta)=\mathbb{E}_{x\sim \mathcal{D},y_1,y_2\sim \pi_\theta(y|x)}\left[M(x,y_1,y_2)\right]$, $\zeta(\bar{p}_\theta;x,y_1,y_2)=p^*(z=1|y_1,y_2,x)\log \bar{p}_\theta\left( x,y_1,y_2\right)+p^*(z=0|y_1,y_2,x)\log \bar{p}_\theta\left( x,y_2,y_1\right)$. Then we have:
\begin{equation}
    \begin{aligned}
&\mathcal{L}_{\mathrm{PRA-P}}\left(\pi_\theta\right) = -\mathbb{E}_{x\sim \mathcal{D},y_1,y_2\sim \pi_\theta(y|x)}\left[ \zeta(\bar{p}_\theta;x,y_1,y_2) \right] + \eta_2(\pi_\theta)\\
=&-\mathbb{E}_{x\sim \mathcal{D},y_1,y_2\sim \pi_0(y|x)}\left[ \zeta(\bar{p}_\theta;x,y_1,y_2) \right] + \left(\sum_{x\in\mathbb{X},y_1,y_2\in\mathbb{Y}}(\pi_0(y|x)-\pi_\theta(y|x))\cdot\zeta(\bar{p}_\theta;x,y_1,y_2)\right) + \eta_2(\pi_\theta).
    \end{aligned}
\end{equation}
Denote $\sum_{x\in\mathbb{X},y_1,y_2\in\mathbb{Y}}(\pi_0(y|x)-\pi_\theta(y|x))\cdot\zeta(\bar{p}_\theta;x,y_1,y_2)$ as $\eta_1(\pi_\theta, \pi_0)$. Based on Eq.\ref{DPO_eq_from_PRA_proof} and $p^*(1|y_1,y_2,x) = \omega(r(x,y_{2-z}),r(x,y_{z+1}))=\sigma(r(x,y_1)- r(x,y_2) )$ because of the BT model, then we have the final conclusion:
\begin{equation}
    \mathcal{L}_{\mathrm{PRA-P}}\left(\pi_\theta\right) = \mathcal{L}_{\mathrm{DPO}}\left(\pi_\theta\right) + \eta_1(\pi_\theta, \pi_0) + \eta_2(\pi_\theta).
\end{equation}
Proof finished.

\section{Proof in Section \ref{Convergence_sec}}\label{Convergence_sec_proof}

\begin{proposition}\label{SGD_theorem}
If $F(\theta)$ is $L$-smooth for $\theta$ and $\|g_t\|^2\leq G^2$. Suppose the parameters $\theta$ are updated by:
\begin{equation}
    \theta_{t+1}=\theta_{t} - \alpha_tg(x,y,\theta_t),
\end{equation}
Then the following inequality holds:
\begin{equation}\label{SGD_eq}
    \small{\begin{aligned}
        \min_{1\leq t\leq T-1}||\nabla F(\theta_t)||^2_2\leq \frac{LG^2\sum_{t=1}^{T-1}\alpha_t^2+2(F(\theta_1)-F^*)}{2\sum_{t=1}^{T-1}\alpha_t}.
    \end{aligned}}
\end{equation}
where $F^*=\arg\min_\theta F(\theta)$. 
\end{proposition}
\textbf{Proof of Sketch:} Before proof, firstly we need to introduce a formal setting. Assume that the parameterized policy $\pi_\theta$ is given by Definition \ref{Softmax} and the parameters $\theta$ is optimized by the Stochastic Gradient Descent (SGD) method. Denote $F(\theta)=\mathbb{E}_{x,y\sim p(x,y;\pi_\theta)}[f(x,y, \theta)]$ where $p(x,y;\pi_\theta)$ is a distribution and $f(x,y, \theta)$ is the loss function on data $(x,y)$ for policy parameters $\theta$. Let $g(x,y,\theta_k)$ be the stochastic gradient of $\nabla F(\theta_k)$, i.e. $E_{x,y\sim p(x,y;\pi_\theta)}[g(x,y,\theta)]=\nabla_\theta F(\theta)$. Usually $g(x,y,\theta_t)$ is simplified to $g_t$ when it does not cause ambiguity. With learning rate $\alpha_t$, the SGD update rule is Eq.\ref{SGD_eq}. For Theorem \ref{SGD_theorem}, since the $L$-smooth property  cannot guarantee convex, thus for a non-convex optimization problem, the commonly used measure is the gradient norm. Therefore, with the properties of $L$-smooth and SGD, Proposition \ref{SGD_theorem} holds. See details in Appendix \ref{SGD_theorem_proof}.

Observing Eq.\ref{SGD_eq}, $||\nabla F(\theta_t)||^2_2$ will converge to 0 when $T\rightarrow\infty$ if $\sum_{t=1}^\infty\alpha_t^2\leq\infty$. The smaller the variance $G$ of stochastic gradient $g_t$ is, the faster the convergence rate of $||\nabla F(\theta_t)||^2_2$ is. For the methods in Figure \ref{Framework_figure}, we prove that the loss functions of these methods all satisfy the $L$-smooth assumption with different coefficients under the softmax parametrization of policy $\pi_\theta$ (See Table \ref{smoothCoefficients}, corresponding to Lemma \ref{BDA_L_RKL_coef}-\ref{PRA_L_coef}). 
\subsection{Proof of Proposition \ref{SGD_theorem}}\label{SGD_theorem_proof}
\textbf{Proof:} Because $F(\theta)$ is $L$-smooth for $\theta$, we have:
\begin{equation}
    \left|F(\theta_{t+1})-F(\theta_{t})-\nabla_\theta F(\theta_{t})(\theta_{t+1}-\theta_{t})\right|\leq \frac{L}{2}\|\theta_{t+1}-\theta_{t}\|_2^2.
\end{equation}
where $\theta_{t+1}=\theta_{t}-\alpha_tg_t$.

Substituting into $\theta_{t+1}=\theta_{t}-\alpha_tg_t$, we get:
\begin{equation}\label{L_smooth_proof2}
    \begin{aligned}
& (F(\theta_{t+1})-F(\theta_{t})+\alpha_t\nabla_\theta F(\theta_{t})g_t) \\
\leq& \left|F(\theta_{t+1})-F(\theta_{t})+\alpha_t\nabla_\theta F(\theta_{t})g_t\right|\\
=&\left|F(\theta_{t+1})-F(\theta_{t})-\nabla_\theta F(\theta_{t})(\theta_{t+1}-\theta_{t})\right| \\
\leq& \frac{L}{2}\|\theta_{t+1}-\theta_{t}\|_2^2 = \frac{L}{2}\|\alpha_t g_t\|_2^2 \leq \frac{L}{2}\alpha_t^2 G^2.
    \end{aligned}
\end{equation}
Base on Eq.\ref{L_smooth_proof2}, we get:
\begin{equation}
    \alpha_t\nabla_\theta F(\theta_{t})g_t \leq \frac{L}{2}\alpha_t^2 G^2 + F(\theta_{t}) -F(\theta_{t+1}).
\end{equation}
Summing both sides with respect to $t$ yields:
\begin{equation}
    \begin{aligned}
        \sum_{t=1}^{T-1}\alpha_t\nabla_\theta F(\theta_{t})g_t \leq \frac{L}{2} G^2\sum_{t=1}^{T-1}\alpha_t^2 + F(\theta_{1}) -F(\theta_{T})\leq \frac{L}{2} G^2\sum_{t=1}^{T-1}\alpha_t^2 + F(\theta_{1}) -F^*.
    \end{aligned}
\end{equation}
where $F^*=\arg\min_\theta F(\theta)$.

Expect the stochastic gradient $g_t$ to get:
\begin{equation}
    \begin{aligned}
        \min_{1\leq t\leq T-1}||\nabla_\theta F(\theta_{t})||_2^2\sum_{t=1}^{T-1}\alpha_t\leq \sum_{t=1}^{T-1}\alpha_t||\nabla_\theta F(\theta_{t})||_2^2 \leq \frac{L}{2} G^2\sum_{t=1}^{T-1}\alpha_t^2 + F(\theta_{1}) -F^*.
    \end{aligned}
\end{equation}
Finally we have:
\begin{equation}
    \begin{aligned}
        \min_{1\leq t\leq T-1}||\nabla F(\theta_t)||^2_2\leq \frac{LG^2\sum_{t=1}^{T-1}\alpha_t^2+2(F(\theta_1)-F^*)}{2\sum_{t=1}^{T-1}\alpha_t}.
    \end{aligned}
\end{equation}
Proof finished.

\subsection{Proof of Theorem \ref{DPO_theorem}}\label{DPO_theorem_proof}
\textbf{Theorem \ref{DPO_theorem}:} Assume $\|g_t\|^2\leq G^2$. Given the Definition \ref{Softmax} for policy $\pi_\theta$ and the learning rate $\alpha_t$, for DPO algorithm, the following properties hold:
    
    (0). $\forall r, \tau$, then $\theta\rightarrow \mathcal{L}_{\mathrm{DPO}}\left(\pi_\theta ; \pi_{\text {ref }}\right)$ is $\frac{4}{\tau^2}$-smooth.

    (1). If there is  $\pi_\theta^*$ such that $\mathbb{E}_{x\sim \mathcal{D},y_1,y_2\sim \pi_0(y|x)}[(p^*(1|y_1,y_2,x) - \sigma(\bar{h}_\theta(x,y_1,y_2)))]=0$, then $\pi_\theta^* = \arg\min_{\pi_\theta} \mathcal{L}_{\mathrm{DPO}}(\pi_{\theta} ; \pi_{\text {ref }}, \pi_0)$. And $\bar{\pi}^\tau(y|x)=\frac{\pi_{ref}(y|x)\exp(\tau r(x,y))}{Z'(x)}$ is an example $\pi_\theta^*$ with the Bradley-Terry (BT) model.

    (2). Denote $\mathcal{L}_{\mathrm{DPO}}^*=\min_{\pi_\theta} \mathcal{L}_{\mathrm{DPO}}(\pi_{\theta} ; \pi_{\text {ref }}, \pi_0)$. $\min_{1\leq i\leq T}||\nabla_\theta \mathcal{L}_{\mathrm{DPO}}(\pi_{\theta_i} ; \pi_{\text {ref }})||^2_2\leq \frac{2G^2\sum_{t=1}^{T-1}\alpha_t^2}{\tau^2\sum_{t=1}^{T-1}\alpha_t} + \frac{\mathcal{L}_{\mathrm{DPO}}(\pi_{\theta_T})-\mathcal{L}_{\mathrm{DPO}}^*}{\sum_{t=1}^{T-1}\alpha_t}$. 

See $\mathcal{L}_{\mathrm{DPO}}(\pi_{\theta} ; \pi_{\text {ref }})$ on Eq.\ref{DPO_eq}.  $\mathcal{D}_{R}\triangleq \{(x,y_w,y_l)|x\sim\mathcal{D},y_w,y_l\sim\pi_0(\cdot|x),(y_w\succ y_l)\sim p^*(1|y_w,y_l,x)\}$ which $\pi_0$ is an unanalytical distribution, $\mathcal{D}$ is an arbitrary distribution and $\{z=1|y_1,y_2,x\}\triangleq \{r(x,y_1)\geq r(x,y_2)\}$.
\begin{equation}
\begin{gathered}
\bar{h}_\theta\left(x, y_w, y_l\right)=\frac{1}{\tau} \log \frac{\pi_\theta\left(y_w \mid x\right)}{\pi_{\text {ref }}\left(y_w \mid x\right)}-\frac{1}{\tau} \log \frac{\pi_\theta\left(y_l \mid x\right)}{\pi_{\text {ref }}\left(y_l \mid x\right)}, \\
\mathcal{L}_{\mathrm{DPO}}\left(\pi_\theta ; \pi_{\text {ref }}\right)=-\mathbb{E}_{\left(x, y_w, y_l\right) \sim \mathcal{D}_R}\left[\log \sigma\left(\bar{h}_\theta\left(x, y_w, y_l\right)\right)\right].
\end{gathered}
\end{equation}

\textbf{Proof:}

\textbf{(0).} $\forall r, \tau$, then $\theta\rightarrow \mathcal{L}_{\mathrm{DPO}}\left(\pi_\theta ; \pi_{\text {ref }}\right)$ is $\frac{8}{\tau^2}$-smooth.

Proof: By Lemma \ref{spectral_radius}, it suffices to show that the spectral radius of the hessian matrix of the second derivative of $\mathcal{L}_{\mathrm{DPO}}(\pi_{\theta} ; \pi_{\text {ref }})$, i.e.
\begin{equation}\label{spectral_radius_DPO}
\begin{aligned}
&\left|\sum_{x,x'\in\mathbb{X}}\sum_{y_i,y_j\in\mathbb{Y}}z(x,y_i) \frac{\partial^2 \mathcal{L}_{\mathrm{DPO}}(\pi_{\theta} ; \pi_{\text {ref }})}{\partial \theta(x,y_i)\partial\theta(x',y_j)} z(x',y_j)\right|    \leq\left(\frac{4}{\tau^2}\right)||z(\cdot,\cdot)||_2^2.
\end{aligned}
\end{equation}

Denote $h(\pi_\theta,x,y_1,y_2)=-p^*(1|y_1,y_2,x)\log\sigma\left( \bar{h}_\theta(x,y_1,y_2)\right)-p^*(0|y_1,y_2,x)\log\sigma\left( \bar{h}_\theta(x,y_2,y_1)\right)$, we have:
\begin{equation}\label{dpo_decompose_psi_with_M_eq}
\begin{aligned}
&\left|\sum_{x,x'\in\mathbb{X}}\sum_{y_i,y_j\in\mathbb{Y}}z(x,y_i) \frac{\partial^2 \mathcal{L}_{\mathrm{DPO}}\left(\pi_\theta ; \pi_{\text {ref }}\right)}{\partial \theta(x,y_i)\partial\theta(x',y_j)} z(x',y_j)\right|\\
=&\left|\sum_{x\in\mathbb{X}}\sum_{y_i,y_j\in\mathbb{Y}}z(x,y_i) \frac{\partial^2 \sum_{x\in\mathbb{X}}\mathcal{D}(x)\sum_{y_1,y_2\in\mathbb{Y}}\pi_0(y_1|x)\pi_0(y_2|x)h(\pi_\theta,x,y_1,y_2)}{\partial \theta(x,y_i)\partial\theta(x,y_j)}  z(x,y_j)\right|\\ =&\left|\sum_{x\in\mathbb{X}}\mathcal{D}(x)\sum_{y_i,y_j\in\mathbb{Y}}z(x,y_i)  \frac{\partial^2 f_{\text{DPO}}(x,\theta)}{\partial \theta(x, y_i) \partial \theta(x, y_j)}  z(x,y_j)\right| \\
\triangleq& |\sum_{x\in\mathbb{X}}\mathcal{D}(x) \psi(x)|  \leq \|\mathcal{D}(\cdot)\|_1\|\psi(\cdot)\|_\infty  = 1\cdot\|\psi(\cdot)\|_\infty  .
\end{aligned}
\end{equation}
where $f_{\text{DPO}}(x,\theta)=\sum_{y_1,y_2\in\mathbb{Y}}\pi_0(y_1|x)\pi_0(y_2|x)h(\pi_\theta,x,y_1,y_2)$ and $\psi(x)=\sum_{y_i,y_j\in\mathbb{Y}}z(x,y_i)  \frac{\partial^2 f_{\text{DPO}}(x,\theta)}{\partial \theta(x, y_i) \partial \theta(x, y_j)}  z(x,y_j)$.

The second derivative of $f_{\text{DPO}}(x,\theta)$ is:
\begin{equation}\label{DPO_decompose}
    \begin{aligned}
        \frac{\partial^2 f_{\text{DPO}}(x, \theta)}{\partial \theta(x, y_i) \partial \theta(x, y_j)} = \sum_{y_1,y_2\in\mathbb{Y}}& \pi_0(y_1|x)\pi_0(y_2|x)\frac{\partial^2 h(\pi_\theta, x, y_1,y_2)}{\partial \theta(x, y_i) \partial \theta(x, y_j)}.
    \end{aligned}
\end{equation}

Consider the first derivative of  $h(\pi_\theta,x,y_1,y_2)$, denote $p^*=p^*(1|y_1,y_2,x)$:
\begin{equation}
    \begin{aligned}
&\frac{\partial h(\pi_\theta,x,y_1,y_2)}{\partial \theta(x, y_i)} \\
=& -p^*\frac{\partial }{\partial \theta(x, y_i) }(\log\sigma(\frac{1}{\tau}\log\frac{\pi_\theta(y_1|x)\pi_{\text{ref}}(y_2|x)}{\pi_\theta(y_2|x)\pi_{\text{ref}}(y_1|x)})) - (1-p^*)\frac{\partial }{\partial \theta(x, y_i) }(\log\sigma(\frac{1}{\tau}\log\frac{\pi_\theta(y_1|x)\pi_{\text{ref}}(y_2|x)}{\pi_\theta(y_2|x)\pi_{\text{ref}}(y_1|x)})) \\
=& -\left(p^*(1-\sigma(\frac{1}{\tau}\log\frac{\pi_\theta(y_1|x)\pi_{\text{ref}}(y_2|x)}{\pi_\theta(y_2|x)\pi_{\text{ref}}(y_1|x)})) - (1-p^*)\sigma(\frac{1}{\tau}\log\frac{\pi_\theta(y_1|x)\pi_{\text{ref}}(y_2|x)}{\pi_\theta(y_2|x)\pi_{\text{ref}}(y_1|x)}) \right)\frac{1}{\tau}\frac{\partial }{\partial \theta(x, y_i) }(\log\frac{\pi_\theta(y_1|x)}{\pi_\theta(y_2|x)}) \\
=& -\left(p^* - \sigma(\frac{1}{\tau}\log\frac{\pi_\theta(y_1|x)\pi_{\text{ref}}(y_2|x)}{\pi_\theta(y_2|x)\pi_{\text{ref}}(y_1|x)}) \right) \frac{1}{\tau}(\delta_{y_1y_i}-\delta_{y_2y_i}).
    \end{aligned}
\end{equation}

Consider the second derivative of  $h(\pi_\theta,x,y_1,y_2)$:
\begin{equation}
    \begin{aligned}
&\frac{\partial^2 h(\pi_\theta,x,y_1,y_2)}{\partial \theta(x, y_i)\partial \theta(x, y_j)}\\ =& -\frac{\partial }{\partial \theta(x, y_j) }( \sigma(\frac{1}{\tau}\log\frac{\pi_\theta(y_2|x)\pi_{\text{ref}}(y_1|x)}{\pi_\theta(y_1|x)\pi_{\text{ref}}(y_2|x)}) \frac{1}{\tau}(\delta_{y_1y_i}-\delta_{y_2y_i}) ) \\
=& \frac{1}{\tau^2}(\delta_{y_1y_i}-\delta_{y_2y_i})(\delta_{y_1y_j}-\delta_{y_2y_j})\sigma(\frac{1}{\tau}\log\frac{\pi_\theta(y_1|x)\pi_{\text{ref}}(y_2|x)}{\pi_\theta(y_2|x)\pi_{\text{ref}}(y_1|x)})\sigma(\frac{1}{\tau}\log\frac{\pi_\theta(y_2|x)\pi_{\text{ref}}(y_1|x)}{\pi_\theta(y_1|x)\pi_{\text{ref}}(y_2|x)}).
    \end{aligned}
\end{equation}

Then:
\begin{equation}\label{DPO_4term_decompose}
\begin{aligned}
&|\psi(x)|=|\sum_{y_i,y_j\in\mathbb{Y}}z(x,y_i)  \frac{\partial^2 f_{\text{DPO}}(x,\theta)}{\partial \theta(x, y_i) \partial \theta(x, y_j)}  z(x,y_j)| \\
=&|\sum_{y_i,y_j\in\mathbb{Y}}z(x,y_i)  \sum_{y_1,y_2\in\mathbb{Y}} \pi_0(y_1|x)\pi_0(y_2|x)\frac{\partial^2 h(\pi_\theta, x, y_1,y_2)}{\partial \theta(x, y_i) \partial \theta(x, y_j)}  z(x,y_j)| \\
\leq &|\sum_{y_i,y_j\in\mathbb{Y}}z(x,y_i)  \sum_{y_1,y_2\in\mathbb{Y}} \pi_0(y_1|x)\pi_0(y_2|x)\frac{1}{\tau^2}(\delta_{y_1y_i}-\delta_{y_2y_i})(\delta_{y_1y_j}-\delta_{y_2y_j})  z(x,y_j)| \\
=& |\sum_{y_i\in\mathbb{Y}}z(x,y_i)   \pi_0(y_i|x)\frac{1}{\tau^2}z(x,y_i) - \sum_{y_i\in\mathbb{Y}}z(x,y_i)   \pi_0(y_i|x)\sum_{y_j\in\mathbb{Y}}\pi_0(y_j|x)\frac{1}{\tau^2}z(x,y_j) \\
&- \sum_{y_i\in\mathbb{Y}}z(x,y_i)   \pi_0(y_i|x)\sum_{y_j\in\mathbb{Y}}\pi_0(y_j|x)\frac{1}{\tau^2}z(x,y_j) + \sum_{y_i\in\mathbb{Y}}z(x,y_i)   \pi_0(y_i|x)\frac{1}{\tau^2}z(x,y_i)| \\
\leq& \frac{2}{\tau^2}|\sum_{y_i\in\mathbb{Y}}z(x,y_i)   \pi_0(y_i|x)z(x,y_i)| + |\sum_{y_i\in\mathbb{Y}}z(x,y_i)   \pi_0(y_i|x)\sum_{y_j\in\mathbb{Y}}\pi_0(y_j|x)z(x,y_j)|\\
=& \frac{2}{\tau^2} (\pi^T_0(\cdot|x)z^2(x,\cdot) + (\pi^T_0(\cdot|x)z(x,\cdot))^2) \leq \frac{2}{\tau^2} (\|\pi_0(\cdot|x)\|_1\|z^2(x,\cdot)\|_\infty + (\|\pi_0(\cdot|x)\|_1\|z(x,\cdot)\|_\infty)^2) \\
\leq& \frac{4}{\tau^2}\|z(x,\cdot)\|_2^2.
\end{aligned}
\end{equation}
The first inequality is because $\sigma(\cdot)\leq 1$. Therefore,
\begin{equation}
    \begin{aligned}
\|\psi(\cdot)\|_\infty &= \max_{x\in\mathbb{X}}|\sum_{y_i,y_j\in\mathbb{Y}}z(x,y_i)  \frac{\partial^2 f_{\text{DPO}}(x,\theta)}{\partial \theta(x, y_i) \partial \theta(x, y_j)}  z(x,y_j)| \leq \max_{x\in\mathbb{X}}\frac{4}{\tau^2}\|z(x,\cdot)\|_2^2 \leq \frac{4}{\tau^2}\|z(\cdot,\cdot)\|_2^2.
    \end{aligned}
\end{equation}
Then Eq.\ref{spectral_radius_DPO} is proved. Proof finished. 

\textbf{(1).} If $\pi_\theta$ such that $\mathbb{E}_{x\sim \mathcal{D},y_1,y_2\sim \pi_0(y|x)}[(p^*(1|y_1,y_2,x) - \sigma(\bar{h}_\theta(x,y_1,y_2)))]=0$, then the DPO's loss function get a local optimal solution (zero gradient point) where $\pi_0$ is another unknown distribution. And $\bar{\pi}^\tau(y|x)=\frac{\pi_{ref}(y|x)\exp(\tau r(x,y))}{Z'(x)}$ is an example with the Bradley-Terry (BT) model.

Proof: From the above theorem, it is evident that the optimal solution for $\pi_\theta$ is influenced by the comparison distribution $p^*(z|y_1,y_2,x)$. Here, we provide an existence proof. We first propose a candidate solution $\bar{\pi}^\tau(y|x)=\frac{\pi_{ref}(y|x)\exp(\tau r(x,y))}{Z'(x)}$ where $Z'(x)=\sum_{y'\in Y}\pi_{ref}(y'|x)\exp({\tau r(x,y')})$, and subsequently demonstrate that if $p^*(z|y_1,y_2,x)$ is modeled by the Bradley-Terry (BT) model \cite{bradley1952rank}, solution $\bar{\pi}^\tau(y|x)=\frac{\pi_{ref}(y|x)\exp(\tau r(x,y))}{Z'(x)}$ satisfies condition $\mathbb{E}_{x\sim \mathcal{D},y_1,y_2\sim \pi_0(y|x)}[(p^*(1|y_1,y_2,x) - \sigma(\bar{h}_\theta(x,y_1,y_2)))]=0$ as a locally optimal solution. But it is not necessarily a global optimum, as we can only verify that $\bar{\pi}^\tau(y|x)=\frac{\pi_{ref}(y|x)\exp(\tau r(x,y))}{Z'(x)}$ is a point where the gradient of the loss function $\mathcal{L}_{\mathrm{DPO}}(\pi_{\theta} ; \pi_{\text {ref }})$ equals zero.

To compute the gradient of the loss function $\mathbb{E}_{x \sim \mathcal{D}, y_1, y_2 \sim \pi_0(y|x)}\left[\mathrm{D}_{\mathrm{KL}}(p^*(z|y_1, y_2, x) || \bar{p}_\theta(z|y_1, y_2,x))\right]$ with respect to $\theta$, we proceed as follows:
\begin{equation}
    \begin{aligned}
&\nabla_\theta \mathbb{E}_{x \sim \mathcal{D}, y_1, y_2 \sim \pi_0(y|x)}\left[\mathrm{D}_{\mathrm{KL}}(p^*(z|y_1, y_2, x) || \bar{p}_\theta(z|y_1, y_2,x))\right] \\
=& -\nabla_\theta \mathbb{E}_{x \sim \mathcal{D}, y_1, y_2 \sim \pi_0(y|x)} \left[p^*(1|y_1, y_2, x) \log \sigma(\bar{h}_\theta(x, y_1, y_2)) + p^*(0|y_1, y_2, x) \log \sigma(\bar{h}_\theta(x, y_2, y_1)) \right] \\
=& -\mathbb{E}_{x \sim \mathcal{D}, y_1, y_2 \sim \pi_0(y|x)} \left[p^*(1|y_1, y_2, x) \nabla_\theta \log \sigma(\bar{h}_\theta(x, y_1, y_2)) + p^*(0|y_1, y_2, x) \nabla_\theta \log \sigma(\bar{h}_\theta(x, y_2, y_1)) \right] \\
=& -\mathbb{E}_{x \sim \mathcal{D}, y_1, y_2 \sim \pi_0(y|x)} \left[ \frac{p^*(1|y_1, y_2, x)}{\sigma(\bar{h}_\theta(x, y_1, y_2))} \nabla_\theta \sigma(\bar{h}_\theta(x, y_1, y_2)) + \frac{p^*(0|y_1, y_2, x)}{\sigma(\bar{h}_\theta(x, y_2, y_1))} \nabla_\theta \sigma(\bar{h}_\theta(x, y_2, y_1)) \right] \\
=& -\mathbb{E}_{x \sim \mathcal{D}, y_1, y_2 \sim \pi_0(y|x)} \left[\frac{p^*(1|y_1, y_2, x)}{\sigma(\bar{h}_\theta(x, y_1, y_2))} \nabla_\theta \sigma(\bar{h}_\theta(x, y_1, y_2)) - \frac{1 - p^*(1|y_1, y_2, x)}{1 - \sigma(\bar{h}_\theta(x, y_1, y_2))} \nabla_\theta \sigma(\bar{h}_\theta(x, y_1, y_2)) \right] \\
=& -\mathbb{E}_{x \sim \mathcal{D}, y_1, y_2 \sim \pi_0(y|x)} \left[\left(\frac{p^*(1|y_1, y_2, x)(1 - \sigma(\bar{h}_\theta)) - (1 - p^*(1|y_1, y_2, x)) \sigma(\bar{h}_\theta)}{\sigma(\bar{h}_\theta(x, y_1, y_2))(1 - \sigma(\bar{h}_\theta)(x, y_1, y_2))}\right) \nabla_\theta \sigma(\bar{h}_\theta)\right] \\
=& -\mathbb{E}_{x \sim \mathcal{D}, y_1, y_2 \sim \pi_0(y|x)} \left[\left(\frac{p^*(1|y_1, y_2, x) - \sigma(\bar{h}_\theta(x, y_1, y_2))}{\sigma(\bar{h}_\theta(x, y_1, y_2))(1 - \sigma(\bar{h}_\theta(x, y_1, y_2)))}\right) \nabla_\theta \sigma(\bar{h}_\theta(x, y_1, y_2))\right] \\
=& -\mathbb{E}_{x \sim \mathcal{D}, y_1, y_2 \sim \pi_0(y|x)} \left[\left(p^*(1|y_1, y_2, x) - \sigma(\bar{h}_\theta(x, y_1, y_2))\right) \nabla_\theta \bar{h}_\theta(x, y_1, y_2)\right].
\end{aligned}
\end{equation}
where $z\in\{0,1\}$, $\bar{p}_\theta(0|y_1,y_2,x)=\sigma\left( \bar{h}_\theta(x,y_2,y_1)\right)$ and $\bar{p}_\theta(1|y_1,y_2,x)=\sigma\left( \bar{h}_\theta(x,y_1,y_2)\right)$.

Under the assumption of the Bradley-Terry (BT) model, which posits:
\begin{equation}
    p^*(1|y_1, y_2, x) = \frac{1}{1 + e^{-(r(x, y_1) - r(x, y_2))}} = \sigma(h(x, y_1, y_2)),
\end{equation}
substituting this into the gradient of the target function, we obtain:
\begin{equation}
    \begin{aligned}
&\nabla_\theta \mathbb{E}_{x \sim \mathcal{D}, y_1, y_2 \sim \pi_0(y|x)}\left[\mathrm{D}_{\mathrm{KL}}(p^*(z|y_1, y_2, x) || \bar{p}_\theta(z|y_1, y_2,x))\right] \\
=& -\mathbb{E}_{x \sim \mathcal{D}, y_1, y_2 \sim \pi_0(y|x)} \left[\left(p^*(1|y_1, y_2, x) - \sigma(\bar{h}_\theta(x, y_1, y_2))\right) \nabla_\theta \bar{h}_\theta(x, y_1, y_2)\right] \\
=& -\mathbb{E}_{x \sim \mathcal{D}, y_1, y_2 \sim \pi_0(y|x)} \left[\left(\sigma(h(x, y_1, y_2)) - \sigma(\bar{h}_\theta(x, y_1, y_2))\right) \nabla_\theta \bar{h}_\theta(x, y_1, y_2)\right].
\end{aligned}
\end{equation}
where $\bar{h}_\theta\left(x, y_1, y_2\right)=\frac{1}{\tau} \log \frac{\pi_\theta\left(y_1 \mid x\right)}{\pi_{\text {ref }}\left(y_1 \mid x\right)}-\frac{1}{\tau} \log \frac{\pi_\theta\left(y_2 \mid x\right)}{\pi_{\text {ref }}\left(y_2 \mid x\right)}$ and ${h}\left(x, y_1, y_2\right)=r(x,y_1)-r(x,y_2)$.

Given that $\sigma(\cdot)$ is a strictly increasing function, it follows that when $\pi_\theta(y|x)=\bar{\pi}^\tau(y|x)=\frac{\pi_{ref}(y|x)\exp(\tau r(x,y))}{Z'(x)}$, we have $\bar{h}_\theta(x, y_1, y_2) = h(x, y_1, y_2)$ and $\nabla_\theta \mathbb{E}_{x \sim \mathcal{D}, y_1, y_2 \sim \pi_0(y|x)} \left[\mathrm{D}_{\mathrm{KL}}(p^*(z|y_1, y_2, x) || \bar{p}_\theta(z|y_1, y_2,x))\right] = 0$. Proof finished.

\textbf{(2).} Denote $\mathcal{L}_{\mathrm{DPO}}^*=\min_{\pi_\theta} \mathcal{L}_{\mathrm{DPO}}(\pi_{\theta} ; \pi_{\text {ref }}, \pi_0)$. $\min_{1\leq i\leq T}||\nabla_\theta \mathcal{L}_{\mathrm{DPO}}(\pi_{\theta_i} ; \pi_{\text {ref }})||^2_2\leq \frac{2G^2\sum_{t=1}^{T-1}\alpha_t^2}{\tau^2T} + \frac{\mathcal{L}_{\mathrm{DPO}}(\pi_{\theta_T})-\mathcal{L}_{\mathrm{DPO}}^*}{T}$. 

This theorem can be obtained from Theorem \ref{DPO_theorem}.(1) and Theorem \ref{SGD_theorem}.

\subsection{Proof of Lemma \ref{data_select}}\label{data_select_proof}
\textbf{Lemma \ref{data_select}:} Assume $\|g_t\|^2\leq G^2$. Given the Definition \ref{Softmax} for policy $\pi_\theta$ and the learning rate $\alpha_t$, for DPO algorithm, let $\gamma(x)\leq\gamma$, when $\pi_0$ is uniform distribution, then:
\begin{equation}  
\small{    \begin{aligned}
\min_{1\leq i\leq T}&||\nabla_\theta \mathcal{L}_{\mathrm{DPO}}(\pi_{\theta_i} ; \pi_{\text {ref }})||^2_2\leq \frac{\mathcal{L}_{\mathrm{DPO}}(\pi_{\theta_T})-\mathcal{L}_{\mathrm{DPO}}^*}{\sum_{t=1}^{T-1}\alpha_t} + \frac{2(\gamma c_0+1)G^2\sum_{t=1}^{T-1}\alpha_t^2}{\tau^2\sum_{t=1}^{T-1}\alpha_t}.
    \end{aligned}}
\end{equation} 
where $\mathcal{L}_{\mathrm{DPO}}^*=\min_{\pi_\theta} \mathcal{L}_{\mathrm{DPO}}(\pi_{\theta} ; \pi_{\text {ref }}, \pi_0)$ and $c_0=\sigma(\frac{\epsilon_0}{\tau})\sigma(-\frac{\epsilon_0}{\tau})-1\in (-1,0)$. See $\mathcal{L}_{\mathrm{DPO}}(\pi_{\theta} ; \pi_{\text {ref }})$ on Eq.\ref{DPO_eq}.

\textbf{Proof:} Based on Theorem \ref{SGD_theorem}, we only need to know $\mathcal{L}_{\mathrm{DPO}}(\pi_{\theta}; \pi_{\text {ref }})$ is $\frac{4(\gamma c_0+1)}{\tau^2}$-smooth.

Denote $h(\pi_\theta,x,y_1,y_2)=-p^*(1|y_1,y_2,x)\log\sigma\left( \bar{h}_\theta(x,y_1,y_2)\right)-p^*(0|y_1,y_2,x)\log\sigma\left( \bar{h}_\theta(x,y_2,y_1)\right)$, we have:
\begin{equation}
\begin{aligned}
&\left|\sum_{x,x'\in\mathbb{X}}\sum_{y_i,y_j\in\mathbb{Y}}z(x,y_i) \frac{\partial^2 \mathcal{L}_{\mathrm{DPO}}\left(\pi_\theta ; \pi_{\text {ref }}\right)}{\partial \theta(x,y_i)\partial\theta(x',y_j)} z(x',y_j)\right|\\
=&\left|\sum_{x\in\mathbb{X}}\sum_{y_i,y_j\in\mathbb{Y}}z(x,y_i) \frac{\partial^2 \sum_{x\in\mathbb{X}}\mathcal{D}(x)\sum_{y_1,y_2\in\mathbb{Y}}\pi_0(y_1|x)\pi_0(y_2|x)h(\pi_\theta,x,y_1,y_2)}{\partial \theta(x,y_i)\partial\theta(x,y_j)}  z(x,y_j)\right|\\ =&\left|\sum_{x\in\mathbb{X}}\mathcal{D}(x)\sum_{y_i,y_j\in\mathbb{Y}}z(x,y_i)  \frac{\partial^2 f_{\text{DPO}}(x,\theta)}{\partial \theta(x, y_i) \partial \theta(x, y_j)}  z(x,y_j)\right| 
\triangleq |\sum_{x\in\mathbb{X}}\mathcal{D}(x) \psi(x)|  \leq \|\mathcal{D}(\cdot)\|_1\|\psi(\cdot)\|_\infty  = 1\cdot\|\psi(\cdot)\|_\infty  .
\end{aligned}
\end{equation}
where $f_{\text{DPO}}(x,\theta)=\sum_{y_1,y_2\in\mathbb{Y}}\pi_0(y_1|x)\pi_0(y_2|x)h(\pi_\theta,x,y_1,y_2)$ and $\psi(x)=\sum_{y_i,y_j\in\mathbb{Y}}z(x,y_i)  \frac{\partial^2 f_{\text{DPO}}(x,\theta)}{\partial \theta(x, y_i) \partial \theta(x, y_j)}  z(x,y_j)$.

Because $\pi_0$ is uniform distribution, the second derivative of $f_{\text{DPO}}(x,\theta)$ is:
\begin{equation}
    \begin{aligned}
\frac{\partial^2 f_{\text{DPO}}(x, \theta)}{\partial \theta(x, y_i) \partial \theta(x, y_j)} = \sum_{y_1,y_2\in\mathbb{Y}}& \frac{1}{K^2} \frac{\partial^2 h(\pi_\theta, x, y_1,y_2)}{\partial \theta(x, y_i) \partial \theta(x, y_j)}.
    \end{aligned}
\end{equation}

Consider the second derivative of  $h(\pi_\theta,x,y_1,y_2)$:
\begin{equation}
    \begin{aligned}
&\frac{\partial^2 h(\pi_\theta,x,y_1,y_2)}{\partial \theta(x, y_i)\partial \theta(x, y_j)}= -\frac{\partial }{\partial \theta(x, y_j) }( \sigma(\frac{1}{\tau}\log\frac{\pi_\theta(y_2|x)\pi_{\text{ref}}(y_1|x)}{\pi_\theta(y_1|x)\pi_{\text{ref}}(y_2|x)}) \frac{1}{\tau}(\delta_{y_1y_i}-\delta_{y_2y_i}) ) \\
=& \frac{1}{\tau^2}(\delta_{y_1y_i}-\delta_{y_2y_i})(\delta_{y_1y_j}-\delta_{y_2y_j})\sigma(\frac{1}{\tau}\log\frac{\pi_\theta(y_1|x)\pi_{\text{ref}}(y_2|x)}{\pi_\theta(y_2|x)\pi_{\text{ref}}(y_1|x)})\sigma(\frac{1}{\tau}\log\frac{\pi_\theta(y_2|x)\pi_{\text{ref}}(y_1|x)}{\pi_\theta(y_1|x)\pi_{\text{ref}}(y_2|x)})\\
=& \frac{1}{\tau^2}(\delta_{y_1y_i}-\delta_{y_2y_i})(\delta_{y_1y_j}-\delta_{y_2y_j})\zeta(x,y_1,y_2).
    \end{aligned}
\end{equation}
where $\zeta(x,y_1,y_2)=\sigma(\frac{1}{\tau}\log\frac{\pi_\theta(y_1|x)\pi_{\text{ref}}(y_2|x)}{\pi_\theta(y_2|x)\pi_{\text{ref}}(y_1|x)})\sigma(\frac{1}{\tau}\log\frac{\pi_\theta(y_2|x)\pi_{\text{ref}}(y_1|x)}{\pi_\theta(y_1|x)\pi_{\text{ref}}(y_2|x)})$.

Let $m_\theta(x,y_1,y_2)=\frac{\pi_\theta(y_1|x)\pi_{\text{ref}}(y_2|x)}{\pi_\theta(y_2|x)\pi_{\text{ref}}(y_1|x)}$ and . Because $\Omega_1(y_1,y_2,x)=\{|\log\frac{p^*(1|y_1,y_2,x)}{p^*(0|y_1,y_2,x)}|\geq\epsilon_0 \}$, $\Omega_2(y_1,y_2,x)=\{|\log m_\theta(x,y_1,y_2)|\geq\epsilon_0 \}$ and $\gamma(x)=\frac{\sum_{y_1,y_2\in\mathbb{Y}} \mathbb{I}(\Omega_1\cap\Omega_2)}{K^2}$.
\begin{equation}
\begin{aligned}
&|\psi(x)|=|\sum_{y_i,y_j\in\mathbb{Y}}z(x,y_i)  \frac{\partial^2 f_{\text{DPO}}(x,\theta)}{\partial \theta(x, y_i) \partial \theta(x, y_j)}  z(x,y_j)| \\
=&|\sum_{y_i,y_j\in\mathbb{Y}}z(x,y_i)  \sum_{y_1,y_2\in\mathbb{Y}} \frac{1}{K^2}\frac{\partial^2 h(\pi_\theta, x, y_1,y_2)}{\partial \theta(x, y_i) \partial \theta(x, y_j)}  z(x,y_j)| \\
= &|\sum_{y_i,y_j\in\mathbb{Y}}z(x,y_i)  \sum_{y_1,y_2\in\mathbb{Y}} \frac{1}{K^2}\frac{1}{\tau^2}(\delta_{y_1y_i}-\delta_{y_2y_i})(\delta_{y_1y_j}-\delta_{y_2y_j})\zeta(x,y_1,y_2)  z(x,y_j)| \\
=& |\sum_{y_i,y_2\in\mathbb{Y}}z(x,y_i)   \frac{1}{K^2}\zeta(x,y_i,y_2)\frac{1}{\tau^2}z(x,y_i) - \sum_{y_j,y_i\in\mathbb{Y}}z(x,y_i)\frac{1}{K^2}\zeta(x,y_i,y_j)\frac{1}{\tau^2}z(x,y_j) \\
&- \sum_{y_i,y_j\in\mathbb{Y}}z(x,y_i)   \frac{1}{K^2}\zeta(x,y_i,y_j)\frac{1}{\tau^2}z(x,y_j) + \sum_{y_1,y_j\in\mathbb{Y}}z(x,y_i)   \frac{1}{K^2}\zeta(x,y_1,y_j)\frac{1}{\tau^2}z(x,y_i)| \\
\leq& \frac{2}{\tau^2}|\sum_{y_i,y_2\in\mathbb{Y}}z(x,y_i)   \frac{1}{K^2}\zeta(x,y_i,y_2)z(x,y_i)|+\frac{2}{\tau^2}|\sum_{y_i,y_j\in\mathbb{Y}}z(x,y_i)   \frac{1}{K^2}\zeta(x,y_i,y_j)z(x,y_j)|\\
=& \frac{2}{\tau^2}|\sum_{y_i\in\mathbb{Y}}z^2(x,y_i)   \sum_{y_2\in\mathbb{Y}}\frac{1}{K^2}\zeta(x,y_i,y_2) |+\frac{2}{\tau^2}|\sum_{y_i,y_j\in\mathbb{Y}}z(x,y_i)   \frac{1}{K^2}\zeta(x,y_i,y_j)z(x,y_j)|.
\end{aligned}
\end{equation}

For the first term $\frac{2}{\tau^2}|\sum_{y_i\in\mathbb{Y}}z^2(x,y_i)   \sum_{y_2\in\mathbb{Y}}\frac{1}{K^2}\zeta(x,y_i,y_2) |$, based on Hölder's inequality we have:
\begin{equation}\label{data_proof_term1_eq}
    \begin{aligned}
&\frac{2}{\tau^2}|\sum_{y_i\in\mathbb{Y}}z^2(x,y_i)   \sum_{y_2\in\mathbb{Y}}\frac{1}{K^2}\zeta(x,y_i,y_2) | \leq \frac{2}{\tau^2}\sum_{y_i,y_2\in\mathbb{Y}}|\frac{1}{K^2}\zeta(x,y_i,y_2)|\cdot \|z^2(x,\cdot)\|_\infty \\
\leq& \frac{2}{\tau^2}(\gamma(x)\sigma(\frac{\epsilon_0}{\tau})\sigma(-\frac{\epsilon_0}{\tau}) + 1-\gamma(x) )\|z^2(x,\cdot)\|_\infty =  \frac{2}{\tau^2}(\gamma(x) c_0+1)\|z(x,\cdot)\|_\infty^2 \leq \frac{2}{\tau^2}(\gamma(x) c_0+1)\|z(x,\cdot)\|_2^2.
    \end{aligned}
\end{equation}

For the second term $\frac{2}{\tau^2}|\sum_{y_i,y_j\in\mathbb{Y}}z(x,y_i)   \frac{1}{K^2}\zeta(x,y_i,y_j)z(x,y_j)|$, let $\mathbb{Y}_1(x)=\{(y_1,y_2)| \mathbb{I}(\Omega_1(x,y_1,y_2)\cap\Omega_2(x,y_1,y_2) )=1 \}$ and $\mathbb{Y}_2(x)=\{(y_1,y_2)| \mathbb{I}(\Omega_1(x,y_1,y_2)\cap\Omega_2(x,y_1,y_2) )=0 \}$, we have:
\begin{equation}\label{data_proof_term2_eq}
    \begin{aligned}
&\frac{2}{\tau^2}|\sum_{y_i,y_j\in\mathbb{Y}}z(x,y_i)   \frac{1}{K^2}\zeta(x,y_i,y_j)z(x,y_j)| \\
\leq&  \frac{2}{\tau^2}|\sum_{(y_i,y_j)\in\mathbb{Y}_1}z(x,y_i)   \frac{1}{K^2}\zeta(x,y_i,y_j)z(x,y_j)| + \frac{2}{\tau^2}|\sum_{(y_i,y_j)\in\mathbb{Y}_2}z(x,y_i)   \frac{1}{K^2}\zeta(x,y_i,y_j)z(x,y_j)| \\
\leq& \frac{2}{\tau^2}|\sum_{(y_i,y_j)\in\mathbb{Y}_1}z(x,y_i)   \frac{1}{K^2}\sigma(\frac{\epsilon_0}{\tau})\sigma(-\frac{\epsilon_0}{\tau})z(x,y_j)| + \frac{2}{\tau^2}|\sum_{(y_i,y_j)\in\mathbb{Y}_2}z(x,y_i)   \frac{1}{K^2}z(x,y_j)| \\
\leq& \frac{2}{\tau^2}\sigma(\frac{\epsilon_0}{\tau})\sigma(-\frac{\epsilon_0}{\tau})\sqrt{\sum_{(y_i,y_j)\in\mathbb{Y}_1}z^2(x,y_i)\frac{1}{K^2}} \sqrt{\sum_{(y_i,y_j)\in\mathbb{Y}_1}\frac{1}{K^2}z^2(x,y_j)} + \frac{2}{\tau^2}|\sum_{(y_i,y_j)\in\mathbb{Y}_2}z(x,y_i)   \frac{1}{K^2}z(x,y_j)|  \\
\leq& \frac{2}{\tau^2}\sigma(\frac{\epsilon_0}{\tau})\sigma(-\frac{\epsilon_0}{\tau})\gamma(x)\|z^2(x,\cdot)\|_\infty + \frac{2}{\tau^2}|\sum_{(y_i,y_j)\in\mathbb{Y}_2}z(x,y_i)   \frac{1}{K^2}z(x,y_j)| \\
\leq& \frac{2}{\tau^2}\sigma(\frac{\epsilon_0}{\tau})\sigma(-\frac{\epsilon_0}{\tau})\gamma(x)\|z^2(x,\cdot)\|_\infty +  \frac{2}{\tau^2}\sqrt{\sum_{(y_i,y_j)\in\mathbb{Y}_2}z^2(x,y_i)\frac{1}{K^2}} \sqrt{\sum_{(y_i,y_j)\in\mathbb{Y}_2}\frac{1}{K^2}z^2(x,y_j)} \\
\leq& \frac{2}{\tau^2}\sigma(\frac{\epsilon_0}{\tau})\sigma(-\frac{\epsilon_0}{\tau})\gamma(x)\|z^2(x,\cdot)\|_\infty + \frac{2}{\tau^2}(1-\gamma(x))\|z^2(x,\cdot)\|_\infty \leq \frac{2}{\tau^2}(\gamma(x) c_0+1)\|z(x,\cdot)\|_2^2.
    \end{aligned}
\end{equation}
Because $c_0=\sigma(\frac{\epsilon_0}{\tau})\sigma(-\frac{\epsilon_0}{\tau})-1\in (-1,0)$.

Therefore,
\begin{equation}
    \begin{aligned}
\|\psi(\cdot)\|_\infty \leq \max_{x\in\mathbb{X}}\frac{4(\gamma(x) c_0+1)}{\tau^2}\|z(x,\cdot)\|_2^2 \leq \frac{4(\gamma c_0+1)}{\tau^2}\|z(\cdot,\cdot)\|_2^2.
    \end{aligned}
\end{equation}
Then $\mathcal{L}_{\mathrm{DPO}}(\pi_{\theta}; \pi_{\text {ref }})$ is $\frac{4(\gamma c_0+1)}{\tau^2}$-smooth is proved. Proof finished.

\subsection{Proof of Theorem \ref{data_select_coro}}\label{data_select_coro_proof}
\textbf{Theorem \ref{data_select_coro}:} Define joint conditional probability distribution $\pi_1(y_1,y_2|x)= \frac{\mu}{K^2}\ \text{if}\ \mathbb{I}(\Omega_1 \cap \Omega_2) = 1; \frac{1-\mu\gamma}{(1-\gamma)K^2} \text{else}$ where $\mu\in(0,1)$. Assume $\|g_t\|^2\leq G^2$. Given the Definition \ref{Softmax} for policy $\pi_\theta$ and the learning rate $\alpha_t$, for DPO algorithm, let $\gamma(x)\leq\gamma$, then:
\begin{equation}
\small{    \begin{aligned}
\min_{1\leq i\leq T}&||\nabla_\theta \mathcal{L}_{\mathrm{DPO}}(\pi_{\theta_i} ; \pi_{\text {ref }})||^2_2\leq \frac{\mathcal{L}_{\mathrm{DPO}}(\pi_{\theta_T})-\mathcal{L}_{\mathrm{DPO}}^*}{\sum_{t=1}^{T-1}\alpha_t}  + \frac{2(\mu\gamma c_0+1)G^2\sum_{t=1}^{T-1}\alpha_t^2}{\tau^2\sum_{t=1}^{T-1}\alpha_t}.
    \end{aligned}}
\end{equation}
where $\mathcal{L}_{\mathrm{DPO}}^*=\min_{\pi_\theta} \mathcal{L}_{\mathrm{DPO}}(\pi_{\theta} ; \pi_{\text {ref }}, \pi_0)$ and $c_0=\sigma(\frac{\epsilon_0}{\tau})\sigma(-\frac{\epsilon_0}{\tau})-1\in (-1,0)$. See $\mathcal{L}_{\mathrm{DPO}}(\pi_{\theta} ; \pi_{\text {ref }})$ on Eq.\ref{DPO_eq}.

\textbf{About $\Omega_1(y_1,y_2,x)$:} As the optimization progresses, $\pi_\theta$ gradually approaches $\bar{\pi}^\tau$, causing the proportion of the event $\Omega_1 \cap \Omega_2$ (denoted by $\gamma$) to increase (See Proposition \ref{pq_equal}), thereby reducing the upper bound in inequality \ref{data_select_coro_eq}. This leads to an increasingly faster convergence rate for DPO. Conversely, if small-margin pairs are selected as event $\Omega_1$ (i.e. $\Omega_1(y_1,y_2,x)=\{|\log\frac{p^*(1|y_1,y_2,x)}{p^*(0|y_1,y_2,x)}|\leq\epsilon_0 \}$), the convergence rate of DPO will slow down as the optimization advances and intensify the distribution shift problem cause $\Omega_1$ and $\Omega_2$ will more and more insistent as the optimization advances.

\textbf{Proof:} Based on Theorem \ref{SGD_theorem}, we only need to know $\mathcal{L}_{\mathrm{DPO}}(\pi_{\theta}; \pi_{\text {ref }})$ is $\frac{4(\mu\gamma c_0+1)}{\tau^2}$-smooth. This corollary shares the same proof process as Lemma \ref{data_select}. The only two differences are Eq.\ref{data_proof_term1_eq} and Eq.\ref{data_proof_term2_eq}. Their $\gamma(x)$ will be changed into $\mu\gamma(x)$. Then we will get the smooth coefficient of $\mathcal{L}_{\mathrm{DPO}}(\pi_{\theta}; \pi_{\text {ref }})$ is $\frac{4(\mu\gamma c_0+1)}{\tau^2}$. Proof finished.

\subsection{Proposition \ref{pq_equal}}\label{pq_equal_sec}
\begin{proposition}\label{pq_equal}
   \begin{equation}
       \lim_{\epsilon_0\rightarrow 0} P(|\log\frac{p^*(z=1)}{p^*(z=0)}|\geq \epsilon_0) - P(|\log\frac{\bar{\pi}^\tau(y_1|x)\pi_{\text{ref}}(y_2|x)}{\bar{\pi}^\tau(y_2|x)\pi_{\text{ref}}(y_1|x)}|\geq \epsilon_0) =0.
   \end{equation} 
\end{proposition}
\textbf{Proof:} comparison probability $p^*(1 \mid y_1, y_2, x)\propto r(x,y_1)-r(x,y_2)$, where it is identified that $r(x, y_1) > r(x, y_2)$, denoted as $y_1 \succ y_2$; similarly, $(x, y_2, y_1)$ is sampled with probability $p^*(0 \mid y_1, y_2, x)\propto r(x,y_2)-r(x,y_1)$, and is identified as $r(x, y_1) < r(x, y_2)$, denoted as $y_2 \succ y_1$. $\bar{\pi}^\tau(y|x)=\frac{\pi_{ref}(y|x)\exp(\tau r(x,y))}{Z'(x)}$. Then
\begin{equation}
\begin{aligned}
    \{|\log\frac{p^*(z=1)}{p^*(z=0)}|\geq \epsilon_0\}\ &\underrightarrow{\epsilon_0\rightarrow0}\ \{r(x,y_1)\geq r(x,y_2)\}, \\
    \{|\log\frac{\bar{\pi}^\tau(y_1|x)\pi_{\text{ref}}(y_2|x)}{\bar{\pi}^\tau(y_2|x)\pi_{\text{ref}}(y_1|x)}|\geq \epsilon_0\}\ &\underrightarrow{\epsilon_0\rightarrow0}\ \{r(x,y_1)\geq r(x,y_2)\}.
\end{aligned}
\end{equation}
Proof finished.

As the optimization progresses, $\pi_\theta$ gradually approaches $\bar{\pi}^\tau$, causing $\Omega_1(y_1,y_2,x)$ close to $\Omega_2(y_1,y_2,x)$, then the proportion of the event $\Omega_1 \cap \Omega_2$ (denoted by $\gamma$) increases. Conversely, if small-margin pairs are selected as event $\Omega_1$ (i.e. $\Omega_1(y_1,y_2,x)=\{|\log\frac{p^*(1|y_1,y_2,x)}{p^*(0|y_1,y_2,x)}|\leq\epsilon_0 \}$), the convergence rate of DPO will slow down as the optimization advances and intensify the distribution shift problem cause $\Omega_1$ and $\Omega_2$ will more and more insistent as the optimization advances.

\subsection{Sub Lemma for Lemma \ref{BDA_L_RKL_coef}-\ref{PRA_L_coef}}\label{SubLemma_sec}
\begin{lemma}\label{softmax_1derivative}
    By the properties of the softmax derivative, we have: 
\begin{equation}
    \frac{\partial \pi_\theta(y|x)}{\partial \theta(x,y')} = \pi_\theta(y|x) (\delta_{yy'} - \pi_\theta(y'|x)),\ \text{where}\ \delta_{yy'}= \begin{cases}1, & \text { if } y=y' \\ 0, & \text { otherwise }\end{cases}.
\end{equation}
\end{lemma}
\begin{lemma}\label{spectral_radius}
    $\forall x$, $x\rightarrow f(x)$ is $L$-smooth is equivalent to the following property:
    \begin{equation}
\left|\sum_{i,j=1}^{|\mathbb{X}|}x_i \frac{\partial^2 f(x)}{\partial x_i\partial x_j} x_j\right|\leq L||\vec{x}||_2^2.
\end{equation}
By Taylor's theorem, it suffices to show that the spectral radius of the hessian matrix of the second derivative of $f(x)$ is bounded by $L$.
\end{lemma}
\begin{lemma}\label{Pair_second_derivative_lemma}
$f(x,\theta)=\sum_{y_1,y_2\in\mathbb{Y}}\pi_\theta(y_1|x)\pi_\theta(y_2|x)h(\pi_\theta,x,y_1,y_2)$. Then the second derivative of $f(x,\theta)$ is:
\begin{equation}
\small{    \begin{aligned}
        \frac{\partial^2 f(x, \theta)}{\partial \theta(x, y_i) \partial \theta(x, y_j)} = \sum_{y_1,y_2\in\mathbb{Y}}& 2\frac{\partial^2 \pi_\theta(y_1|x)}{\partial \theta(x, y_i) \partial \theta(x, y_j)} \pi_\theta(y_2|x)h(\pi_\theta,x,y_1,y_2) + 2\frac{\partial \pi_\theta(y_1|x)}{\partial \theta(x, y_i)}\frac{\partial \pi_\theta(y_2|x)}{\partial \theta(x, y_j)}h(\pi_\theta,x,y_1,y_2) \\
        &+ 2\frac{\partial \pi_\theta(y_1|x)}{\partial \theta(x, y_i)}\pi_\theta(y_2|x) \frac{\partial h(\pi_\theta, x, y_1,y_2)}{\partial \theta(x, y_j)}+ 2\frac{\partial \pi_\theta(y_1|x)}{\partial \theta(x, y_j)}\pi_\theta(y_2|x) \frac{\partial h(\pi_\theta, x, y_1,y_2)}{\partial \theta(x, y_i)}\\
        & + \pi_\theta(y_1|x)\pi_\theta(y_2|x)\frac{\partial^2 h(\pi_\theta, x, y_1,y_2)}{\partial \theta(x, y_i) \partial \theta(x, y_j)}.
    \end{aligned} }
\end{equation}
\end{lemma}
\begin{lemma}\label{SelfEntropy}
    For $f(p)=-p\log p-(1-p)\log(1-p)$, $f(p)\leq\log2$.

    Proof: For $f(p)=-p\log p-(1-p)\log(1-p)\geq 0, p\in[0,1]$, $f'(p)=\log(\frac{1}{p}-1)$ is a monotonically decreasing function that reaches 0 when $p=0.5$. Thus $f(p)$ increases first and then decreases, and reaches its maximum value at $p=0.5$. So $f(p)\leq\log2$.
\end{lemma}

\begin{lemma}\label{TV_KL_inequal}
$D_{\mathrm{TV}}(p \| q)^2 \leq D_{\mathrm{KL}}(p \| q)$.
\end{lemma}
\subsection{Proof of Lemma \ref{BDA_L_RKL_coef}}\label{BDA_L_RKL_coef_proof}
\begin{lemma}\label{BDA_L_RKL_coef}
    (BDA(Reverse-KL) Smoothness ) Given softmax parametrization of Definition \ref{Softmax} for policy $\pi_\theta$,  $\forall r, \tau$, $\theta\rightarrow \mathbb{E}_{x\sim \mathcal{D}}\left[\mathrm{D}_{\mathrm{KL}}\left( \pi^\tau\left(\cdot \mid x\right) \| \pi_\theta\left(\cdot \mid x\right) \right)\right]$ is 2-smooth (see Eq.\ref{reverse-KL}).

    See proof in Appendix \ref{BDA_L_RKL_coef_proof}.
\end{lemma}

\textbf{Proof:} By Lemma \ref{spectral_radius}, it suffices to show that the spectral radius of the hessian matrix of the second derivative of $\mathbb{E}_{x\sim \mathcal{D}}\left[\mathrm{D}_{\mathrm{KL}}\left( \pi^\tau\left(\cdot \mid x\right) \| \pi_\theta\left(\cdot \mid x\right) \right)\right]$ is bounded by $2$, i.e.
\begin{equation}
\left|\sum_{x,x'\in\mathbb{X}}\sum_{y,y'\in\mathbb{Y}}z(x,y) \frac{\partial^2 \mathbb{E}_{x\sim \mathcal{D}}\left[\mathrm{D}_{\mathrm{KL}}\left( \pi^\tau\left(\cdot \mid x\right) \| \pi_\theta\left(\cdot \mid x\right) \right)\right]}{\partial \theta(x,y)\partial\theta(x',y')} z(x',y')\right|\leq2||z(\cdot,\cdot)||_2^2.
\end{equation}
By Lemma \ref{softmax_1derivative}, the first derivative of $\mathbb{E}_{x\sim \mathcal{D}}\left[\mathrm{D}_{\mathrm{KL}}\left( \pi^\tau\left(\cdot \mid x\right) \| \pi_\theta\left(\cdot \mid x\right) \right)\right]$ is:
\begin{equation}
    \begin{aligned}
& \frac{\partial \mathbb{E}_{x\sim \mathcal{D}}\left[\mathrm{D}_{\mathrm{KL}}\left( \pi^\tau\left(\cdot \mid x\right) \| \pi_\theta\left(\cdot \mid x\right) \right)\right]}{\partial \theta(x,y)}  \\
=& -\frac{\partial \mathbb{E}_{x\sim \mathcal{D}}[\sum_{y'\in Y}\frac{\exp(\tau r(x,y'))}{Z(x)}(\log\pi_\theta(y'|x))]}{\partial \theta(x,y)} \\
=&-\mathcal{D}(x)\sum_{y'\in Y}\frac{\exp(\tau r(x,y'))}{Z(x)}\frac{\partial \log\pi_\theta(y'|x))}{\partial \theta(x,y)} \\
=&-\mathcal{D}(x)\sum_{y'\in Y}\frac{\exp(\tau r(x,y'))}{Z(x)} (\delta_{y'y} - \pi_\theta(y|x))\\
=& \mathcal{D}(x)\pi_\theta(y|x) - \mathcal{D}(x)\frac{\exp(\tau r(x,y))}{Z(x)}.
\end{aligned}
\end{equation}
The second derivative of $\mathbb{E}_{x\sim \mathcal{D}}\left[\mathrm{D}_{\mathrm{KL}}\left( \pi^\tau\left(\cdot \mid x\right) \| \pi_\theta\left(\cdot \mid x\right) \right)\right]$ is:
\begin{equation}
    \begin{aligned}
        & \frac{\partial^2 \mathbb{E}_{x\sim \mathcal{D}}\left[\mathrm{D}_{\mathrm{KL}}\left( \pi^\tau\left(\cdot \mid x\right) \| \pi_\theta\left(\cdot \mid x\right) \right)\right]}{\partial \theta(x,y)\partial \theta(x',y')}  \\
        =& \frac{\partial}{\partial \theta(x',y')}(\frac{\partial \mathbb{E}_{x\sim \mathcal{D}}\left[\mathrm{D}_{\mathrm{KL}}\left( \pi^\tau\left(\cdot \mid x\right) \| \pi_\theta\left(\cdot \mid x\right) \right)\right]}{\partial \theta(x,y)})\\
        =& \frac{\partial \mathcal{D}(x)\pi_\theta(y|x)}{\partial \theta(x',y')} = \delta_{xx'}\mathcal{D}(x)\pi_\theta(y|x)(\delta_{yy'} - \pi_\theta(y'|x)).
    \end{aligned}
\end{equation}
Then the spectral radius of $\mathbb{E}_{x\sim \mathcal{D}}\left[\mathrm{D}_{\mathrm{KL}}\left( \pi^\tau\left(\cdot \mid x\right) \| \pi_\theta\left(\cdot \mid x\right) \right)\right]$ is:
\begin{equation}
\begin{aligned}
&\left|\sum_{x,x'\in\mathbb{X}}\sum_{y,y'\in\mathbb{Y}}z(x,y) \frac{\partial^2 \mathbb{E}_{x\sim \mathcal{D}}\left[\mathrm{D}_{\mathrm{KL}}\left( \pi^\tau\left(\cdot \mid x\right) \| \pi_\theta\left(\cdot \mid x\right) \right)\right]}{\partial \theta(x,y)\partial\theta(x',y')} z(x',y')\right|\\
=&\left|\sum_{x\in\mathbb{X}}\sum_{y,y'\in\mathbb{Y}}z(x,y) \delta_{xx'}\mathcal{D}(x)\pi_\theta(y|x)(\delta_{yy'} - \pi_\theta(y'|x)) z(x,y')\right| \\
=&\left|\sum_{x\in\mathbb{X}}\mathcal{D}(x)\sum_{y,y'\in\mathbb{Y}}z(x,y)  \pi_\theta(y|x)(\delta_{yy'} - \pi_\theta(y'|x))  z(x,y')\right|\\
\triangleq& |\sum_{x\in\mathbb{X}}\mathcal{D}(x) \psi(x) |\leq \|\mathcal{D}(\cdot)\|_1 \|\psi(\cdot)\|_\infty=1\cdot\|\psi(\cdot)\|_\infty.
\end{aligned}
\end{equation}
Then we have:
\begin{equation}
\begin{aligned}
    &\|\psi(\cdot)\|_\infty = \max_{x\in\mathbb{X}}\left| \sum_{y,y'\in\mathbb{Y}}z(x,y)  \pi_\theta(y|x)(\delta_{yy'} - \pi_\theta(y'|x))  z(x,y')\right| \\
    =& \max_{x\in\mathbb{X}}\left| \sum_{y,y'\in\mathbb{Y}}z(x,y)  \pi_\theta(y|x)\delta_{yy'}z(x,y') - z(x,y)\pi_\theta(y|x)\pi_\theta(y'|x)  z(x,y')\right| \\
    \leq& \max_{x\in\mathbb{X}}\left| \sum_{y,y'\in\mathbb{Y}}z(x,y)  \pi_\theta(y|x)\delta_{yy'}z(x,y')\right|+ \max_{x\in\mathbb{X}}\left| \sum_{y,y'\in\mathbb{Y}}z(x,y)\pi_\theta(y|x)\pi_\theta(y'|x)  z(x,y')\right| \\
    =& \max_{x\in\mathbb{X}}\left| \sum_{y,y'\in\mathbb{Y}}z(x,y)  \pi_\theta(y|x)z(x,y)\right|+ \max_{x\in\mathbb{X}} (\sum_{y\in\mathbb{Y}}z(x,y)\pi_\theta(y|x))^2\\
    \leq& \max_{x\in\mathbb{X}} \sum_{y\in\mathbb{Y}}z(x,y)z(x,y) + \max_{x\in\mathbb{X}}(z(x,\cdot)^T\pi_\theta(\cdot|x))^2 \\
    \leq& \max_{x\in\mathbb{X}}  ||z(x,\cdot)||_2^2+ \max_{x\in\mathbb{X}}(||\pi_\theta(\cdot|x)||_1 ||z(x,\cdot)||_\infty )^2 \\
    \leq& 2\max_{x\in\mathbb{X}}  ||z(x,\cdot)||_2^2 = 2 \|||z(x,\cdot)||_2^2\|_\infty \leq 2||z(\cdot,\cdot)||_2^2.
\end{aligned}
\end{equation}
where $\|||z(x,\cdot)||_2^2\|_\infty$ is taking the infinite norm of $||z(x,\cdot)||_2^2$ with $x$ as the coordinate axis. 

Therefore, we have the $2$-smooth conclusion:
\begin{equation}
\left|\sum_{x,x'\in\mathbb{X}}\sum_{y,y'\in\mathbb{Y}}z(x,y) \frac{\partial^2 \mathbb{E}_{x\sim \mathcal{D}}\left[\mathrm{D}_{\mathrm{KL}}\left( \pi^\tau\left(\cdot \mid x\right) \| \pi_\theta\left(\cdot \mid x\right) \right)\right]}{\partial \theta(x,y)\partial\theta(x',y')} z(x',y')\right|\leq2||z(\cdot,\cdot)||_2^2.
\end{equation}
Proof finished.

\subsection{Proof of Lemma \ref{BDA_L_FKL_coef}}\label{BDA_L_FKL_coef_proof}
\begin{lemma}\label{BDA_L_FKL_coef}
    (BDA(Forward-KL) Smoothness ) Given softmax parametrization of Definition \ref{Softmax} for policy $\pi_\theta$, assume $|\log(\pi_\theta(y|x)) - \log(\pi^\tau(y|x))|\leq\epsilon_1$,  $\forall r, \tau$, $\theta\rightarrow \mathbb{E}_{x\sim \mathcal{D}}\left[\mathrm{D}_{\mathrm{KL}}\left(\pi_\theta\left(\cdot \mid x\right) \| \pi^\tau\left(\cdot \mid x\right)\right)\right]$ is $\left((4+K)\epsilon_1+6+2K\right)$-smooth (see Eq.\ref{forward-KL}).

    See proof in Appendix \ref{BDA_L_FKL_coef_proof}.
\end{lemma}

\textbf{Proof:} By Lemma \ref{spectral_radius}, it suffices to show that the spectral radius of the hessian matrix of the second derivative of $\mathbb{E}_{x\sim \mathcal{D}}\left[\mathrm{D}_{\mathrm{KL}}\left(\pi_\theta\left(\cdot \mid x\right) \| \pi^\tau\left(\cdot \mid x\right)\right)\right]$ is bounded by $\left(6\epsilon_1+10\right)$, i.e.
\begin{equation}
\left|\sum_{x,x'\in\mathbb{X}}\sum_{y,y'\in\mathbb{Y}}z(x,y) \frac{\partial^2 \mathbb{E}_{x\sim \mathcal{D}}\left[\mathrm{D}_{\mathrm{KL}}\left(\pi_\theta\left(\cdot \mid x\right) \| \pi^\tau\left(\cdot \mid x\right)\right)\right]}{\partial \theta(x,y)\partial\theta(x',y')} z(x',y')\right|\leq\left(6\epsilon_1+10\right)||z(\cdot,\cdot)||_2^2.
\end{equation}
For Equation \ref{forward-KL}, we have:
\begin{equation}
\begin{aligned}
&\mathbb{E}_{x\sim \mathcal{D}}\left[\mathrm{D}_{\mathrm{KL}}\left(\pi_\theta\left(\cdot \mid x\right) \| \pi^\tau\left(\cdot \mid x\right)\right)\right]\\ 
=&-\mathbb{E}_{x\sim \mathcal{D},y\sim\pi_\theta(\cdot|x)}\left[(\log(\pi^\tau(y|x) - \log(\pi_\theta(y|x)))\right]\\
=&\sum_{x\in\mathbb{X}}\mathcal{D}(x)\pi_\theta(\cdot|x)^Th(\pi_\theta,x)\quad \text{where}\ h(\pi_\theta,x,y)=(-\log(\pi^\tau(y|x) + \log(\pi_\theta(y|x))),\ h(\pi_\theta,x)=h(\pi_\theta,x,\cdot).
\end{aligned}
\end{equation}
Denote $f_{\text{fKL}}(x,\theta) =\pi_\theta(\cdot|x)^Th(\pi_\theta,x)$. We can calculate the spectral radius of $\frac{\partial^2 \mathbb{E}_{x\sim \mathcal{D}}\left[\mathrm{D}_{\mathrm{KL}}\left(\pi_\theta\left(\cdot \mid x\right) \| \pi^\tau\left(\cdot \mid x\right)\right)\right]}{\partial \theta(x,y)\partial\theta(x',y')}$:
\begin{equation}
\begin{aligned}
&\left|\sum_{x,x'\in\mathbb{X}}\sum_{y,y'\in\mathbb{Y}}z(x,y) \frac{\partial^2 \mathbb{E}_{x\sim \mathcal{D}}\left[\mathrm{D}_{\mathrm{KL}}\left(\pi_\theta\left(\cdot \mid x\right) \| \pi^\tau\left(\cdot \mid x\right)\right)\right]}{\partial \theta(x,y)\partial\theta(x',y')} z(x',y')\right|\\
=&\left|\sum_{x\in\mathbb{X}}\sum_{y,y'\in\mathbb{Y}}z(x,y) \frac{\partial^2 \sum_{x\in\mathbb{X}}\mathcal{D}(x)\pi_\theta(\cdot|x)^Th(\pi_\theta,x)}{\partial \theta(x,y)\partial\theta(x,y')}  z(x,y')\right|\\ =&\left|\sum_{x\in\mathbb{X}}\sum_{y,y'\in\mathbb{Y}}z(x,y) \mathcal{D}(x) \frac{\partial^2 f_{\text{fKL}}(x,\theta)}{\partial \theta(x, y) \partial \theta(x, y')}  z(x,y')\right|.
\end{aligned}
\end{equation}
The first derivative of $f_{\text{fKL}}(x,\theta)$ is:
\begin{equation}
    \frac{\partial f_{\text{fKL}}(x, \theta)}{\partial \theta(x, y_i)} = \sum_{y} \frac{\partial \pi_\theta(y|x)}{\partial \theta(x, y_i)} h(\pi_\theta, x, y) + \pi_\theta(y|x) \frac{\partial h(\pi_\theta, x, y)}{\partial \theta(x, y_i)}.
\end{equation}

The second derivative of $f_{\text{fKL}}(x,\theta)$ is:
\begin{equation}
    \begin{aligned}
        \frac{\partial^2 f_{\text{fKL}}(x, \theta)}{\partial \theta(x, y_i) \partial \theta(x, y_j)} = \sum_{y}& \frac{\partial^2 \pi_\theta(y|x)}{\partial \theta(x, y_i) \partial \theta(x, y_j)} h(\pi_\theta, x, y) + \frac{\partial \pi_\theta(y|x)}{\partial \theta(x, y_i)} \frac{\partial h(\pi_\theta, x, y)}{\partial \theta(x, y_j)}\\
        &+ \frac{\partial \pi_\theta(y|x)}{\partial \theta(x, y_j)} \frac{\partial h(\pi_\theta, x, y)}{\partial \theta(x, y_i)} + \pi_\theta(y|x) \frac{\partial^2 h(\pi_\theta, x, y)}{\partial \theta(x, y_i) \partial \theta(x, y_j)}.
    \end{aligned}
\end{equation}
By Lemma \ref{softmax_1derivative}, we have:
\begin{equation}
    \frac{\partial \pi_\theta(y|x)}{\partial \theta(x, y_i)} = \pi_\theta(y|x)(\delta_{yy_i}-\pi_\theta(y_i|x)).
\end{equation}
\begin{equation}
    \frac{\partial\left(\pi_\theta(y | x)(\delta_{yy_i} - \pi_\theta(y_i | x))\right)}{\partial \theta(x, y_j)} = \pi_\theta(y | x)(\delta_{yy_j} - \pi_\theta(y_j | x))(\delta_{yy_i} - \pi_\theta(y_i | x)) - \pi_\theta(y | x) \pi_\theta(y_i | x)(\delta_{y_iy_j} - \pi_\theta(y_j | x)).
\end{equation}
And:
\begin{equation}
    \frac{\partial h(\pi_\theta, x, y)}{\partial \theta(x, y_j)} = \frac{\partial (-\log(\pi^\tau(y|x) + \log(\pi_\theta(y|x)))}{\partial \theta(x, y_j)} = (\delta_{yy_j}-\pi_\theta(y_j|x)).
\end{equation}
\begin{equation}
    \frac{\partial^2 h(\pi_\theta, x, y)}{\partial \theta(x, y_i) \partial \theta(x, y_j)} = \frac{\partial}{\partial \theta(x, y_j)}\left( \frac{\partial h(\pi_\theta, x, y)}{\partial \theta(x, y_i)} \right) = \frac{\partial (\delta_{yy_i}-\pi_\theta(y_i|x))}{\partial \theta(x, y_j)}=-\pi_\theta(y_i|x)(\delta_{y_iy_j}-\pi_\theta(y_j|x)).
\end{equation}
Thus the second derivative of $f_{\text{fKL}}(x,\theta)$ is changed into:
\begin{equation}
    \begin{aligned}
        &\frac{\partial^2 f_{\text{fKL}}(x, \theta)}{\partial \theta(x, y_i) \partial \theta(x, y_j)} \\
        =&\sum_{y} \pi_\theta(y | x)(\delta_{yy_j} - \pi_\theta(y_j | x))(\delta_{yy_i} - \pi_\theta(y_i | x))(h(\pi_\theta, x, y) +2) \\
        &\quad\quad - \pi_\theta(y | x) \pi_\theta(y_i | x)(\delta_{y_iy_j} - \pi_\theta(y_j | x))(h(\pi_\theta, x, y) + 1). \\
    \end{aligned}
\end{equation}
Let $\vec{h}(\theta,x) = (h(\pi_\theta, x, y_1), h(\pi_\theta, x, y_2), \ldots, h(\pi_\theta, x, y_{|\mathbb{Y}|}))$, $\vec{H}_1(\theta,x) =-\vec{h}(\theta,x) - 1$, and $\vec{H}_2(\theta,x) = \vec{h}(\theta,x) + 2 $. So ${H}_1(\pi_\theta, x, y)=-h(\pi_\theta, x, y)-1$ and $H_2(\pi_\theta, x, y)$ is similar. Then, the Hessian matrix of $f_{\text{fKL}}(x,\theta) = \pi_\theta(\cdot|x)^Th(\pi_\theta,x)$ simplifies as follows:
\begin{equation}
    \begin{aligned}
\nabla^2_\theta f_{\text{fKL}}(x,\theta)_{i,j} = \sum_{y \in \mathbb{Y}} [&\pi_\theta(y | x) \pi_\theta(y_i | x)(\delta_{y_iy_j} - \pi_\theta(y_j | x)) {H}_1(\pi_\theta, x, y) \\
&+ \pi_\theta(y | x)(\delta_{yy_j} - \pi_\theta(y_j | x))(\delta_{yy_i} - \pi_\theta(y_i | x)) {H}_2(\pi_\theta, x, y)].
\end{aligned}
\end{equation}

Then we compute the quadratic form of the Hessian matrix:
\begin{equation}
    \small{\begin{aligned}
&|\sum_{x\in\mathbb{X}}\sum_{y,y'\in\mathbb{Y}}z(x,y) \mathcal{D}(x) \frac{\partial^2 f_{\text{fKL}}(x,\theta)}{\partial \theta(x, y) \partial \theta(x, y')}  z(x,y')| \\
=& |\sum_{x\in\mathbb{X}}\mathcal{D}(x) \sum_{i=1}^{|\mathbb{Y}|} \sum_{j=1}^{|\mathbb{Y}|} z(x,y_i) \left(\nabla^2_\theta f_{\text{fKL}}(x,\theta)_{i,j} \right) z(x,y_j)| \\
=& |\sum_{x\in\mathbb{X}}\mathcal{D}(x) ( \pi_\theta(\cdot|x)^\top \vec{H}_1(\theta,x) (\pi_\theta(\cdot|x)^\top z(x,\cdot))^2 + \pi_\theta(\cdot|x)^\top (\vec{H}_2(\theta,x) - \vec{H}_1(\theta,x)) (\pi_\theta(\cdot|x)^\top z(x,\cdot))^2 \\
&\quad + \sum_{i=1}^K z(x,y_i)^2 \pi_\theta(y_i | x) H_2(\pi_\theta, x, y_i) - 2 \pi_\theta(\cdot|x)^\top z(x,\cdot) \pi_\theta(y_i | x) H_2(\pi_\theta, x, y_i) z(x,y_i) ) |\\
=& | \sum_{x\in\mathbb{X}}\mathcal{D}(x) ( \pi_\theta(\cdot|x)^\top \vec{H}_1(\theta,x) (\pi_\theta(\cdot|x)^\top z(x,\cdot))^2 + \pi_\theta(\cdot|x)^\top (\vec{H}_2(\theta,x) - \vec{H}_1(\theta,x)) (\pi_\theta(\cdot|x)^\top z(x,\cdot))^2 \\
&\quad+ \sum_{i=1}^K (z(x,y_i) - 2 \pi_\theta(\cdot|x)^\top z(x,\cdot)) \pi_\theta(y_i | x) H_2(\pi_\theta, x, y_i) z(x,y_i) ) | \\
\triangleq& |\sum_{x\in\mathbb{X}}\mathcal{D}(x) \psi(x) |\leq \|\mathcal{D}(\cdot)\|_1 \|\psi(\cdot)\|_\infty=1\cdot\|\psi(\cdot)\|_\infty.
\end{aligned}}
\end{equation}

To establish an upper bound for the spectral radius of the Hessian matrix, let $\vec{H}_3(\theta,x) = \vec{H}_2(\theta,x) - \vec{H}_1(\theta,x)$. We can then write:
\begin{equation}
    \begin{aligned}
& \|\psi(\cdot)\|_\infty \\ 
=& \max_{x\in\mathbb{X}}\left|\pi_\theta(\cdot|x)^\top \vec{H}_1(\theta,x) \cdot (\pi_\theta(\cdot|x)^\top z(x,\cdot))^2 + \pi_\theta(\cdot|x)^\top \vec{H}_3(\theta,x) \cdot (\pi_\theta(\cdot|x)^\top z(x,\cdot) )^2 \right.\\
&\quad+ \left. \sum_{i=1}^K z(x,y_i)^2 \pi_\theta(y_i | x) H_2(\pi_\theta, x, y_i) + \sum_{i=1}^K 2 \pi_\theta(\cdot|x)^\top z(x,\cdot) \pi_\theta(y_i | x) H_2(\pi_\theta, x, y_i) z(x,y_i) \right| \\
\leq& \max_{x\in\mathbb{X}}\left|\pi_\theta(\cdot|x)^\top \vec{H}_1(\theta,x) \cdot (\pi_\theta(\cdot|x)^\top z(x,\cdot))^2\right| + \left|\pi_\theta(\cdot|x)^\top \vec{H}_3(\theta,x) \cdot (\pi_\theta(\cdot|x)^\top z(x,\cdot))^2\right| \\
&\quad+ \left|\sum_{i=1}^K z(x,y_i)^2 \pi_\theta(y_i | x) H_2(\pi_\theta, x, y_i)\right| + \left|\sum_{i=1}^K 2 \pi_\theta(\cdot|x)^\top z(x,\cdot) \pi_\theta(y_i | x) H_2(\pi_\theta, x, y_i) z(x,y_i)\right| ,
\end{aligned}
\end{equation}
Continuing from the previous equation:
\begin{equation}
    \begin{aligned}
& \|\psi(\cdot)\|_\infty \\ 
\leq& \max_{x\in\mathbb{X}}\|\pi_\theta(\cdot|x)\|_1 \|\vec{H}_1(\theta,x)\|_\infty \|z(x,\cdot)\|_2^2 + \|\pi_\theta(\cdot|x)\|_1 \|\vec{H}_3(\theta,x)\|_\infty \|z(x,\cdot)\|_2^2 \\
&\quad+ \|\vec{H}_2(\theta,x)\|_\infty \|z(x,\cdot)\|_2^2 + 2\|\pi_\theta(\cdot|x)\odot\vec{H}_2(\theta,x)\|_1 \|z(x,\cdot)\|_2^2 \\
=& \max_{x\in\mathbb{X}}(\|\vec{H}_1(\theta,x)\|_\infty + \|\vec{H}_2(\theta,x)\|_\infty + \|\vec{H}_3(\theta,x)\|_\infty) \|z(x,\cdot)\|_2^2 + 2\|\pi_\theta(\cdot|x)\odot\vec{H}_2(\theta,x)\|_1 \|z(x,\cdot)\|_2^2 \\
\leq& \max_{x\in\mathbb{X}}(\|\vec{H}_1(\theta,x)\|_\infty + \|\vec{H}_2(\theta,x)\|_\infty + \|\vec{H}_3(\theta,x)\|_\infty) \|z(x,\cdot)\|_2^2 + 2\|\vec{H}_2(\theta,x)\|_\infty \|z(x,\cdot)\|_2^2 \\
=& \max_{x\in\mathbb{X}}(\|\vec{H}_1(\theta,x)\|_\infty + 3\|\vec{H}_2(\theta,x)\|_\infty + \|\vec{H}_3(\theta,x)\|_\infty) \|z(x,\cdot)\|_2^2\\
\leq& \max_{x\in\mathbb{X}}(\|\vec{H}_1(\theta,x)\|_\infty + 3\|\vec{H}_2(\theta,x)\|_\infty + \|\vec{H}_3(\theta,x)\|_\infty) \max_{x\in\mathbb{X}}\|z(x,\cdot)\|_2^2 \\
\leq& \max_{x\in\mathbb{X}}(\|\vec{H}_1(\theta,x)\|_\infty + 3\|\vec{H}_2(\theta,x)\|_\infty + \|\vec{H}_3(\theta,x)\|_\infty) \|z(\cdot,\cdot)\|_2^2 .
\end{aligned}
\end{equation}

In the second inequality, the first term arises from Hölder's inequality, which states that $ \pi_\theta^\top z(x,\cdot) \leq \|\pi_\theta\|_1 \|z(x,\cdot)\|_\infty = \|z(x,\cdot)\|_\infty $, where $ \|\pi_\theta\|_1 = 1 $ and $ \|z^2(x,\cdot)\|_1 = \|z(x,\cdot)\|_2^2 $. For the second term, we have the bound on $ |(\pi_\theta^\top z(x,\cdot))^2| $:
\begin{equation}
    |(\pi_\theta^\top z(x,\cdot))^2| = \pi_\theta^\top z(x,\cdot) \cdot \pi_\theta^\top z(x,\cdot) \leq \|\pi_\theta\|_1 \|z(x,\cdot)\|_\infty \cdot \|\pi_\theta\|_1 \|z(x,\cdot)\|_\infty \leq \|z(x,\cdot)\|_2^2.
\end{equation}

For the third term:
\begin{equation}
    \left| \sum_{i=1}^K z(x,y_i)^2 \pi_\theta(y_i | x) H_2(\pi_\theta, x, y_i) \right| \leq \left| \sum_{i=1}^K z(x,y_i)^2 |H_2(\pi_\theta, x, y_i)| \right| \leq \|\vec{H}_2(\theta,x)\|_\infty \|z(x,\cdot)\|_2^2.
\end{equation}

For the fourth term:
\begin{equation}
    \begin{aligned}
|\sum_{i=1}^K 2\pi_\theta^Tz(x,\cdot)\pi_\theta(y_i|x)H_2(\pi_\theta, x, y_i)z(x,y_i)|&\leq2||\pi_\theta||_1||z(x,\cdot)||_\infty|\sum_{i=1}^K \pi_\theta(y_i|x)H_2(\pi_\theta, x, y_i)z(x,y_i)||\\
&\leq2||z(x,\cdot)||_2||\pi_\theta(\cdot|x)\odot\vec{H}_2(\theta,x)||_1||z(x,\cdot)||_\infty\\
&\leq2||\vec{H}_2(\theta,x)||_\infty||z(x,\cdot)||_2^2.
\end{aligned}
\end{equation}
In this Lemma, we assume $|h(\pi_\theta,x,y)|=|\log(\pi_\theta(y|x)) - \log(\pi^\tau(y|x))|\leq\epsilon_1$ where $ \epsilon_1 $ denotes the upper bound of the error of the current parameterized distribution $ \pi_\theta $.

For $ ||\vec{H}_1(\theta,x)||_\infty $, we have:
\begin{equation}
    ||\vec{H}_1(\theta,x)||_\infty = || -\vec{h}(\theta,x) - 1 ||_\infty \leq \epsilon_1 + 1.
\end{equation}

For $ ||\vec{H}_2(\theta,x)||_\infty $, it follows that:
\begin{equation}
    ||\vec{H}_2(\theta,x)||_\infty = ||\vec{h}(\theta,x) + 2||_\infty \leq \epsilon_1 + 2.
\end{equation}

For $ ||\vec{H}_3(\theta,x)||_\infty $, we find:
\begin{equation}
    ||\vec{H}_3(\theta,x)||_\infty = ||\vec{H}_2(\theta,x) - \vec{H}_1(\theta)||_\infty = ||\vec{h}(\theta,x) + 2 + \vec{h}(\theta,x) + 1||_\infty \leq 2\epsilon_1 + 3.
\end{equation}

Combining these results yields:
\begin{equation}
    \begin{aligned}
&||\vec{H}_1(\theta)||_\infty + ||\vec{H}_2(\theta)||_\infty + ||\vec{H}_3(\theta)||_\infty \leq 6\epsilon_1+10.
\end{aligned}
\end{equation}
Thus,
\begin{equation}
\left|\sum_{x,x'\in\mathbb{X}}\sum_{y,y'\in\mathbb{Y}}z(x,y) \frac{\partial^2 \mathbb{E}_{x\sim \mathcal{D}}\left[\mathrm{D}_{\mathrm{KL}}\left(\pi_\theta\left(\cdot \mid x\right) \| \pi^\tau\left(\cdot \mid x\right)\right)\right]}{\partial \theta(x,y)\partial\theta(x',y')} z(x',y')\right|\leq\left(6\epsilon_1+10\right)||z(\cdot,\cdot)||_2^2.
\end{equation}
Proof finished.

\subsection{Proof of Lemma \ref{RA_L_coef}}\label{RA_L_coef_proof}
\begin{lemma}\label{RA_L_coef}
(RA Smoothness) Given softmax parametrization of Definition \ref{Softmax} for policy $\pi_\theta$, assume $|r_\theta(x,y)- r(x,y)|\leq\epsilon_2$,  $\forall r, \tau$, $\theta\rightarrow \mathbb{E}_{x\sim \mathcal{D},y\sim\pi_\theta(\cdot|x)}\left[\left(r_\theta(x,y)- r(x,y)\right)^2\right]$ is $( 3\epsilon_1^2 + \frac{18\epsilon_1}{\tau} + \frac{8}{\tau^2} + \max\left\{ \epsilon_1^2 + \frac{2}{\tau} \epsilon_1, \frac{1}{\tau} \right\} )$-smooth (see Eq.\ref{RA_eq}). See proof in Appendix \ref{RA_L_coef_proof}.

\end{lemma}

\textbf{Proof:} Denote $\mathcal{L}_{RA}(\pi_\theta)=\mathbb{E}_{x\sim \mathcal{D},y\sim\pi_\theta(\cdot|x)}\left[\left(r_\theta(x,y)- r(x,y)\right)^2\right]$ Let $S\triangleq S(\theta) \in \mathbb{R}^{K \times K}$ be the second derivative of the value map $\theta \rightarrow \mathcal{L}_{RA}(\pi_\theta)$, i.e. $S(\theta,x_1,y_1,x_2,y_2)=\frac{\partial^2 \mathcal{L}_{RA}(\pi_\theta)}{\partial \theta(x_1,y_1)\partial\theta(x_2,y_2)}$. Denote $L_{\text{RA}}=( 3\epsilon_1^2 + \frac{18\epsilon_1}{\tau} + \frac{8}{\tau^2} + \max\left\{ \epsilon_1^2 + \frac{2}{\tau} \epsilon_1, \frac{1}{\tau} \right\} )$. By Lemma \ref{spectral_radius}, it suffices to show that the spectral radius of $S$ is bounded by $L_{\text{RA}}$. Because $r_\theta(x, y) = \frac{1}{\tau} \log(Z(x)\pi_\theta(y|x))$, denote $z(x,y)=\frac{1}{\tau}\log\sum_{y\in Y}e^{\tau r(x,y)}- r(x,y)$, then we have:
\begin{equation}
\begin{aligned}
&\mathbb{E}_{x\sim \mathcal{D},y\sim\pi_\theta(\cdot|x)}\left[\left(r_\theta(x,y)- r(x,y)\right)^2\right]\\
=&\mathbb{E}_{x\sim \mathcal{D},y\sim\pi_\theta(y|x)}\left[\left(\frac{1}{\tau}\log\pi_\theta(y)+z(x,y)\right)^2\right] \\
=&\sum_{x\in\mathbb{X},y\in\mathbb{Y}}\mathcal{D}(x)\pi_\theta(y|x)\left(\frac{1}{\tau}\log\pi_\theta(y|x)+z(x,y)\right)^2\\
=&\sum_{x\in\mathbb{X}}\mathcal{D}(x)\pi_\theta(\cdot|x)^Th(\pi_\theta,x)\quad \text{where}\ h(\pi_\theta,x,y)=\left(\frac{1}{\tau}\log\pi_\theta(y|x)+z(x,y)\right)^2,\ h(\pi_\theta,x)=h(\pi_\theta,x,\cdot).
\end{aligned}
\end{equation}

Now, by Definition \ref{Softmax} we have:
\begin{equation}
    \pi_\theta(y \mid x)=\frac{\exp \{\theta(x, y)\}}{\sum_{y^{\prime}} \exp \left\{\theta\left(x, y^{\prime}\right)\right\}}.
\end{equation}
By Lemma \ref{softmax_1derivative}, we have: 
\begin{equation}
    \frac{\partial \pi_\theta(y|x)}{\partial \theta(x,y')} = \pi_\theta(y|x) (\delta_{yy'} - \pi_\theta(y'|x)),\ \text{where}\ \delta_{yy'}= \begin{cases}1, & \text { if } y=y' \\ 0, & \text { otherwise }\end{cases}.
\end{equation}

If $x\neq x'$, then the second derivative of $\pi_\theta(y|x)$ is 0:
\begin{equation}
    \frac{\partial^2 \pi_\theta(y|x)}{\partial \theta(x,y_1)\partial\theta(x',y_2)} = \frac{\partial \pi_\theta(y|x) (\delta_{yy_1} - \pi_\theta(y_1|x))}{\partial \theta(x',y_2)}=0.
\end{equation}
Similarly, because $\pi_\theta(y|x)$ and $h(\pi_\theta,x)$ don't have relevant parameters to $\theta(x',\cdot)$, for $x\neq x'$, we have:
\begin{equation}
    \frac{\partial^2 \pi_\theta(\cdot|x)^Th(\pi_\theta,x)}{\partial \theta(x,y_1)\partial\theta(x',y_2)} =0.
\end{equation}
Therefore if $x_1\neq x_2$,
\begin{equation}
    \frac{\partial^2 \mathbb{E}_{x\sim \mathcal{D},y\sim\pi_\theta(\cdot|x)}\left[\left(r_\theta(x,y)- r(x,y)\right)^2\right]}{\partial \theta(x_1,y_1)\partial\theta(x_2,y_2)} = \mathcal{D}(x_1) \frac{\partial}{\partial\theta(x_2,y_2)} \left(\frac{\partial \pi_\theta(\cdot|x_1)^Th(\pi_\theta,x_1)}{\partial \theta(x_1,y_1)}\right)=0.
\end{equation}

So the only care about the term when $x_1=x_2$, i.e.
\begin{equation}
    \frac{\partial^2 \mathbb{E}_{x\sim \mathcal{D},y\sim\pi_\theta(\cdot|x)}\left[\left(r_\theta(x,y)- r(x,y)\right)^2\right]}{\partial \theta(x,y_1)\partial\theta(x,y_2)} = \mathcal{D}(x) \frac{\partial}{\partial\theta(x,y_2)} \left(\frac{\partial \pi_\theta(\cdot|x)^Th(\pi_\theta,x)}{\partial \theta(x,y_1)}\right).
\end{equation}
Denote $f_{RA}(x,\theta) = \pi_\theta(\cdot|x)^Th(\pi_\theta,x) = \sum_{y \in \mathbb{Y}} \pi_\theta(y|x)h(\pi_\theta,x,y)$ and $h(\pi_\theta,x,y)={g}^2(\pi_\theta,x,y),\ g(\pi_\theta,x,y) = \frac{1}{\tau} \log \pi_\theta(y|x) + z(x,y)=r_\theta(x,y)- r(x,y)$, the first derivative of $ f_{RA}(x,\theta) $ is given by:
\begin{equation}
    \frac{\partial f_{RA}(x,\theta)}{\partial \theta(x,y')} = \sum_{y \in \mathbb{Y}} \left[\frac{\partial \pi_\theta(y|x)}{\partial \theta(x,y')} g^2(\pi_\theta,x,y) + 2\pi_\theta(y|x) g(\pi_\theta,x,y) \frac{\partial g(\pi_\theta,x,y)}{\partial \theta(x,y')}\right].
\end{equation}
where, $\frac{\partial g(\pi_\theta,x,y)}{\partial \theta(x,y')} = \frac{1}{\tau} \frac{\partial\log \pi_\theta(y|x)}{\partial \theta(x,y')}  = \frac{1}{\tau} (\delta_{yy'} - \pi_\theta(y'|x))$.

To compute the Hessian matrix, we take the derivative of this expression with respect to $\theta(y_2)$, yielding the $(y_1, y_2)$-th element as
\begin{equation}
    \frac{\partial^2 f_{RA}(x,\theta)}{\partial \theta(x,y_1) \partial \theta(x,y_2)} = \frac{\partial}{\partial \theta(x,y_2)} \left[\sum_{y \in \mathbb{Y}} \left(\frac{\partial \pi_\theta(y|x)}{\partial \theta(x,y_1)} g^2(\pi_\theta,x,y) + 2\pi_\theta(y|x) g(\pi_\theta,x,y) \frac{\partial g(\pi_\theta,x,y)}{\partial \theta(x,y_1)}\right)\right].
\end{equation}
We proceed by decomposing this expression into two parts for further derivation. The first part of our derivation is given by:
\begin{equation}
    \sum_{y \in \mathbb{Y}} \frac{\partial \pi_\theta(y | x)}{\partial \theta(x, y_1)} g^2(\pi_\theta, x, y) = \sum_{y \in \mathbb{Y}} \pi_\theta(y | x) (\delta_{yy_1} - \pi_\theta(y_1 | x)) g^2(\pi_\theta, x, y).
\end{equation}
Applying the chain rule, we first differentiate $\pi_\theta(y | x)(\delta_{yy_1} - \pi_\theta(y_1 | x))$ with respect to $\theta(x, y_2)$:
\begin{equation}
    \frac{\partial}{\partial \theta(x, y_2)} \left(\pi_\theta(y | x)(\delta_{yy_1} - \pi_\theta(y_1 | x))\right) = \frac{\partial \pi_\theta(y | x)}{\partial \theta(x, y_2)} (\delta_{yy_1} - \pi_\theta(y_1 | x)) - \pi_\theta(y | x) \frac{\partial \pi_\theta(y_1 | x)}{\partial \theta(x, y_2)}.
\end{equation}
Since $\frac{\partial \pi_\theta(y | x)}{\partial \theta(x, y_2)} = \pi_\theta(y | x)(\delta_{yy_2} - \pi_\theta(y_2 | x))$ and $\frac{\partial \pi_\theta(y_1 | x)}{\partial \theta(x, y_2)} = \pi_\theta(y_1 | x)(\delta_{y_1y_2} - \pi_\theta(y_2 | x))$, the derivative of the first part is
\begin{equation}
    \frac{\partial\left(\pi_\theta(y | x)(\delta_{yy_1} - \pi_\theta(y_1 | x))\right)}{\partial \theta(x, y_2)} = \pi_\theta(y | x)(\delta_{yy_2} - \pi_\theta(y_2 | x))(\delta_{yy_1} - \pi_\theta(y_1 | x)) - \pi_\theta(y | x) \pi_\theta(y_1 | x)(\delta_{y_1y_2} - \pi_\theta(y_2 | x)).
\end{equation}

Next, we differentiate $g^2(\pi_\theta, x, y)$ with respect to $\theta(x, y_2)$. Noting that $\frac{\partial g(\pi_\theta, x, y)}{\partial \theta(x, y_2)} = \frac{1}{\tau} (\delta_{yy_2} - \pi_\theta(y_2 | x))$, we obtain
\begin{equation}
    \frac{\partial g^2(\pi_\theta, x, y)}{\partial \theta(x, y_2)} = 2 g(\pi_\theta, x, y) \frac{\partial g(\pi_\theta, x, y)}{\partial \theta(x, y_2)} = \frac{2}{\tau} g(\pi_\theta, x, y)(\delta_{yy_2} - \pi_\theta(y_2 | x)).
\end{equation}
Combining all results, we obtain the derivative of the first part as follows:
\begin{equation}
    \begin{aligned}
\sum_{y \in \mathbb{Y}} &\left[\left(\pi_\theta(y | x)(\delta_{yy_2} - \pi_\theta(y_2 | x))(\delta_{yy_1} - \pi_\theta(y_1 | x)) - \pi_\theta(y | x) \pi_\theta(y_1 | x)(\delta_{y_1y_2} - \pi_\theta(y_2 | x))\right) g^2(\pi_\theta, x, y)\right. \\
&+ \left.\pi_\theta(y | x)(\delta_{yy_1} - \pi_\theta(y_1 | x)) \frac{2}{\tau} g(\pi_\theta, x, y)(\delta_{yy_2} - \pi_\theta(y_2 | x)) \right].
\end{aligned}
\end{equation}

For the second part, we begin with the expression:
\begin{equation}
    2\sum_{y \in \mathbb{Y}} \pi_\theta(y | x) g(\pi_\theta, x, y) \frac{\partial g(\pi_\theta, x, y)}{\partial \theta(x, y_1)} = \frac{2}{\tau} \sum_{y \in \mathbb{Y}} \pi_\theta(y | x) g(\pi_\theta, x, y)(\delta_{yy_1} - \pi_\theta(y_1 | x)).
\end{equation}

Using the product rule, this derivative with respect to $\theta(x, y_2)$ consists of two components:

1. The derivative of $\pi_\theta(y | x)$: 
   $\frac{2}{\tau} \sum_{y \in \mathbb{Y}} \frac{\partial \pi_\theta(y | x)}{\partial \theta(x, y_2)} g(\pi_\theta, x, y)(\delta_{yy_1} - \pi_\theta(y_1 | x))$,
   
2. The derivative of $g(\pi_\theta, x, y)(\delta_{yy_1} - \pi_\theta(y_1 | x))$: $\frac{2}{\tau} \sum_{y \in \mathbb{Y}} \pi_\theta(y | x) \left[\frac{\partial g(\pi_\theta, x, y)}{\partial \theta(x, y_2)}(\delta_{yy_1} - \pi_\theta(y_1 | x)) - g(\pi_\theta, x, y)\frac{\partial \pi_\theta(y_1 | x)}{\partial \theta(x, y_2)}\right]$.

For the derivative of $\pi_\theta(y | x)$, noting that $\frac{\partial \pi_\theta(y | x)}{\partial \theta(x, y_2)} = \pi_\theta(y | x)(\delta_{yy_2} - \pi_\theta(y_2 | x))$, we obtain
\begin{equation}
    \frac{2}{\tau} \sum_{y \in \mathbb{Y}} \pi_\theta(y | x)(\delta_{yy_2} - \pi_\theta(y_2 | x)) g(\pi_\theta, x, y)(\delta_{yy_1} - \pi_\theta(y_1 | x)).
\end{equation}

For the derivative of $g(\pi_\theta, x, y)(\delta_{yy_1} - \pi_\theta(y_1 | x))$, using $g(\pi_\theta, x, y) = \frac{1}{\tau} \log \pi_\theta(y | x) + z(y)$, we find
\begin{equation}
    \frac{\partial g(\pi_\theta, x, y)}{\partial \theta(x, y_2)} = \frac{1}{\tau} (\delta_{yy_2} - \pi_\theta(y_2 | x)),
\end{equation}
and since $\frac{\partial \pi_\theta(y_1 | x)}{\partial \theta(x, y_2)} = \pi_\theta(y_1 | x)(\delta_{y_1y_2} - \pi_\theta(y_2 | x))$, the derivative of the second part simplifies to:
\begin{equation}
    \small{\begin{aligned}
&2\sum_{y \in \mathbb{Y}} \pi_\theta(y | x) g(\pi_\theta, x, y) \frac{\partial g(\pi_\theta, x, y)}{\partial \theta(x, y_1)} = \frac{2}{\tau} \sum_{y \in \mathbb{Y}} \pi_\theta(y | x)(\delta_{yy_2} - \pi_\theta(y_2 | x)) g(\pi_\theta, x, y)(\delta_{yy_1} - \pi_\theta(y_1 | x))\\
&\quad+ \frac{2}{\tau^2} \sum_{y \in \mathbb{Y}} \pi_\theta(y | x)(\delta_{yy_2} - \pi_\theta(y_2 | x))(\delta_{yy_1} - \pi_\theta(y_1 | x)) - \frac{2}{\tau} \sum_{y \in \mathbb{Y}} \pi_\theta(y | x) g(\pi_\theta, x, y) \pi_\theta(y_1 | x)(\delta_{y_1y_2} - \pi_\theta(y_2 | x)).
\end{aligned}}
\end{equation}

Finally, combining terms, we derive the compact expression for the derivative of the second part with respect to $\theta(x, y_2)$:
\begin{equation}
    \small{\begin{aligned}
&2\sum_{y \in \mathbb{Y}} \pi_\theta(y | x) g(\pi_\theta, x, y) \frac{\partial g(\pi_\theta, x, y)}{\partial \theta(x, y_1)}\\
=& \sum_{y \in \mathbb{Y}} \left(\frac{2}{\tau} g(\pi_\theta, x, y) + \frac{2}{\tau^2}\right)\pi_\theta(y | x)(\delta_{yy_2} - \pi_\theta(y_2 | x))(\delta_{yy_1} - \pi_\theta(y_1 | x)) - \frac{2}{\tau} \pi_\theta(y | x) g(\pi_\theta, x, y) \pi_\theta(y_1 | x)(\delta_{y_1y_2} - \pi_\theta(y_2 | x)).
\end{aligned}}
\end{equation}

By gathering all terms, we derive the full expression for the Hessian matrix:
\begin{equation}
\small{    \begin{aligned}
&\frac{\partial}{\partial\theta(x,y_2)} \left(\frac{\partial \pi_\theta(\cdot|x)^Th(\pi_\theta,x)}{\partial \theta(x,y_1)}\right) = \frac{\partial^2 f_{RA}(x,\theta)}{\partial \theta(x, y_1) \partial \theta(x, y_2)} \\
=& \frac{\partial}{\partial \theta(x, y_2)} \left[\sum_{y \in \mathbb{Y}} \left(\frac{\partial \pi_\theta(y | x)}{\partial \theta(x, y_1)} g(\pi_\theta, x, y)^2 + 2 \pi_\theta(y | x) g(\pi_\theta, x, y) \frac{\partial g(\pi_\theta, x, y)}{\partial \theta(x, y_1)}\right)\right] \\
=& \sum_{y \in \mathbb{Y}} \left[\left(\pi_\theta(y | x)(\delta_{yy_2} - \pi_\theta(y_2 | x))(\delta_{yy_1} - \pi_\theta(y_1 | x)) - \pi_\theta(y | x) \pi_\theta(y_1 | x)(\delta_{y_1y_2} - \pi_\theta(y_2 | x))\right) g(\pi_\theta, x, y)^2\right. \\
&\quad+ \left.\pi_\theta(y | x)(\delta_{yy_1} - \pi_\theta(y_1 | x)) \frac{2}{\tau} g(\pi_\theta, x, y)(\delta_{yy_2} - \pi_\theta(y_2 | x))\right] \\
&+ \sum_{y \in \mathbb{Y}} \left(\frac{2}{\tau} g(\pi_\theta, x, y) + \frac{2}{\tau^2}\right) \pi_\theta(y | x)(\delta_{yy_2} - \pi_\theta(y_2 | x))(\delta_{yy_1} - \pi_\theta(y_1 | x)) \\
&\quad- \frac{2}{\tau} \pi_\theta(y | x) g(\pi_\theta, x, y) \pi_\theta(y_1 | x)(\delta_{y_1y_2} - \pi_\theta(y_2 | x)) \\
=& \sum_{y \in \mathbb{Y}} \left(g(\pi_\theta, x, y)^2 + \frac{4}{\tau} g(\pi_\theta, x, y) + \frac{2}{\tau^2}\right) \pi_\theta(y | x)(\delta_{yy_2} - \pi_\theta(y_2 | x))(\delta_{yy_1} - \pi_\theta(y_1 | x)) \\
&\quad - \sum_{y \in \mathbb{Y}} \left(g(\pi_\theta, x, y)^2 + \frac{2}{\tau} g(\pi_\theta, x, y)\right) \pi_\theta(y | x) \pi_\theta(y_1 | x)(\delta_{y_1y_2} - \pi_\theta(y_2 | x)).
\end{aligned}}
\end{equation}

Let $\vec{g}(\theta,x) = (g(\pi_\theta, x, y_1), g(\pi_\theta, x, y_2), \ldots, g(\pi_\theta, x, y_{|\mathbb{Y}|}))$, $\vec{G}_1(\theta,x) = -\vec{g}^2(\theta,x) - \frac{2}{\tau} \vec{g}(\theta,x)$, and $\vec{G}_2(\theta,x) = (\vec{g}(\theta,x) + \frac{2}{\tau})^2 - \frac{2}{\tau^2}$. So ${G}_1(\pi_\theta, x, y)=-g^2(\pi_\theta, x, y)-\frac{2}{\tau}g(\pi_\theta, x, y)$ and $G_2(\pi_\theta, x, y)$ is similar. Then, the Hessian matrix of $f_{RA}(x,\theta) = \pi_\theta(\cdot|x)^Th(\pi_\theta,x)$ simplifies as follows:
\begin{equation}
    \begin{aligned}
\nabla^2_\theta f_{RA}(x,\theta)_{i,j} = \sum_{y \in \mathbb{Y}} [&\pi_\theta(y | x) \pi_\theta(y_i | x)(\delta_{y_iy_j} - \pi_\theta(y_j | x)) {G}_1(\pi_\theta, x, y) \\
&+ \pi_\theta(y | x)(\delta_{yy_j} - \pi_\theta(y_j | x))(\delta_{yy_i} - \pi_\theta(y_i | x)) {G}_2(\pi_\theta, x, y)].
\end{aligned}
\end{equation}

Then we can calculate the spectral radius of $S(\theta)$, because if $x_1\neq x_2$, $\frac{\partial^2 \mathbb{E}_{x\sim \mathcal{D},y\sim\pi_\theta(\cdot|x)}\left[\left(r_\theta(x,y)- r(x,y)\right)^2\right]}{\partial \theta(x_1,y_1)\partial\theta(x_2,y_2)}=0$. So
\begin{equation}
\begin{aligned}
&\left|\sum_{x,x'\in\mathbb{X}}\sum_{y,y'\in\mathbb{Y}}z(x,y)S(\theta,x,y,x',y')z(x',y')\right| = \left|\sum_{x,x'\in\mathbb{X}}\sum_{y,y'\in\mathbb{Y}}z(x,y) \frac{\partial^2 \mathcal{L}_{RA}(\pi_\theta)}{\partial \theta(x,y)\partial\theta(x',y')} z(x',y')\right|\\
=& \left|\sum_{x\in\mathbb{X}}\sum_{y,y'\in\mathbb{Y}}z(x,y) \frac{\partial^2 \mathcal{L}_{RA}(\pi_\theta)}{\partial \theta(x,y)\partial\theta(x,y')} z(x,y')\right| =\left|\sum_{x\in\mathbb{X}}\sum_{y,y'\in\mathbb{Y}}z(x,y) \mathcal{D}(x) \frac{\partial^2 f_{RA}(x,\theta)}{\partial \theta(x, y) \partial \theta(x, y')}  z(x,y')\right|.
\end{aligned}
\end{equation}

To compute the quadratic form of the Hessian matrix, we proceed as follows:
\begin{equation}
    \small{\begin{aligned}
&|\sum_{x\in\mathbb{X}}\sum_{y,y'\in\mathbb{Y}}z(x,y) \mathcal{D}(x) \frac{\partial^2 f_{RA}(x,\theta)}{\partial \theta(x, y) \partial \theta(x, y')}  z(x,y')| \\
=& |\sum_{x\in\mathbb{X}}\mathcal{D}(x) \sum_{i=1}^{|\mathbb{Y}|} \sum_{j=1}^{|\mathbb{Y}|} z(x,y_i) \left(\nabla^2_\theta f_{RA}(x,\theta)_{i,j} \right) z(x,y_j)| \\
=& |\sum_{x\in\mathbb{X}}\mathcal{D}(x) ( \pi_\theta(\cdot|x)^\top \vec{G}_1(\theta,x) (\pi_\theta(\cdot|x)^\top z(x,\cdot))^2 + \pi_\theta(\cdot|x)^\top (\vec{G}_2(\theta,x) - \vec{G}_1(\theta,x)) (\pi_\theta(\cdot|x)^\top z(x,\cdot))^2 \\
&\quad + \sum_{i=1}^K z(x,y_i)^2 \pi_\theta(y_i | x) G_2(\pi_\theta, x, y_i) - 2 \pi_\theta(\cdot|x)^\top z(x,\cdot) \pi_\theta(y_i | x) G_2(\pi_\theta, x, y_i) z(x,y_i) ) |\\
=& | \sum_{x\in\mathbb{X}}\mathcal{D}(x) ( \pi_\theta(\cdot|x)^\top \vec{G}_1(\theta,x) (\pi_\theta(\cdot|x)^\top z(x,\cdot))^2 + \pi_\theta(\cdot|x)^\top (\vec{G}_2(\theta,x) - \vec{G}_1(\theta,x)) (\pi_\theta(\cdot|x)^\top z(x,\cdot))^2 \\
&\quad+ \sum_{i=1}^K (z(x,y_i) - 2 \pi_\theta(\cdot|x)^\top z(x,\cdot)) \pi_\theta(y_i | x) G_2(\pi_\theta, x, y_i) z(x,y_i) ) | \\
\triangleq& |\sum_{x\in\mathbb{X}}\mathcal{D}(x) \psi(x) |\leq \|\mathcal{D}(\cdot)\|_1 \|\psi(\cdot)\|_\infty=1\cdot\|\psi(\cdot)\|_\infty.
\end{aligned}}
\end{equation}

To establish an upper bound for the spectral radius of the Hessian matrix, let $\vec{G}_3(\theta,x) = \vec{G}_2(\theta,x) - \vec{G}_1(\theta,x)$. We can then write:
\begin{equation}
    \begin{aligned}
& \|\psi(\cdot)\|_\infty \\ 
=& \max_{x\in\mathbb{X}}\left|\pi_\theta(\cdot|x)^\top \vec{G}_1(\theta,x) \cdot (\pi_\theta(\cdot|x)^\top z(x,\cdot))^2 + \pi_\theta(\cdot|x)^\top \vec{G}_3(\theta,x) \cdot (\pi_\theta(\cdot|x)^\top z(x,\cdot) )^2 \right.\\
&\quad+ \left. \sum_{i=1}^K z(x,y_i)^2 \pi_\theta(y_i | x) G_2(\pi_\theta, x, y_i) + \sum_{i=1}^K 2 \pi_\theta(\cdot|x)^\top z(x,\cdot) \pi_\theta(y_i | x) G_2(\pi_\theta, x, y_i) z(x,y_i) \right| \\
\leq& \max_{x\in\mathbb{X}}\left|\pi_\theta(\cdot|x)^\top \vec{G}_1(\theta,x) \cdot (\pi_\theta(\cdot|x)^\top z(x,\cdot))^2\right| + \left|\pi_\theta(\cdot|x)^\top \vec{G}_3(\theta,x) \cdot (\pi_\theta(\cdot|x)^\top z(x,\cdot))^2\right| \\
&\quad+ \left|\sum_{i=1}^K z(x,y_i)^2 \pi_\theta(y_i | x) G_2(\pi_\theta, x, y_i)\right| + \left|\sum_{i=1}^K 2 \pi_\theta(\cdot|x)^\top z(x,\cdot) \pi_\theta(y_i | x) G_2(\pi_\theta, x, y_i) z(x,y_i)\right| \\
\leq& \max_{x\in\mathbb{X}}\|\pi_\theta(\cdot|x)\|_1 \|\vec{G}_1(\theta,x)\|_\infty \|z(x,\cdot)\|_2^2 + \|\pi_\theta(\cdot|x)\|_1 \|\vec{G}_3(\theta,x)\|_\infty \|z(x,\cdot)\|_2^2 \\
&\quad+ \|\vec{G}_2(\theta,x)\|_\infty \|z(x,\cdot)\|_2^2 + 2\|\pi_\theta(\cdot|x)\odot\vec{G}_2(\theta,x)\|_1 \|z(x,\cdot)\|_2^2 \\
=& \max_{x\in\mathbb{X}}\|\vec{G}_1(\theta,x)\|_\infty \|z(x,\cdot)\|_2^2 + \|\vec{G}_3(\theta,x)\|_\infty \|z(x,\cdot)\|_2^2 + \|\vec{G}_2(\theta,x)\|_\infty \|z(x,\cdot)\|_2^2 + 2\|\vec{G}_2(\theta,x)\|_\infty \|z(x,\cdot)\|_2^2 \\
=& \max_{x\in\mathbb{X}}(\|\vec{G}_1(\theta,x)\|_\infty + 3\|\vec{G}_2(\theta,x)\|_\infty + \|\vec{G}_3(\theta,x)\|_\infty) \|z(x,\cdot)\|_2^2\\
\leq& \max_{x\in\mathbb{X}}(\|\vec{G}_1(\theta,x)\|_\infty + 3\|\vec{G}_2(\theta,x)\|_\infty + \|\vec{G}_3(\theta,x)\|_\infty) \max_{x\in\mathbb{X}}\|z(x,\cdot)\|_2^2 \\
\leq& \max_{x\in\mathbb{X}}(\|\vec{G}_1(\theta,x)\|_\infty + 3\|\vec{G}_2(\theta,x)\|_\infty + \|\vec{G}_3(\theta,x)\|_\infty) \|z(\cdot,\cdot)\|_2^2 .
\end{aligned}
\end{equation}
In the second inequality, the first term arises from Hölder's inequality, which states that $ \pi_\theta^\top z(x,\cdot) \leq \|\pi_\theta\|_1 \|z(x,\cdot)\|_\infty = \|z(x,\cdot)\|_\infty $, where $ \|\pi_\theta\|_\infty \leq 1 $ and $ \|z^2(x,\cdot)\|_1 = \|z(x,\cdot)\|_2^2 $. For the second term, we have the bound on $ |(\pi_\theta^\top z(x,\cdot))^2| $:
\begin{equation}
    |(\pi_\theta^\top z(x,\cdot))^2| = \pi_\theta^\top z(x,\cdot) \cdot \pi_\theta^\top z(x,\cdot) \leq \|\pi_\theta\|_1 \|z(x,\cdot)\|_\infty \cdot \|\pi_\theta\|_1 \|z(x,\cdot)\|_\infty \leq \|z(x,\cdot)\|_2^2.
\end{equation}

For the third term:
\begin{equation}
    \left| \sum_{i=1}^K z(x,y_i)^2 \pi_\theta(y_i | x) G_2(y_i) \right| \leq \left| \sum_{i=1}^K z(x,y_i)^2 |G_2(y_i)| \right| \leq \|\vec{G}_2(\theta)\|_\infty \|z^2(x,\cdot)\|_1 = \|\vec{G}_2(\theta)\|_\infty \|z(x,\cdot)\|_2^2.
\end{equation}

For the fourth term:
\begin{equation}
    \begin{aligned}
|\sum_{i=1}^K 2\pi_\theta^Tz(x,\cdot)\pi_\theta(y_i)G_2(y_i)z(x,y_i)|&\leq2||\pi_\theta||_1||z(x,\cdot)||_\infty|\sum_{i=1}^K \pi_\theta(y_i)G_2(y_i)z(x,y_i)||\\
&\leq2||z(x,\cdot)||_2||\pi_\theta(\cdot|x)\odot\vec{G}_2(\theta)||_1||z(x,\cdot)||_\infty\\
&\leq2||\vec{G}_2(\theta)||_\infty||z(x,\cdot)||_2^2.
\end{aligned}
\end{equation}
Next, we analyze the coefficient $ (\|\vec{G}_1(\theta,x)\|_\infty + \|\vec{G}_2(\theta,x)\|_\infty + \|\vec{G}_3(\theta,x)\|_\infty + \|\vec{G}_2(\theta,x)\|_1) $. The term $ |g(\pi_\theta, x, y)| $ represents the error between the parameterized distribution $ \pi_\theta $ and the true distribution for component $ (\pi_\theta, x, y) $. In this Lemma, we assume $|r_\theta(x,y)- r(x,y)|\leq\epsilon_1$. Therefore $ |g(\pi_\theta, x, y)| \leq \epsilon_1 $ for all components $ (\pi_\theta, x, y) $, where $ \epsilon_1 $ denotes the upper bound of the error of the current parameterized distribution $ \pi_\theta $.

For $ ||\vec{G}_1(\theta,x)||_\infty $, we have:
\begin{equation}
    ||\vec{G}_1(\theta,x)||_\infty = ||\vec{g}^2(\theta,x) + \frac{2}{\tau} \vec{g}(\theta,x)||_\infty \leq \max\{\epsilon_1^2 + \frac{2}{\tau} \epsilon_1, \frac{1}{\tau}\}.
\end{equation}

For $ ||\vec{G}_2(\theta,x)||_\infty $, it follows that:
\begin{equation}
    ||\vec{G}_2(\theta,x)||_\infty = ||\vec{g}^2(\theta,x) + \frac{4}{\tau} \vec{g}(\theta,x) + \frac{2}{\tau^2}||_\infty \leq \epsilon_1^2 + \frac{4\epsilon_1}{\tau} + \frac{2}{\tau^2}.
\end{equation}

For $ ||\vec{G}_3(\theta,x)||_\infty $, we find:
\begin{equation}
    ||\vec{G}_3(\theta,x)||_\infty = ||\vec{G}_2(\theta,x) - \vec{G}_1(\theta)||_\infty = ||\frac{6}{\tau} \vec{g}(\theta,x) + \frac{2}{\tau^2}||_\infty \leq \frac{6\epsilon_1}{\tau} + \frac{2}{\tau^2}.
\end{equation}

Combining these results yields:
\begin{equation}
    \begin{aligned}
&||\vec{G}_1(\theta)||_\infty + 3||\vec{G}_2(\theta)||_\infty + ||\vec{G}_3(\theta)||_\infty \\
\leq & 3\epsilon_1^2 + \frac{18\epsilon_1}{\tau} + \frac{8}{\tau^2} + \max\left\{ \epsilon_1^2 + \frac{2}{\tau} \epsilon_1, \frac{1}{\tau} \right\} .
\end{aligned}
\end{equation}

Thus, the Lipschitz coefficient of the gradient of the target function with respect to the parameter $ \theta $ is given by:
\begin{equation}
    \begin{aligned}
&\left|\sum_{x,x'\in\mathbb{X}}\sum_{y,y'\in\mathbb{Y}}z(x,y)S(\theta,x,y,x',y')z(x',y')\right| \\
\leq& \max_{x\in\mathbb{X}}(\|\vec{G}_1(\theta,x)\|_\infty + 3\|\vec{G}_2(\theta,x)\|_\infty + \|\vec{G}_3(\theta,x)\|_\infty) \|z(\cdot,\cdot)\|_2^2\\
\leq & ( 3\epsilon_1^2 + \frac{18\epsilon_1}{\tau} + \frac{8}{\tau^2} + \max\left\{ \epsilon_1^2 + \frac{2}{\tau} \epsilon_1, \frac{1}{\tau} \right\} )\|z(\cdot,\cdot)\|_2^2.
    \end{aligned}
\end{equation}
Proof finished.

\subsection{Proof of Lemma \ref{RDA_L_coef}}\label{RDA_L_coef_proof}
\begin{lemma}\label{RDA_L_coef}
    (RDA Smoothness) Given softmax parametrization of Definition \ref{Softmax} for policy $\pi_\theta$, assume $|(r_\theta(x,y_1)-r_\theta(x,y_2))-(r(x,y_1)-r(x,y_2))|\leq\epsilon_3$,  $\forall r, \tau$, $\theta\rightarrow \mathbb{E}_{\mathcal{D}_{\text{pw}}}\left[\left( \frac{1}{\tau}\log\frac{\pi_\theta(y_1|x)}{\pi_\theta(y_2|x)}-(r(x,y_1)-r(x,y_2))\right)^2\right]$ is $\left(20\epsilon_2^2+\frac{32\epsilon_2}{\tau}+\frac{8}{\tau^2}\right)$-smooth where $\mathcal{D}_{\text{pw}}\triangleq\{(x, y_1, y_2) \mid x \in \mathcal{D}, y_1, y_2 \sim \pi_\theta(y|x)\}$ (see Eq.\ref{pair-wised_RA_eq}). See proof in Appendix \ref{RDA_L_coef_proof}.

\end{lemma}

\textbf{Proof:} By Lemma \ref{spectral_radius}, it suffices to show that the spectral radius of the hessian matrix of the second derivative of $\mathcal{L}_{\text{RDA}}(\pi_\theta) = \mathbb{E}_{\mathcal{D}_{\text{pw}}}\left[\left( \frac{1}{\tau}\log\frac{\pi_\theta(y_1|x)}{\pi_\theta(y_2|x)}-(r(x,y_1)-r(x,y_2))\right)^2\right]$ is bounded by $\left(20\epsilon_2^2+\frac{32\epsilon_2}{\tau}+\frac{8}{\tau^2}\right)$, i.e.
\begin{equation}\label{spectral_radius_RDA}
\left|\sum_{x,x'\in\mathbb{X}}\sum_{y,y'\in\mathbb{Y}}z(x,y) \frac{\partial^2 \mathcal{L}_{\text{RDA}}(\pi_\theta)}{\partial \theta(x,y)\partial\theta(x',y')} z(x',y')\right|\leq\left(20\epsilon_2^2+\frac{32\epsilon_2}{\tau}+\frac{8}{\tau^2}\right)||z(\cdot,\cdot)||_2^2.
\end{equation}
To show the bound, denote $h(\pi_\theta,x,y_1,y_2)=g^2\left( \pi_\theta,x,y_1,y_2\right)$ where $g(\pi_\theta,x,y_1,y_2)=\frac{1}{\tau}\log\frac{\pi_\theta(y_1|x)}{\pi_\theta(y_2|x)}-(r(x,y_1)-r(x,y_2))$, we have:
\begin{equation}
\begin{aligned}
&\left|\sum_{x,x'\in\mathbb{X}}\sum_{y,y'\in\mathbb{Y}}z(x,y) \frac{\partial^2 \mathbb{E}_{\mathcal{D}_{\text{pw}}}\left[\left( \frac{1}{\tau}\log\frac{\pi_\theta(y_1|x)}{\pi_\theta(y_2|x)}-(r(x,y_1)-r(x,y_2))\right)^2\right]}{\partial \theta(x,y)\partial\theta(x',y')} z(x',y')\right|\\
=&\left|\sum_{x\in\mathbb{X}}\sum_{y,y'\in\mathbb{Y}}z(x,y) \frac{\partial^2 \sum_{x\in\mathbb{X}}\mathcal{D}(x)\sum_{y_1,y_2\in\mathbb{Y}}\pi_\theta(y_1|x)\pi_\theta(y_2|x)h(\pi_\theta,x,y_1,y_2)}{\partial \theta(x,y)\partial\theta(x,y')}  z(x,y')\right|\\ =&\left|\sum_{x\in\mathbb{X}}\mathcal{D}(x)\sum_{y,y'\in\mathbb{Y}}z(x,y)  \frac{\partial^2 f_{\text{RDA}}(x,\theta)}{\partial \theta(x, y) \partial \theta(x, y')}  z(x,y')\right| \\
\triangleq& |\sum_{x\in\mathbb{X}}\mathcal{D}(x) \psi(x) |\leq \|\mathcal{D}(\cdot)\|_1 \|\psi(\cdot)\|_\infty=1\cdot\|\psi(\cdot)\|_\infty.
\end{aligned}
\end{equation}
where $f_{\text{RDA}}(x,\theta)=\sum_{y_1,y_2\in\mathbb{Y}}\pi_\theta(y_1|x)\pi_\theta(y_2|x)h(\pi_\theta,x,y_1,y_2)$.

Based on Lemma \ref{Pair_second_derivative_lemma}, the second derivative of $f_{\text{RDA}}(x,\theta)$ is:
\begin{equation}\label{RDA_decompose}
    \begin{aligned}
        \frac{\partial^2 f_{\text{RDA}}(x, \theta)}{\partial \theta(x, y_i) \partial \theta(x, y_j)} = \sum_{y_1,y_2\in\mathbb{Y}}& 2\frac{\partial^2 \pi_\theta(y_1|x)}{\partial \theta(x, y_i) \partial \theta(x, y_j)} \pi_\theta(y_2|x)h(\pi_\theta,x,y_1,y_2) + 2\frac{\partial \pi_\theta(y_1|x)}{\partial \theta(x, y_i)}\frac{\partial \pi_\theta(y_2|x)}{\partial \theta(x, y_j)}h(\pi_\theta,x,y_1,y_2) \\
        &+ 2\frac{\partial \pi_\theta(y_1|x)}{\partial \theta(x, y_i)}\pi_\theta(y_2|x) \frac{\partial h(\pi_\theta, x, y_1,y_2)}{\partial \theta(x, y_j)}+ 2\frac{\partial \pi_\theta(y_1|x)}{\partial \theta(x, y_j)}\pi_\theta(y_2|x) \frac{\partial h(\pi_\theta, x, y_1,y_2)}{\partial \theta(x, y_i)}\\
        & + \pi_\theta(y_1|x)\pi_\theta(y_2|x)\frac{\partial^2 h(\pi_\theta, x, y_1,y_2)}{\partial \theta(x, y_i) \partial \theta(x, y_j)}.
    \end{aligned}
\end{equation}

Because $|(r_\theta(x,y_1)-r_\theta(x,y_2))-(r(x,y_1)-r(x,y_2))|\leq\epsilon_2$, $|g(\pi_\theta,x,y_1,y_2)|\leq\epsilon_2$.

According to the absolute value inequality, $|\psi(x)|=\left|\sum_{y_i,y_j\in\mathbb{Y}}z(x,y_i) \frac{\partial^2 f_{\text{RDA}}(x,\theta)}{\partial \theta(x, y_i) \partial \theta(x, y_j)}  z(x,y_j)\right|$ can be decomposed into five parts based on Eq.\ref{RDA_decompose} for further analysis.
\begin{equation}\label{RDA_decompose_abs}
    \begin{aligned}
&\left|\sum_{y_i,y_j\in\mathbb{Y}}z(x,y_i) \frac{\partial^2 f_{\text{RDA}}(x,\theta)}{\partial \theta(x, y_i) \partial \theta(x, y_j)}  z(x,y_j)\right| \\
\leq& \left|\sum_{y_i,y_j\in\mathbb{Y}}z(x,y_i)  \sum_{y_1,y_2\in\mathbb{Y}} 2\frac{\partial^2 \pi_\theta(y_1|x)}{\partial \theta(x, y_i) \partial \theta(x, y_j)} \pi_\theta(y_2|x)h(\pi_\theta,x,y_1,y_2)  z(x,y_j)\right| \\
&+\left|\sum_{y_i,y_j\in\mathbb{Y}}z(x,y_i) \sum_{y_1,y_2\in\mathbb{Y}}  2\frac{\partial \pi_\theta(y_1|x)}{\partial \theta(x, y_i)}\frac{\partial \pi_\theta(y_2|x)}{\partial \theta(x, y_j)}h(\pi_\theta,x,y_1,y_2) z(x,y_j)\right| \\
&+\left|\sum_{y_i,y_j\in\mathbb{Y}}z(x,y_i) \sum_{y_1,y_2\in\mathbb{Y}}  2\frac{\partial \pi_\theta(y_1|x)}{\partial \theta(x, y_i)}\pi_\theta(y_2|x) \frac{\partial h(\pi_\theta, x, y_1,y_2)}{\partial \theta(x, y_j)}  z(x,y_j)\right|  \\
&+ \left|\sum_{y_i,y_j\in\mathbb{Y}}z(x,y_i) \sum_{y_1,y_2\in\mathbb{Y}}   2\frac{\partial \pi_\theta(y_1|x)}{\partial \theta(x, y_j)}\pi_\theta(y_2|x) \frac{\partial h(\pi_\theta, x, y_1,y_2)}{\partial \theta(x, y_i)}  z(x,y_j)\right| \\
&+\left|\sum_{y_i,y_j\in\mathbb{Y}}z(x,y_i) \sum_{y_1,y_2\in\mathbb{Y}} \pi_\theta(y_1|x)\pi_\theta(y_2|x)\frac{\partial^2 h(\pi_\theta, x, y_1,y_2)}{\partial \theta(x, y_i) \partial \theta(x, y_j)}  z(x,y_j)\right|.
    \end{aligned}
\end{equation}

For the first and second terms of Eq.\ref{RDA_decompose_abs}, we have:
\begin{equation}\label{RDA_1_2_terms}
\begin{aligned}
    & \left|\sum_{y_i,y_j\in\mathbb{Y}}z(x,y_i)  \sum_{y_1,y_2\in\mathbb{Y}} 2\frac{\partial^2 \pi_\theta(y_1|x)}{\partial \theta(x, y_i) \partial \theta(x, y_j)} \pi_\theta(y_2|x)h(\pi_\theta,x,y_1,y_2)  z(x,y_j)\right| \\
&+ \left|\sum_{y_i,y_j\in\mathbb{Y}}z(x,y_i) \sum_{y_1,y_2\in\mathbb{Y}}  2\frac{\partial \pi_\theta(y_1|x)}{\partial \theta(x, y_i)}\frac{\partial \pi_\theta(y_2|x)}{\partial \theta(x, y_j)}h(\pi_\theta,x,y_1,y_2) z(x,y_j)\right| \\
=&\left|\sum_{y_i,y_j\in\mathbb{Y}}z(x,y_i)  \sum_{y_1,y_2\in\mathbb{Y}}4  \pi_\theta(y_1 | x)\pi_\theta(y_2 | x)(\delta_{y_1y_i} - \pi_\theta(y_i | x))(\delta_{y_2y_j} - \pi_\theta(y_j | x))h(\pi_\theta,x,y_1,y_2)  z(x,y_j)\right| \\
&+\left|\sum_{y_i,y_j\in\mathbb{Y}}z(x,y_i) \sum_{y_1,y_2\in\mathbb{Y}}2\pi_\theta(y_1 | x)\pi_\theta(y_2 | x) \pi_\theta(y_i | x)(\delta_{y_iy_j} - \pi_\theta(y_j | x))h(\pi_\theta,x,y_1,y_2)  z(x,y_j)\right| \\
\leq&\left|\sum_{y_i,y_j\in\mathbb{Y}}z(x,y_i)  \sum_{y_1,y_2\in\mathbb{Y}}4  \pi_\theta(y_1 | x)\pi_\theta(y_2 | x)(\delta_{y_1y_i} - \pi_\theta(y_i | x))(\delta_{y_2y_j} - \pi_\theta(y_j | x))\epsilon_2^2  z(x,y_j)\right| \\
&+\left|\sum_{y_i,y_j\in\mathbb{Y}}z(x,y_i) \sum_{y_1,y_2\in\mathbb{Y}}2\pi_\theta(y_1 | x)\pi_\theta(y_2 | x) \pi_\theta(y_i | x)(\delta_{y_iy_j} - \pi_\theta(y_j | x))\epsilon_2^2  z(x,y_j)\right|.
\end{aligned}
\end{equation}
The first equality is because:
\begin{equation}
    \frac{\partial^2 \pi_\theta(y|x)}{\partial \theta(x, y_i) \partial \theta(x, y_j)} = \pi_\theta(y | x)(\delta_{yy_j} - \pi_\theta(y_j | x))(\delta_{yy_i} - \pi_\theta(y_i | x)) - \pi_\theta(y | x) \pi_\theta(y_i | x)(\delta_{y_iy_j} - \pi_\theta(y_j | x)).
\end{equation}
The second inequality is because:
\begin{equation}
|h(\pi_\theta,x,y_1,y_2)|=|g^2(\pi_\theta,x,y_1,y_2)|\leq\epsilon_2^2.
\end{equation}
Then:
\begin{equation}\label{RDA_1_2_terms_1part}
\begin{aligned}
&\left|\sum_{y_i,y_j\in\mathbb{Y}}z(x,y_i)   \sum_{y_1,y_2\in\mathbb{Y}}4  \pi_\theta(y_1 | x)\pi_\theta(y_2 | x)(\delta_{y_1y_i} - \pi_\theta(y_i | x))(\delta_{y_2y_j} - \pi_\theta(y_j | x))\epsilon_2^2  z(x,y_j)\right| \leq 16\epsilon_2^2 \|z(x,\cdot)\|_2^2.
\end{aligned}
\end{equation}
The inequality is because we can expand the absolute value operation in the second-to-last line into four terms, and these four terms can be deduced to be less than $\|z(x,\cdot)\|_2^2$ using some existing conclusions. The relevant conclusions are:
$ \pi_\theta^\top z(x,\cdot) \leq \|\pi_\theta\|_1 \|z(x,\cdot)\|_\infty = \|z(x,\cdot)\|_\infty\leq\|z(x,\cdot)\|_2 $, $ \pi_\theta^\top z^2(x,\cdot) \leq \|z(x,\cdot)\|_2^2$, $ \|\pi_\theta\|_\infty \leq 1 $, $ \|z^2(x,\cdot)\|_1 = \|z(x,\cdot)\|_2^2 $ and $|(\pi_\theta^\top z(x,\cdot))^2|  \leq \|z(x,\cdot)\|_2^2$.

And similarly, we have:
\begin{equation}\label{RDA_1_2_terms_2part}
    \begin{aligned}
&\left|\sum_{y_i,y_j\in\mathbb{Y}}z(x,y_i) \sum_{y_1,y_2\in\mathbb{Y}}2\pi_\theta(y_1 | x)\pi_\theta(y_2 | x) \pi_\theta(y_i | x)(\delta_{y_iy_j} - \pi_\theta(y_j | x))\epsilon_2^2  z(x,y_j)\right| \\
\leq& 2\left|\sum_{y_i,y_j\in\mathbb{Y}}z(x,y_i) \delta_{y_iy_j}\epsilon_2^2  z(x,y_j)\right| +2\left|\sum_{y_i,y_j\in\mathbb{Y}}z(x,y_i)\pi_\theta(y_i | x)\pi_\theta(y_j | x)\epsilon_2^2  z(x,y_j)\right| \\
\leq& 2\epsilon_2^2+2\left|\sum_{y_i,y_j\in\mathbb{Y}}z(x,y_i)\pi_\theta(y_i | x)\pi_\theta(y_j | x)\epsilon_2^2  z(x,y_j)\right| \\
\leq& 2\epsilon_2^2+2\epsilon_2^2\left|\sum_{y_i\in\mathbb{Y}}z(x,y_i)\pi_\theta(y_i | x)\sum_{y_j\in\mathbb{Y}}\pi_\theta(y_j | x)  z(x,y_j)\right| \\
\leq& 2\epsilon_2^2+2\epsilon_2^2\left|\|\pi(\cdot|x)\|_1\|z(x,\cdot)\|_\infty\|\pi(\cdot|x)\|_1\|z(x,\cdot)\|_\infty\right|
\leq 4\epsilon_2^2\|z(x,\cdot)\|_2^2.
    \end{aligned}
\end{equation}

For the third and forth terms of Eq.\ref{RDA_decompose_abs}, we have:
\begin{equation}\label{RDA_3_4_terms}
\begin{aligned}
    &\left|\sum_{y_i,y_j\in\mathbb{Y}}z(x,y_i) \sum_{y_1,y_2\in\mathbb{Y}}  2\frac{\partial \pi_\theta(y_1|x)}{\partial \theta(x, y_i)}\pi_\theta(y_2|x) \frac{\partial h(\pi_\theta, x, y_1,y_2)}{\partial \theta(x, y_j)}  z(x,y_j)\right|  \\
&+ \left|\sum_{y_i,y_j\in\mathbb{Y}}z(x,y_i) \sum_{y_1,y_2\in\mathbb{Y}}   2\frac{\partial \pi_\theta(y_1|x)}{\partial \theta(x, y_j)}\pi_\theta(y_2|x) \frac{\partial h(\pi_\theta, x, y_1,y_2)}{\partial \theta(x, y_i)}  z(x,y_j)\right| \\
=& 2\left|\sum_{y_i,y_j\in\mathbb{Y}}z(x,y_i) \sum_{y_1,y_2\in\mathbb{Y}}  2\frac{\partial \pi_\theta(y_1|x)}{\partial \theta(x, y_i)}\pi_\theta(y_2|x) \frac{\partial h(\pi_\theta, x, y_1,y_2)}{\partial \theta(x, y_j)}  z(x,y_j)\right|  \\
=& 8\left|\sum_{y_i,y_j\in\mathbb{Y}}z(x,y_i) \sum_{y_1,y_2\in\mathbb{Y}}  \frac{\partial \pi_\theta(y_1|x)}{\partial \theta(x, y_i)}\pi_\theta(y_2|x) g(\pi_\theta, x, y_1,y_2)\frac{\partial g(\pi_\theta,x,y_1,y_2)}{\partial \theta(x, y_j)}  z(x,y_j)\right| \\
=& 8\left|\sum_{y_i,y_j\in\mathbb{Y}}z(x,y_i) \sum_{y_1,y_2\in\mathbb{Y}}  \pi_\theta(y_1|x)(\delta_{y_1y_i}-\pi_\theta(y_i|x))\pi_\theta(y_2|x) g(\pi_\theta, x, y_1,y_2)\frac{1}{\tau}(\delta_{y_1y_j} - \delta_{y_2y_j})  z(x,y_j)\right| \\
\leq&\frac{8\epsilon_2}{\tau}\left|\sum_{y_i,y_j\in\mathbb{Y}}z(x,y_i) \sum_{y_1,y_2\in\mathbb{Y}}  \pi_\theta(y_1|x)(\delta_{y_1y_i}-\pi_\theta(y_i|x))\pi_\theta(y_2|x) \frac{1}{\tau}(\delta_{y_1y_j} - \delta_{y_2y_j})  z(x,y_j)\right| \\
\leq& \frac{32\epsilon_2}{\tau}\|z(x,\cdot)\|_2^2.
\end{aligned}
\end{equation}
where the first derivative of $g(\pi_\theta,x,y_1,y_2)$ is:
\begin{equation}
    \begin{aligned}
\frac{\partial g(\pi_\theta,x,y_1,y_2)}{\partial \theta(x, y_i)} = \frac{\partial \frac{1}{\tau}(\log{\pi_\theta(y_1|x)} -\log{\pi_\theta(y_2|x)})}{\partial \theta(x, y_i)} = \frac{1}{\tau}(\delta_{y_1y_i} - \pi_\theta(y_i|x) - \delta_{y_2y_i} + \pi_\theta(y_i|x)) = \frac{1}{\tau}(\delta_{y_1y_i} - \delta_{y_2y_i}).
    \end{aligned}
\end{equation}

And the last inequality is because we can expand the absolute value operation in the second-to-last line into four terms, and these four terms can be deduced to be less than $\|z(x,\cdot)\|_2^2$ using some existing conclusions. The relevant conclusions are:
$ \pi_\theta^\top z(x,\cdot) \leq \|\pi_\theta\|_1 \|z(x,\cdot)\|_\infty = \|z(x,\cdot)\|_\infty\leq\|z(x,\cdot)\|_2 $, $ \pi_\theta^\top z^2(x,\cdot) \leq \|z(x,\cdot)\|_2^2$, $ \|\pi_\theta\|_\infty \leq 1 $, $ \|z^2(x,\cdot)\|_1 = \|z(x,\cdot)\|_2^2 $ and $|(\pi_\theta^\top z(x,\cdot))^2|  \leq \|z(x,\cdot)\|_2^2$.

For the fifth term of Eq.\ref{RDA_decompose_abs}, we have:
\begin{equation}\label{RDA_5_terms}
    \begin{aligned}
&\left|\sum_{y_i,y_j\in\mathbb{Y}}z(x,y_i) \sum_{y_1,y_2\in\mathbb{Y}} \pi_\theta(y_1|x)\pi_\theta(y_2|x)\frac{\partial^2 h(\pi_\theta, x, y_1,y_2)}{\partial \theta(x, y_i) \partial \theta(x, y_j)}  z(x,y_j)\right| \\
=& \left|\sum_{y_i,y_j\in\mathbb{Y}}z(x,y_i) \sum_{y_1,y_2\in\mathbb{Y}} \pi_\theta(y_1|x)\pi_\theta(y_2|x)2\frac{\partial g(\pi_\theta, x, y_1,y_2)}{\partial \theta(x, y_i)}\frac{\partial g(\pi_\theta, x, y_1,y_2)}{\partial \theta(x, y_j)}  z(x,y_j)\right| \\
=& \left|\sum_{y_i,y_j\in\mathbb{Y}}z(x,y_i) \sum_{y_1,y_2\in\mathbb{Y}} \pi_\theta(y_1|x)\pi_\theta(y_2|x)\frac{2}{\tau^2}(\delta_{y_1y_i} - \delta_{y_2y_i})(\delta_{y_1y_j} - \delta_{y_2y_j})  z(x,y_j)\right|\\
=& |\sum_{y_i\in\mathbb{Y}}z(x,y_i)   \pi_\theta(y_i|x)\frac{2}{\tau^2}z(x,y_i) - \sum_{y_i\in\mathbb{Y}}z(x,y_i)   \pi_\theta(y_i|x)\sum_{y_j\in\mathbb{Y}}\pi_\theta(y_j|x)\frac{2}{\tau^2}z(x,y_j) \\
&- \sum_{y_i\in\mathbb{Y}}z(x,y_i)   \pi_\theta(y_i|x)\sum_{y_j\in\mathbb{Y}}\pi_\theta(y_j|x)\frac{2}{\tau^2}z(x,y_j) + \sum_{y_i\in\mathbb{Y}}z(x,y_i)   \pi_\theta(y_i|x)\frac{2}{\tau^2}z(x,y_i)|,
    \end{aligned}
\end{equation}
$$    \begin{aligned}
\leq& \frac{4}{\tau^2}|\sum_{y_i\in\mathbb{Y}}z(x,y_i)   \pi_\theta(y_i|x)z(x,y_i)| + |\sum_{y_i\in\mathbb{Y}}z(x,y_i)   \pi_\theta(y_i|x)\sum_{y_j\in\mathbb{Y}}\pi_\theta(y_j|x)z(x,y_j)|\\
=& \frac{4}{\tau^2} (\pi^T_\theta(\cdot|x)z^2(x,\cdot) + (\pi^T_\theta(\cdot|x)z(x,\cdot))^2) \leq \frac{4}{\tau^2} (\|\pi_\theta(\cdot|x)\|_1\|z^2(x,\cdot)\|_\infty + (\|\pi_\theta(\cdot|x)\|_1\|z(x,\cdot)\|_\infty)^2) \\
\leq& \frac{8}{\tau^2}\|z(x,\cdot)\|_2^2.
    \end{aligned}$$
where the second derivative of $g(\pi_\theta,x,y_1,y_2)$ is:
\begin{equation}
    \begin{aligned}
\frac{\partial^2 g(\pi_\theta,x,y_1,y_2)}{\partial \theta(x, y_i)\partial \theta(x, y_j)} = \frac{\partial}{\partial \theta(x, y_j)}(\frac{1}{\tau}(\delta_{y_1y_i} - \delta_{y_2y_i}))=0.
    \end{aligned}
\end{equation}
In summary, we have the upper bound of Eq.\ref{RDA_decompose_abs}:
\begin{equation}
    \begin{aligned}
\|\psi(x)\|_\infty=&\max_{x\in\mathbb{X}}|\psi(x)|=\max_{x\in\mathbb{X}}\left|\sum_{y_i,y_j\in\mathbb{Y}}z(x,y_i) \frac{\partial^2 f_{\text{RDA}}(x,\theta)}{\partial \theta(x, y_i) \partial \theta(x, y_j)}  z(x,y_j)\right| \\
\leq& \left(20\epsilon_2^2+\frac{32\epsilon_2}{\tau}+\frac{8}{\tau^2}\right)\max_{x\in\mathbb{X}}\|z(x,\cdot)\|_2^2 \\
\leq& \left(20\epsilon_2^2+\frac{32\epsilon_2}{\tau}+\frac{8}{\tau^2}\right)\|z(\cdot,\cdot)\|_2^2
    \end{aligned}
\end{equation}
Then Eq.\ref{spectral_radius_RDA} is proved. Proof finished.

\subsection{Proof of Lemma \ref{PRA_L_coef}}\label{PRA_L_coef_proof}
\begin{lemma}\label{PRA_L_coef}
    (PRA Smoothness) Given softmax parametrization of Definition \ref{Softmax} for policy $\pi_\theta$, assume $|\theta(x_1,y_1)-\theta(x_2,y_2)|\leq d$ and $|p^* - \sigma(\frac{1}{\tau}\log\frac{\pi_\theta(y_1|x)}{\pi_\theta(y_2|x)}) |\leq \epsilon_4$,  $\forall r, \tau$, $\theta\rightarrow \mathbb{E}_{\mathcal{D}_{\theta}}\left[\mathrm{D}_{\mathrm{KL}}(p^*(z|y_1,y_2,x)||p_\theta(z|y_1,y_2,x))\right]$ is $(20\log(1+ e^{\frac{d}{\tau}})+\frac{16\epsilon_3}{\tau}+\frac{4}{\tau^2}+16\log2)$-smooth where $\mathcal{D}_{\theta} \triangleq \{(x, y_1, y_2, z) \mid x \in \mathcal{D}, y_1, y_2\sim\pi_\theta(\cdot|x), z = 1 \text{ if } r(x, y_1) > r(x, y_2); 0 \text{ otherwise}\}$ (see Eq.\ref{PRA_eq}). See proof in Appendix \ref{PRA_L_coef_proof}.
\end{lemma}

\textbf{Proof:} Based on Appendix \ref{Derivation_PRA}, we have:
\begin{equation}
\begin{aligned}
&\mathbb{E}_{\mathcal{D}_{\theta}}\left[\mathrm{D}_{\mathrm{KL}}(p^*(z|y_1,y_2,x)||\sigma\left( h_\theta(x,y_1,y_2)\right))\right]\\
=&-\mathbb{E}_{x\sim \mathcal{D},y_1,y_2\sim \pi_\theta(y|x)}\left[p^*(z=1|y_1,y_2,x)\log\sigma\left( h_\theta(x,y_1,y_2)\right)+p^*(z=0|y_1,y_2,x)\log\sigma\left( h_\theta(x,y_2,y_1)\right)\right]\\
&+\mathbb{E}_{x\sim \mathcal{D},y_1,y_2\sim \pi_\theta(y|x)}\left[p^*(z=1|y_1,y_2,x)\log p^*(z=1|y_1,y_2,x)+p^*(z=0|y_1,y_2,x)\log p^*(z=0|y_1,y_2,x)\right]\\
=&-\mathbb{E}_{x\sim \mathcal{D},y_1,y_2\sim \pi_\theta(y|x)}\left[p^*(z=1|y_1,y_2,x)\log\sigma\left( h_\theta(x,y_1,y_2)\right)+p^*(z=0|y_1,y_2,x)\log\sigma\left( h_\theta(x,y_2,y_1)\right)\right]\\
&+\mathbb{E}_{x\sim \mathcal{D},y_1,y_2\sim \pi_\theta(y|x)}\left[M(x,y_1,y_2)\right]\\
\triangleq& \mathcal{L}_{\text{PRA}}(\pi_\theta).
\end{aligned}
\end{equation}
where $\sigma(x)=\frac{1}{1+e^{-x}}$ is the sigmoid function, $M(x,y_1,y_2)=\sum_{z=0,1}p^*(z|y_1,y_2,x)\log p^*(z|y_1,y_2,x)$ and $(x,y_1,y_2)\sim \mathcal{D}_{\theta}\triangleq y_1\succ y_2\sim p^*(z=1|y_1,y_2,x),y_1,y_2\sim\pi_\theta(y|x),x\sim\mathcal{D}$. 

By Lemma \ref{spectral_radius}, it suffices to show that the spectral radius of the hessian matrix of the second derivative of $\mathcal{L}_{\text{PRA}}(\pi_\theta)$, i.e.
\begin{equation}\label{spectral_radius_pra}
\begin{aligned}
    &\left|\sum_{x,x'\in\mathbb{X}}\sum_{y_i,y_j\in\mathbb{Y}}z(x,y_i) \frac{\partial^2 \mathcal{L}_{\text{PRA}}(\pi_\theta)}{\partial \theta(x,y_i)\partial\theta(x',y_j)} z(x',y_j)\right|\\
    \leq&\left(20\log(1+ e^{\frac{d}{\tau}})+\frac{16\epsilon_3}{\tau}+\frac{4}{\tau^2} + 16\log2\right)||z(\cdot,\cdot)||_2^2.
\end{aligned}
\end{equation}
Denote $h(\pi_\theta,x,y_1,y_2)=-p^*(z=1|y_1,y_2,x)\log\sigma\left( h_\theta(x,y_1,y_2)\right)-p^*(z=0|y_1,y_2,x)\log\sigma\left( h_\theta(x,y_2,y_1)\right)$, we have:
\begin{equation}\label{pra_decompose_psi_with_M_eq}
\begin{aligned}
&\left|\sum_{x,x'\in\mathbb{X}}\sum_{y_i,y_j\in\mathbb{Y}}z(x,y_i) \frac{\partial^2 \mathcal{L}_{\text{PRA}}(\pi_\theta)}{\partial \theta(x,y_i)\partial\theta(x',y_j)} z(x',y_j)\right|\\
=&\left|\sum_{x\in\mathbb{X}}\sum_{y_i,y_j\in\mathbb{Y}}z(x,y_i) \frac{\partial^2 \sum_{x\in\mathbb{X}}\mathcal{D}(x)\sum_{y_1,y_2\in\mathbb{Y}}\pi_\theta(y_1|x)\pi_\theta(y_2|x)(h(\pi_\theta,x,y_1,y_2)+M(x,y_1,y_2))}{\partial \theta(x,y_i)\partial\theta(x,y_j)}  z(x,y_j)\right|\\ =&\left|\sum_{x\in\mathbb{X}}\mathcal{D}(x)\sum_{y_i,y_j\in\mathbb{Y}}z(x,y_i)  \frac{\partial^2 f_{\text{PRA}}(x,\theta)}{\partial \theta(x, y_i) \partial \theta(x, y_j)}  z(x,y_j)\right| \\
&+ \left|\sum_{x\in\mathbb{X}}\mathcal{D}(x)\sum_{y_i,y_j\in\mathbb{Y}}z(x,y_i)  \frac{\partial^2 \sum_{y_1,y_2\in\mathbb{Y}}\pi_\theta(y_1|x)\pi_\theta(y_2|x)M(x,y_1,y_2)}{\partial \theta(x, y_i) \partial \theta(x, y_j)}  z(x,y_j)\right| \\
\triangleq& |\sum_{x\in\mathbb{X}}\mathcal{D}(x) \psi(x) | + \left|\sum_{x\in\mathbb{X}}\mathcal{D}(x)\sum_{y_i,y_j\in\mathbb{Y}}z(x,y_i)  \frac{\partial^2 \sum_{y_1,y_2\in\mathbb{Y}}\pi_\theta(y_1|x)\pi_\theta(y_2|x)M(x,y_1,y_2)}{\partial \theta(x, y_i) \partial \theta(x, y_j)}  z(x,y_j)\right|  \\
\leq& \|\mathcal{D}(\cdot)\|_1 \|\psi(\cdot)\|_\infty  + \left|\sum_{x\in\mathbb{X}}\mathcal{D}(x)\sum_{y_i,y_j\in\mathbb{Y}}z(x,y_i)  \frac{\partial^2 \sum_{y_1,y_2\in\mathbb{Y}}\pi_\theta(y_1|x)\pi_\theta(y_2|x)M(x,y_1,y_2)}{\partial \theta(x, y_i) \partial \theta(x, y_j)}  z(x,y_j)\right|\\
=& 1\cdot\|\psi(\cdot)\|_\infty  + \left|\sum_{x\in\mathbb{X}}\mathcal{D}(x)\sum_{y_i,y_j\in\mathbb{Y}}z(x,y_i)  \frac{\partial^2 \sum_{y_1,y_2\in\mathbb{Y}}\pi_\theta(y_1|x)\pi_\theta(y_2|x)M(x,y_1,y_2)}{\partial \theta(x, y_i) \partial \theta(x, y_j)}  z(x,y_j)\right|.
\end{aligned}
\end{equation}
where $f_{\text{PRA}}(x,\theta)=\sum_{y_1,y_2\in\mathbb{Y}}\pi_\theta(y_1|x)\pi_\theta(y_2|x)h(\pi_\theta,x,y_1,y_2)$ and $\psi(x)=\sum_{y_i,y_j\in\mathbb{Y}}z(x,y_i)  \frac{\partial^2 f_{\text{PRA}}(x,\theta)}{\partial \theta(x, y_i) \partial \theta(x, y_j)}  z(x,y_j)$.

Because $|\theta(x_1,y_1)-\theta(x_2,y_2)|\leq d$ and the Definition \ref{Softmax} for policy $\pi_\theta$, consider the upper bound of $h(\pi_\theta,x,y_1,y_2)$, we have:
\begin{equation}
\begin{aligned}
    |\log\sigma\left( h_\theta(x,y_1,y_2)\right)| =& |\log\sigma\left( \frac{1}{\tau}\log\frac{\pi_\theta(y_1|x)}{\pi_\theta(y_2|x)}\right)| = |\log\sigma\left( \frac{1}{\tau}\log\frac{\exp(\theta(y_1,x))}{\exp(\theta(y_2,x))}\right)| \\
    =& |\log\sigma\left( \frac{1}{\tau}(\theta(y_1,x)-\theta(y_2,x))\right)|
    \leq -\log\sigma\left( \frac{-d}{\tau}\right)\\
    =& \log \sigma^{-1}\left( \frac{-d}{\tau}\right) = \log(1+ e^{\frac{d}{\tau}}).
\end{aligned}
\end{equation}
Then 
\begin{equation}
\begin{aligned}
    &|h(\pi_\theta,x,y_1,y_2)|\\
    =&|p^*(z=1|y_1,y_2,x)\log\sigma\left( h_\theta(x,y_1,y_2)\right)+p^*(z=0|y_1,y_2,x)\log\sigma\left( h_\theta(x,y_2,y_1)\right)| \\
    \leq& p^*(z=1|y_1,y_2,x)\log(1+ e^{\frac{d}{\tau}}) + p^*(z=0|y_1,y_2,x)\log(1+ e^{\frac{d}{\tau}})=\log(1+ e^{\frac{d}{\tau}}).
\end{aligned}
\end{equation}

Consider the first derivative of  $h(\pi_\theta,x,y_1,y_2)$, denote $p^*=p^*(1|y_1,y_2,x)$:
\begin{equation}
    \begin{aligned}
\frac{\partial h(\pi_\theta,x,y_1,y_2)}{\partial \theta(x, y_i)} =& -p^*\frac{\partial }{\partial \theta(x, y_i) }(\log\sigma(\frac{1}{\tau}\log\frac{\pi_\theta(y_1|x)}{\pi_\theta(y_2|x)})) - (1-p^*)\frac{\partial }{\partial \theta(x, y_i) }(\log\sigma(\frac{1}{\tau}\log\frac{\pi_\theta(y_2|x)}{\pi_\theta(y_1|x)})) \\
=& -\left(p^*(1-\sigma(\frac{1}{\tau}\log\frac{\pi_\theta(y_1|x)}{\pi_\theta(y_2|x)})) - (1-p^*)\sigma(\frac{1}{\tau}\log\frac{\pi_\theta(y_1|x)}{\pi_\theta(y_2|x)}) \right)\frac{1}{\tau}\frac{\partial }{\partial \theta(x, y_i) }(\log\frac{\pi_\theta(y_1|x)}{\pi_\theta(y_2|x)}) \\
=& -\left(p^* - \sigma(\frac{1}{\tau}\log\frac{\pi_\theta(y_1|x)}{\pi_\theta(y_2|x)}) \right) \frac{1}{\tau}(\delta_{y_1y_i}-\delta_{y_2y_i}).
    \end{aligned}
\end{equation}
Consider the second derivative of  $h(\pi_\theta,x,y_1,y_2)$:
\begin{equation}\label{pra_h_pi_theta_2rd_eq}
    \begin{aligned}
\frac{\partial^2 h(\pi_\theta,x,y_1,y_2)}{\partial \theta(x, y_i)\partial \theta(x, y_j)} =& -\frac{\partial }{\partial \theta(x, y_j) }( \sigma(\frac{1}{\tau}\log\frac{\pi_\theta(y_2|x)}{\pi_\theta(y_1|x)}) \frac{1}{\tau}(\delta_{y_1y_i}-\delta_{y_2y_i}) ) \\
=& \frac{1}{\tau^2}(\delta_{y_1y_i}-\delta_{y_2y_i})(\delta_{y_1y_j}-\delta_{y_2y_j})\sigma(\frac{1}{\tau}\log\frac{\pi_\theta(y_2|x)}{\pi_\theta(y_1|x)})\sigma(\frac{1}{\tau}\log\frac{\pi_\theta(y_1|x)}{\pi_\theta(y_2|x)}).
    \end{aligned}
\end{equation}
Based on Lemma \ref{Pair_second_derivative_lemma}, the second derivative of $f_{\text{PRA}}(x,\theta)$ is:
\begin{equation}\label{PRA_decompose}
    \begin{aligned}
        \frac{\partial^2 f_{\text{PRA}}(x, \theta)}{\partial \theta(x, y_i) \partial \theta(x, y_j)} = \sum_{y_1,y_2\in\mathbb{Y}}& 2\frac{\partial^2 \pi_\theta(y_1|x)}{\partial \theta(x, y_i) \partial \theta(x, y_j)} \pi_\theta(y_2|x)h(\pi_\theta,x,y_1,y_2) + 2\frac{\partial \pi_\theta(y_1|x)}{\partial \theta(x, y_i)}\frac{\partial \pi_\theta(y_2|x)}{\partial \theta(x, y_j)}h(\pi_\theta,x,y_1,y_2) \\
        &+ 2\frac{\partial \pi_\theta(y_1|x)}{\partial \theta(x, y_i)}\pi_\theta(y_2|x) \frac{\partial h(\pi_\theta, x, y_1,y_2)}{\partial \theta(x, y_j)}+ 2\frac{\partial \pi_\theta(y_1|x)}{\partial \theta(x, y_j)}\pi_\theta(y_2|x) \frac{\partial h(\pi_\theta, x, y_1,y_2)}{\partial \theta(x, y_i)}\\
        & + \pi_\theta(y_1|x)\pi_\theta(y_2|x)\frac{\partial^2 h(\pi_\theta, x, y_1,y_2)}{\partial \theta(x, y_i) \partial \theta(x, y_j)}.
    \end{aligned}
\end{equation}

According to the absolute value inequality, $|\psi(x)|=\left|\sum_{y_i,y_j\in\mathbb{Y}}z(x,y_i) \frac{\partial^2 f_{\text{PRA}}(x,\theta)}{\partial \theta(x, y_i) \partial \theta(x, y_j)}  z(x,y_j)\right|$ can be decomposed into five parts based on Eq.\ref{PRA_decompose} for further analysis.
\begin{equation}\label{PRA_decompose_abs}
    \begin{aligned}
&\left|\sum_{y_i,y_j\in\mathbb{Y}}z(x,y_i) \frac{\partial^2 f_{\text{PRA}}(x,\theta)}{\partial \theta(x, y_i) \partial \theta(x, y_j)}  z(x,y_j)\right| \\
\leq& \left|\sum_{y_i,y_j\in\mathbb{Y}}z(x,y_i)  \sum_{y_1,y_2\in\mathbb{Y}} 2\frac{\partial^2 \pi_\theta(y_1|x)}{\partial \theta(x, y_i) \partial \theta(x, y_j)} \pi_\theta(y_2|x)h(\pi_\theta,x,y_1,y_2)  z(x,y_j)\right| \\
&+\left|\sum_{y_i,y_j\in\mathbb{Y}}z(x,y_i) \sum_{y_1,y_2\in\mathbb{Y}}  2\frac{\partial \pi_\theta(y_1|x)}{\partial \theta(x, y_i)}\frac{\partial \pi_\theta(y_2|x)}{\partial \theta(x, y_j)}h(\pi_\theta,x,y_1,y_2) z(x,y_j)\right| \\
&+\left|\sum_{y_i,y_j\in\mathbb{Y}}z(x,y_i) \sum_{y_1,y_2\in\mathbb{Y}}  4\frac{\partial \pi_\theta(y_1|x)}{\partial \theta(x, y_i)}\pi_\theta(y_2|x) \frac{\partial h(\pi_\theta, x, y_1,y_2)}{\partial \theta(x, y_j)}  z(x,y_j)\right|  \\
&+\left|\sum_{y_i,y_j\in\mathbb{Y}}z(x,y_i) \sum_{y_1,y_2\in\mathbb{Y}} \pi_\theta(y_1|x)\pi_\theta(y_2|x)\frac{\partial^2 h(\pi_\theta, x, y_1,y_2)}{\partial \theta(x, y_i) \partial \theta(x, y_j)}  z(x,y_j)\right|.
    \end{aligned}
\end{equation}

Similar to Eq.\ref{RDA_1_2_terms} and its following derivation, we have:
\begin{equation} 
\begin{aligned}
    & \left|\sum_{y_i,y_j\in\mathbb{Y}}z(x,y_i)  \sum_{y_1,y_2\in\mathbb{Y}} 2\frac{\partial^2 \pi_\theta(y_1|x)}{\partial \theta(x, y_i) \partial \theta(x, y_j)} \pi_\theta(y_2|x)h(\pi_\theta,x,y_1,y_2)  z(x,y_j)\right| \\
&+ \left|\sum_{y_i,y_j\in\mathbb{Y}}z(x,y_i) \sum_{y_1,y_2\in\mathbb{Y}}  2\frac{\partial \pi_\theta(y_1|x)}{\partial \theta(x, y_i)}\frac{\partial \pi_\theta(y_2|x)}{\partial \theta(x, y_j)}h(\pi_\theta,x,y_1,y_2) z(x,y_j)\right| \\
\leq&\left|\sum_{y_i,y_j\in\mathbb{Y}}z(x,y_i)  \sum_{y_1,y_2\in\mathbb{Y}}4  \pi_\theta(y_1 | x)\pi_\theta(y_2 | x)(\delta_{y_1y_i} - \pi_\theta(y_i | x))(\delta_{y_2y_j} - \pi_\theta(y_j | x))\log(1+ e^{\frac{d}{\tau}}) z(x,y_j)\right| \\
&+\left|\sum_{y_i,y_j\in\mathbb{Y}}z(x,y_i) \sum_{y_1,y_2\in\mathbb{Y}}2\pi_\theta(y_1 | x)\pi_\theta(y_2 | x) \pi_\theta(y_i | x)(\delta_{y_iy_j} - \pi_\theta(y_j | x))\log(1+ e^{\frac{d}{\tau}})  z(x,y_j)\right| \\
\leq& 20\log(1+ e^{\frac{d}{\tau}}).
\end{aligned}
\end{equation}

Similar to Eq.\ref{RDA_3_4_terms}, for the third term of Eq.\ref{PRA_decompose_abs}, we have:
\begin{equation}
\begin{aligned}
& \left|\sum_{y_i,y_j\in\mathbb{Y}}z(x,y_i) \sum_{y_1,y_2\in\mathbb{Y}}  4\frac{\partial \pi_\theta(y_1|x)}{\partial \theta(x, y_i)}\pi_\theta(y_2|x) \frac{\partial h(\pi_\theta, x, y_1,y_2)}{\partial \theta(x, y_j)}  z(x,y_j)\right|  \\
=& 4\left|\sum_{y_i,y_j\in\mathbb{Y}}z(x,y_i) \sum_{y_1,y_2\in\mathbb{Y}}  \frac{\partial \pi_\theta(y_1|x)}{\partial \theta(x, y_i)}\pi_\theta(y_2|x) \left(p^* - \sigma(\frac{1}{\tau}\log\frac{\pi_\theta(y_1|x)}{\pi_\theta(y_2|x)}) \right) \frac{1}{\tau}(\delta_{y_1y_j}-\delta_{y_2y_j})  z(x,y_j)\right| \\
\leq& 4\left|\sum_{y_i,y_j\in\mathbb{Y}}z(x,y_i) \sum_{y_1,y_2\in\mathbb{Y}}  \frac{\partial \pi_\theta(y_1|x)}{\partial \theta(x, y_i)}\pi_\theta(y_2|x) \epsilon_3    \frac{1}{\tau}(\delta_{y_1y_j}-\delta_{y_2y_j})  z(x,y_j)\right| \\
=& \frac{4\epsilon_3}{\tau} \left|\sum_{y_i,y_j\in\mathbb{Y}}z(x,y_i) \sum_{y_1,y_2\in\mathbb{Y}}  \pi_\theta(y_1|x)(\delta_{y_1y_i}-\pi_\theta(y_i|x)) \pi_\theta(y_2|x) \epsilon_3    \frac{1}{\tau}(\delta_{y_1y_j}-\delta_{y_2y_j})  z(x,y_j)\right| \\
\leq&  \frac{16\epsilon_3}{\tau}\|z(x,\cdot)\|_2^2.
\end{aligned}
\end{equation}
The first inequality is because $|p^* - \sigma(\frac{1}{\tau}\log\frac{\pi_\theta(y_1|x)}{\pi_\theta(y_2|x)}) |\leq \epsilon_3$ where $p^*=p^*(1|x,y_1,y_2)$. The last inequality is because we can expand the absolute value operation in the second-to-last line into four terms, and these four terms can be deduced to be less than $\|z(x,\cdot)\|_2^2$ using some existing conclusions. The relevant conclusions are:
$ \pi_\theta^\top z(x,\cdot) \leq \|\pi_\theta\|_1 \|z(x,\cdot)\|_\infty = \|z(x,\cdot)\|_\infty\leq\|z(x,\cdot)\|_2 $, $ \pi_\theta^\top z^2(x,\cdot) \leq \|z(x,\cdot)\|_2^2$, $ \|\pi_\theta\|_\infty \leq 1 $, $ \|z^2(x,\cdot)\|_1 = \|z(x,\cdot)\|_2^2 $ and $|(\pi_\theta^\top z(x,\cdot))^2|  \leq \|z(x,\cdot)\|_2^2$.

Similar to Eq.\ref{RDA_5_terms}, for the fifth term of Eq.\ref{PRA_decompose_abs}, based on Eq.\ref{pra_h_pi_theta_2rd_eq} and Eq.\ref{DPO_4term_decompose}, we have:
\begin{equation}
        \begin{aligned}
&\left|\sum_{y_i,y_j\in\mathbb{Y}}z(x,y_i) \sum_{y_1,y_2\in\mathbb{Y}} \pi_\theta(y_1|x)\pi_\theta(y_2|x)\frac{\partial^2 h(\pi_\theta, x, y_1,y_2)}{\partial \theta(x, y_i) \partial \theta(x, y_j)}  z(x,y_j)\right| \\
=& \left|\sum_{y_i,y_j\in\mathbb{Y}}z(x,y_i) \sum_{y_1,y_2\in\mathbb{Y}} \pi_\theta(y_1|x)\pi_\theta(y_2|x)  (\delta_{y_1y_i}-\delta_{y_2y_i})(\delta_{y_1y_j}-\delta_{y_2y_j})(\frac{1}{\tau^2}\sigma(\frac{1}{\tau}\log\frac{\pi_\theta(y_2|x)}{\pi_\theta(y_1|x)})\sigma(\frac{1}{\tau}\log\frac{\pi_\theta(y_1|x)}{\pi_\theta(y_2|x)}))
    z(x,y_j)\right| \\
\leq& \left|\sum_{y_i,y_j\in\mathbb{Y}}z(x,y_i) \sum_{y_1,y_2\in\mathbb{Y}} \pi_\theta(y_1|x)\pi_\theta(y_2|x)(\delta_{y_1y_i}-\delta_{y_2y_i})(\delta_{y_1y_j}-\delta_{y_2y_j})\frac{1}{\tau^2}  z(x,y_j)\right| \\
\leq& \frac{4}{\tau^2}\|z(x,\cdot)\|_2^2.
    \end{aligned}
\end{equation}
In summary, we have the upper bound of Eq.\ref{PRA_decompose_abs}:
\begin{equation}
    \begin{aligned}
\|\psi(x)\|_\infty=&\max_{x\in\mathbb{X}}|\psi(x)|=\max_{x\in\mathbb{X}}\left|\sum_{y_i,y_j\in\mathbb{Y}}z(x,y_i) \frac{\partial^2 f_{\text{PRA}}(x,\theta)}{\partial \theta(x, y_i) \partial \theta(x, y_j)}  z(x,y_j)\right| \\
\leq& \left(20\log(1+ e^{\frac{d}{\tau}})+\frac{16\epsilon_3}{\tau}+\frac{4}{\tau^2}\right)\|z(\cdot,\cdot)\|_2^2
    \end{aligned}
\end{equation}
As for the second term $\left|\sum_{x\in\mathbb{X}}\mathcal{D}(x)\sum_{y_i,y_j\in\mathbb{Y}}z(x,y_i)  \frac{\partial^2 \sum_{y_1,y_2\in\mathbb{Y}}\pi_\theta(y_1|x)\pi_\theta(y_2|x)M(x,y_1,y_2)}{\partial \theta(x, y_i) \partial \theta(x, y_j)}  z(x,y_j)\right|$ of Eq.\ref{pra_decompose_psi_with_M_eq}, we have:
\begin{equation}
    \begin{aligned}
&|\frac{\partial^2 \sum_{y_1,y_2\in\mathbb{Y}}\pi_\theta(y_1|x)\pi_\theta(y_2|x)M(x,y_1,y_2)}{\partial \theta(x, y_i) \partial \theta(x, y_j)}| \\
=&| \frac{\partial}{\partial \theta(x, y_j) }(\frac{\partial \sum_{y_1,y_2\in\mathbb{Y}}\pi_\theta(y_1|x)\pi_\theta(y_2|x)M(x,y_1,y_2)}{\partial \theta(x, y_i) })|\\
=& |\frac{\partial}{\partial \theta(x, y_j) }\left(\sum_{y_1,y_2\in\mathbb{Y}}\pi_\theta(y_1|x)\pi_\theta(y_2|x)(\delta_{y_1y_i}+\delta_{y_2y_i} -2\pi_\theta(y_i|x))M(x,y_1,y_2)\right) |\\
=& |\sum_{y_1,y_2\in\mathbb{Y}}\pi_\theta(y_1|x)\pi_\theta(y_2|x)(\delta_{y_1y_i}+\delta_{y_2y_i} -2\pi_\theta(y_i|x))(\delta_{y_1y_j}+\delta_{y_2y_j} -2\pi_\theta(y_j|x))M(x,y_1,y_2).
    \end{aligned}
\end{equation}
Based on Lemma \ref{SelfEntropy}. We have:
\begin{equation}
    \begin{aligned}
&\left|\sum_{x\in\mathbb{X}}\mathcal{D}(x)\sum_{y_i,y_j\in\mathbb{Y}}z(x,y_i)  \frac{\partial^2 \sum_{y_1,y_2\in\mathbb{Y}}\pi_\theta(y_1|x)\pi_\theta(y_2|x)M(x,y_1,y_2)}{\partial \theta(x, y_i) \partial \theta(x, y_j)}  z(x,y_j)\right| 
\leq 16\log2 \cdot\|z(\cdot,\cdot)\|_2^2
    \end{aligned}
\end{equation}
This inequality is because we can expand the absolute value operation in the second-to-last line into nine terms, and these nine terms can be deduced to be less than $\|z(x,\cdot)\|_2^2$ or $2\|z(x,\cdot)\|_2^2$ using some existing conclusions. The relevant conclusions are:
$ \pi_\theta^\top z(x,\cdot) \leq \|\pi_\theta\|_1 \|z(x,\cdot)\|_\infty = \|z(x,\cdot)\|_\infty\leq\|z(x,\cdot)\|_2 $, $ \pi_\theta^\top z^2(x,\cdot) \leq \|z(x,\cdot)\|_2^2$, $ \|\pi_\theta\|_\infty \leq 1 $, $ \|z^2(x,\cdot)\|_1 = \|z(x,\cdot)\|_2^2 $ and $|(\pi_\theta^\top z(x,\cdot))^2|  \leq \|z(x,\cdot)\|_2^2$.

Then Eq.\ref{spectral_radius_pra} is proved. Proof finished.

\end{document}